%% file: main.tex
\documentclass[
fontsize=10pt,
abstract,
twocolumn,
numbers=enddot  
]{scrartcl}

\input{packages/structure}
\input{packages/math}
\input{packages/add_ons}
\input{packages/draw}

\makeatletter
\let\Subject\@subject
\title{Inertial Line-Of-Sight Stabilization Using a 3-DOF Spherical Parallel Manipulator with Coaxial Input Shafts}
\let\Title\@title
\subtitle{}
\let\Subtitle\@subtitle
\author[(1),(2),(3)]{Alexandre \textsc{Lê}}
\author[(1)]{Guillaume \textsc{Rance}}
\author[(2),(3)]{Fabrice \textsc{Rouillier}}
\author[(4)]{Damien \textsc{Chablat}}
\affil[(1)]{Safran Electronics \& Defense, 100 avenue de Paris, 91344 Massy CEDEX, Île-de-France, France (e-mail: \{alexandre-thanh.le, guillaume.rance\}@safrangroup.com).}
\affil[(2)]{Sorbonne Université, Université de Paris Cité, Institut de
Mathématiques de Jussieu Paris Rive Gauche, 4 place Jussieu, 75252 Paris CEDEX 05, Île-de-France, France.}
\affil[(3)]{Inria Paris, 2 rue Simone Iff, 75012 Paris, Île-de-France, France (e-mail: \{alexandre.le, fabrice.rouillier\}@inria.fr).}
\affil[(4)]{Nantes Université, École Centrale Nantes, CNRS, LS2N, UMR 6004, F-44000, Nantes, France (e-mail: damien.chablat@cnrs.fr)}
\let\Author\@author
\date{}
\let\Date\@date

\newcommand{\logo}{
}
\titlehead{\logo}
\publishers{\vspace*{-2em}}
\newcommand{\paperID}{00037}
\makeatother

\begin{document}
	\maketitle
    \hypersetup{hidelinks}
	
	\begin{spacing}{1}
        \input{contents/1}
	\end{spacing}

    \bibliographystyle{apalike}
    \bibliography{IEEEabrv,contents/ref}

    \begin{appendices}
        \input{contents/appendix}
    \end{appendices}
\end{document}

%% file: packages/structure.tex
\usepackage[T1]{fontenc}
\usepackage[utf8]{inputenc}

\usepackage{helvet}

\usepackage[left=2cm,right=2cm,top=2.5cm,bottom=2.5cm,
columnsep=1cm]{geometry}

\setkomafont{subject}{\large\bfseries}
\setkomafont{title}{\bfseries}

\usepackage{xpatch}
\makeatletter
\patchcmd{\@maketitle}{\huge}{\large\MakeUppercase}{}{}
\makeatother

\setkomafont{subtitle}{\large\bfseries}
\setkomafont{author}{\normalsize}
\setkomafont{date}{\small}
\setkomafont{publishers}{\small}

\setkomafont{section}{\normalsize}
\addtokomafont{section}{\MakeUppercase}
\setkomafont{subsection}{\normalsize}
\setkomafont{subsubsection}{\normalsize\normalfont\itshape}

\usepackage[
colorlinks=true,
linkcolor=dblue,
bookmarks=true,
bookmarksopen=true,
citecolor=magenta,
urlcolor=cyan,
backref=page,
breaklinks,
pdfusetitle
]{hyperref}

\makeatletter
\hypersetup{
pdftitle={\@title},
pdfauthor={\@author},
pdfsubject={\@subject},
pdfkeywords={}
}
\makeatother

\usepackage{lastpage}
\AtEndDocument{\label{lastpage}}

\setkomafont{pageheadfoot}{\normalfont\normalcolor} 

\usepackage[
]{scrlayer-scrpage}
\deftriplepagestyle{first}
    {}
    {}
    {\textbf{OPTRO2024\_\paperID}}
    {}
    {\thepage}
    {}
\clearpairofpagestyles
\ihead{}
\ohead{}
\usepackage[page,toc]{appendix} 

\let\appendixpagenameorig\appendixpagename
\renewcommand{\appendixpagename}{\normalsize\MakeUppercase\appendixpagenameorig}

\usepackage{authblk}

\usepackage{caption}
\DeclareCaptionLabelSeparator{period}{.~}
\let\figurenameorig\figurename
\renewcommand{\figurename}{\itshape\figurenameorig}
\let\tablenameorig\tablename
\renewcommand{\tablename}{\itshape\tablenameorig}

%% file: packages/math.tex
\usepackage{amsmath,amsthm,amsfonts,amssymb,esint} 

\docsvlist{A,B,C,D,E,F,G,H,I,J,K,L,M,N,O,P,Q,R,S,T,U,V,W,X,Y,Z}

\docsvlist{A,B,C,D,E,F,G,H,I,J,K,L,M,N,O,P,Q,R,S,T,U,V,W,X,Y,Z}

\docsvlist{a,b,c,d,e,f,g,h,i,j,k,l,m,n,o,p,q,r,s,t,u,v,w,x,y,z,A,B,C,D,E,F,G,H,I,J,K,L,M,N,O,P,Q,R,S,T,U,V,W,X,Y,Z}

\docsvlist{a,b,c,d,e,f,g,h,i,j,k,l,m,n,o,p,q,r,s,t,u,v,w,x,y,z,A,B,C,D,E,F,G,H,I,J,K,L,M,N,O,P,Q,R,S,T,U,V,W,X,Y,Z}

\usepackage{bigints} 

\usepackage{mathtools}
\newcommand{\byaff}{\coloneqq}
\newcommand{\bydef}{\triangleq}

\DeclareMathOperator{\e}{e}
\DeclareMathOperator{\im}{i}

\usepackage{xstring}
\noexpandarg 

	\DeclareRobustCommand{\dder}{
		\mathop{}\mathopen{}\mathrm{d}
	}

	\newcommand{\dd}[2][0]{
		\IfStrEq{#1}{0}{
			\dder #2
		}{
			\IfBeginWith{#2}{f}{
				\dder^{#1} \! #2
			}{
				\dder^{#1}  #2
			}
		}
	}

	\newcommand{\dv}[3][0]{
		\IfStrEq{#1}{0}{
			\ensuremath{\frac{\dd{#2}}{\dd{#3}}}
		}{
			\ensuremath{\frac{\dd[#1]{#2}}{\dd{#3}^{#1}}}
		}
	}

	\makeatletter
		\let\original@partial\partial
		\renewcommand{\partial}{
			\original@partial\mathopen{}
		}
	\makeatother

	\newcommand{\pp}[2][0]{
		\IfStrEq{#1}{0}{
			\partial #2
		}{
			\IfBeginWith{#2}{f}{
				\partial^{#1} \! #2
			}{
				\partial^{#1} #2
			}
		}
	}

	\newcommand\pdv[3][0]{%
		\def\deno{\StrSubstitute{ #3}{ }\partial}
		\def\denoOK{\StrSubstitute{\deno}{\partial f}{\partial \! f}}
		\IfStrEq{#1}{0}{
			\ensuremath{\frac{\pp{#2}}{\denoOK}}
		}{
			\ensuremath{\frac{\pp^{#1}{#2}}{\denoOK}}
		}
	}

\usepackage{scalerel}

\DeclareMathOperator{\discrim}{discrim}

\DeclareMathOperator{\sgn}{sgn}

\DeclareMathOperator{\diag}{\bm{\mathrm{diag}}}

\DeclareMathOperator{\id}{\mathbf{I}}

\usepackage{stmaryrd} 
\SetSymbolFont{stmry}{bold}{U}{stmry}{m}{n}
\usepackage{bm}

\newcommand{\rep}[2]{\prescript{#2}{}{#1}}

\newcommand\abs[1]{\left|#1\right|}

\newcommand\mat[1]{\left[\;\begin{matrix}#1\end{matrix}\;\right]}
\newcommand\bra[1]{\left\langle#1\right\rangle}
\newcommand\itvd[1]{\left\llbracket#1\right\rrbracket}

\newcommand{\1}{\text{\usefont{U}{bbold}{m}{n}1}}
\MakeRobust{\1}

\usepackage{siunitx}

%% file: packages/add_ons.tex
\usepackage{blindtext}
\usepackage{lipsum}

\newcommand\tss[1]{\texorpdfstring{\textsuperscript{#1}}{#1}}	  

\usepackage{setspace}							  
\usepackage{graphicx} 						  
\usepackage[percent]{overpic}       
\usepackage{enumitem}               

\usepackage{subcaption}			
\usepackage{multirow}
\usepackage{hhline}

\usepackage{standalone}			
\usepackage{pdfpages}			

\makeatletter
\def\th@plain{%
  \thm@notefont{}
  \itshape 
}
\def\th@definition{%
  \thm@notefont{}
  \normalfont 
}
\makeatother

\newtheorem{theorem}{Theorem}[]
\newtheorem{proposition}{Proposition}[]

\newtheorem{lemma}[theorem]{Lemma}

\newtheorem{remark}{Remark}[]

\renewcommand*{\backref}[1]{}
\renewcommand*{\backrefalt}[4]{%
	{
 	\ifcase #1(No~citation)
	\or(Citation~on~page~#2)
	\else(Citation~on~pages~#2)
	\fi%
	}
}

\usepackage[auto-label=true]{exsheets}
\DeclareInstance{exsheets-heading}{block-subtitle}{default}{
  join = {
    title[r,B]number[l,B](.333em,0pt) ;
    title[r,B]subtitle[l,B](1em,0pt) ;
    title[l,B]points[l,B](\linewidth+\marginparsep,0pt) ;
    main[l,T]title[l,T](0pt,0pt) ;
  } ,
  subtitle-pre-code = \parbox[t]{.8\linewidth}
}
\SetupExSheets{
  headings = runin, 
  question/type = exam,
  counter-format=qu[1]~--,
  points/name=p.,
  points/format=\color{black!50},
  question/skip-below=0.25\baselineskip
}

\usepackage[ddmmyyyy]{datetime}

\usepackage[short]{optidef}

\setlist[0]{noitemsep}

%% file: packages/draw.tex
\usepackage{xcolor}
\definecolor{main}{RGB}{0,0,0}
\definecolor{dgreen}{RGB}{0,111,0}
\definecolor{dblue}{RGB}{0,0,160}
\definecolor{dred}{RGB}{223,0,0}
\definecolor{safran}{RGB}{20,119,198}
\definecolor{bleu_identitaire}{RGB}{1,68,145}
\definecolor{gris_identitaire}{RGB}{82,86,104}

\usepackage{pgf,pgfplots,tikz}
\usetikzlibrary{calc,arrows}
\pgfplotsset{width=10cm,compat=1.9}
\usepackage{tikz-cd}
\usepackage{tikz-3dplot}
\usepackage{graphicx}
\usetikzlibrary{decorations.pathmorphing,patterns}
\usetikzlibrary{automata,positioning}				

\usepackage[oldvoltagedirection]{circuitikz}
\usetikzlibrary{babel}

\usepackage{schemabloc}
\usepackage{bodegraph}

\tikzstyle{block} = [draw, fill=white, rectangle, 
        minimum height=3em, minimum width=3em]
\tikzstyle{mblock} = [draw, fill=white, rectangle, 
        minimum height=1.75em, minimum width=3em]       
\tikzstyle{tblock} = [draw, fill=white, rectangle,       
        minimum height=5em, minimum width=3em]          
\tikzstyle{bblock} = [draw, fill=white, rectangle,       
        minimum height=10em, minimum width=3em]          
\tikzstyle{sblock} = [draw, fill=white, rectangle, 
        minimum height=1.5em, minimum width=3em]
\tikzstyle{sum} = [draw, fill=white, circle, node distance=1cm]
\tikzstyle{input} = [coordinate]
\tikzstyle{output} = [coordinate]
\tikzstyle{pinstyle} = [pin edge={to-,thin,black}]

\tikzstyle{decision} = [diamond, draw, fill=blue!10, 
    text width=7.5em, text badly centered, node distance=4cm, inner sep=0pt]
\tikzstyle{etape} = [rectangle, draw, fill=blue!10, 
    text centered, rounded corners, minimum height=4em]
\tikzstyle{line} = [draw, -latex']
\tikzstyle{cloud} = [draw, ellipse,fill=red!10, node distance=3cm,
    minimum height=2em]

\newcommand{\nvar}[2]{%
    \newlength{#1}
    \setlength{#1}{#2}
}

\nvar{\dg}{0.3cm}

\nvar{\ddx}{1.15cm} 

\pgfdeclarelayer{background}
\pgfdeclarelayer{foreground}
\pgfsetlayers{background,main,foreground}

\newcommand{\tdcyl}[5]{
    \path (1,0,0);
    \pgfgetlastxy{\cylxx}{\cylxy}
    \path (0,1,0);
    \pgfgetlastxy{\cylyx}{\cylyy}
    \path (0,0,1);
    \pgfgetlastxy{\cylzx}{\cylzy}
    \pgfmathsetmacro{\cylt}{(\cylzy * \cylyx - \cylzx * \cylyy)/ (\cylzy * \cylxx - \cylzx * \cylxy)}
    \pgfmathsetmacro{\ang}{atan(\cylt)}
    \pgfmathsetmacro{\ct}{1/sqrt(1 + (\cylt)^2)}
    \pgfmathsetmacro{\st}{\cylt * \ct}
    \filldraw[fill=white] (#4*\ct+#1,#4*\st+#2,#3) -- ++(0,0,#5) arc[start angle=\ang,delta angle=-180,radius=#4] -- ++(0,0,-#5) arc[start angle=\ang+180,delta angle=180,radius=#4];
    \filldraw[fill=white] (#1,#2,#3+#5) circle[radius=#4];
    \filldraw[fill=white] (#1,#2,#3) circle[radius=#4];
}
\usetikzlibrary{spy}    

%% file: contents/1.tex
\thispagestyle{first}
\medskip\par
\noindent\textbf{ABSTRACT:} 
\par\smallskip
\noindent This article dives into the use of a 3-\underline{R}RR Spherical Parallel Manipulator (SPM) for the purpose of inertial Line Of Sight (LOS) stabilization. Such a parallel robot provides three Degrees of Freedom (DOF) in orientation and is studied from the kinematic point of view. In particular, one guarantees that the singular loci (with the resulting numerical instabilities and inappropriate behavior of the mechanism) are far away from the prescribed workspace. Once the kinematics of the device is certified, a control strategy needs to be implemented in order to stabilize the LOS through the upper platform of the mechanism. Such a work is done with \textsc{Matlab} Simulink\textsuperscript{\textregistered} using a SimMechanics\texttrademark{} model of our robot.

\section{Introduction}

\subsection{Context of the study}

A classical approach for Line Of Sight (LOS) stabilization involves the use of gimbal-based systems \cite{Mas08,Hil08}. Such devices behave like \emph{serial} robots and provide up to two Degrees Of Freedom (DOF) in orientation. This article is focusing on LOS stabilization using a \emph{parallel robot} that provides three DOF in orientation. Unlike their serial counterparts, parallel manipulators are closed-loop kinematic chains with at least two legs mostly actuated at their bases (the other joints are then passive). As a result, many parts of the robot are subject to traction/compression constraints so that it is possible to use less powerful actuators. They are also known for presenting very good performance in terms of dynamics, stiffness and accuracy. More general information about parallel robots can be found in \cite{Mer06}. Nowadays, the vast majority of commercial gyrostabilized sights uses gimbal systems for a LOS stabilization w.r.t.~two rotational DOF: unlimited bearing, and elevation. Given all the before-mentioned advantages of parallel robots, the idea is to use a Spherical Parallel Manipulator (SPM) \cite{GH94} with coaxial input shafts (CoSPM) to inertially stabilize the LOS w.r.t.~a third rotational DOF called bank. Such an upgrade improves the quality of LOS stabilization in the sense that the field of view is mechanically non-rotating around the LOS axis.

\begin{figure}[htbp]
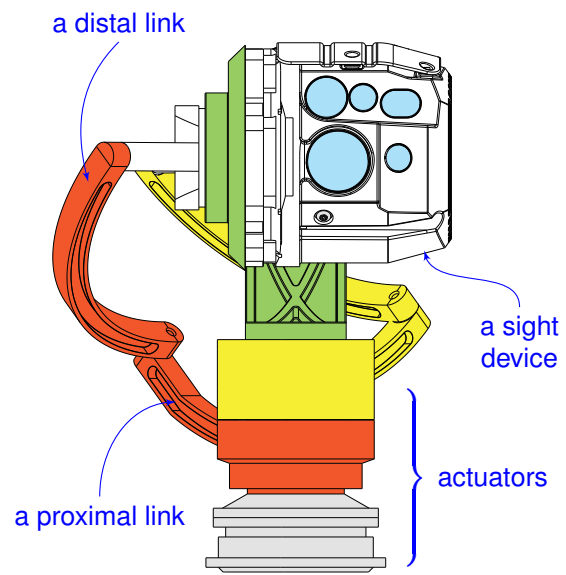

	\centering
	\includestandalone[width=7.5cm]{TikZ/spm_d1_legend}
	\caption{Illustration of the SPM with coaxial input shafts of interest}
	\label{fig:spm_d1_legend}
\end{figure}

\vspace*{-1.5em}
\subsection{Presentation of the mechanism}\label{ss:pres}

\begin{figure*}[ht!]
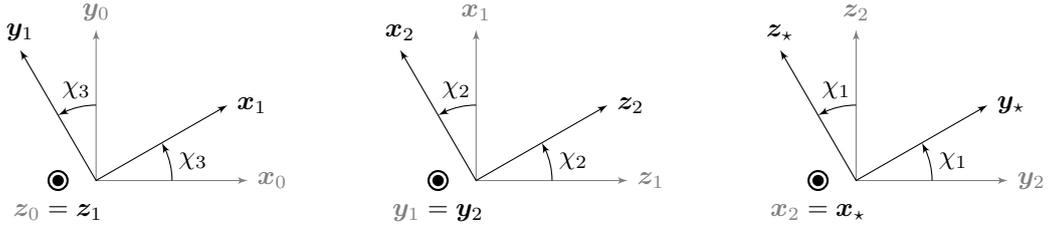

	\centering
	\includestandalone[]{TikZ/rotation_order_zyx}
	\caption{Rotation order using the Euler Tait-Bryan ZYX convention}
	\label{fig:rotation_order}
\end{figure*}

The CoSPM of interest is depicted in Figure \ref{fig:spm_d1_legend}. It is a parallel robot with three kinematic chains (the first one in red, the second one in yellow and the third one in green). Such a manipulator is \emph{spherical} in the sense that the upper platform only makes pure spherical motions that are described in this article as a composition of three elementary rotations using the Euler Tait-Bryan ZYX convention as illustrated in Figure \ref{fig:rotation_order}. Given that formalism, the device can \emph{at least} make:
\begin{itemize}
	\item unlimited bearing $\chi_3$ (rotation around the $z_0$-axis) thanks to its coaxial input shafts;
	\item an elevation angle $\chi_2$ (rotation around the $y_1$-axis) of $100^\circ$;
	\item a bank angle $\chi_1$ (rotation around the $x_2$-axis) of $\pm10^\circ$.
\end{itemize}

\begin{figure}[htbp]
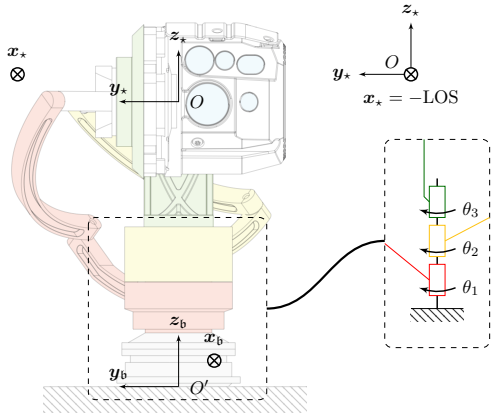

	\centering
	\includestandalone[width=6.5cm]{TikZ/spm_d1_frames} 
	\caption{Tools for the description of the SPM}
	\label{fig:spm_d1_frames}
\end{figure}

This subset of the workspace of interest is called \emph{prescribed regular workspace} and will be denoted as $\calW^\star$ in the sequel. The three before-mentioned DOFs in $\chi_1$, $\chi_2$ and $\chi_3$ are called \emph{orientation} of the sight. It is worth stressing that such orientations are obtained through the angular motion of the three actuators located at the base of the SPM as shown in Figure \ref{fig:spm_d1_frames} (the other joints being passive).

In the sequel, one can put the orientation angles into the vector $\bm{\chi}\bydef\mat{\chi_1 & \chi_2 & \chi_3}^\sfT$ whereas the three actuators can be described by the actuated joint vector $\bm{\theta}\bydef\mat{\theta_1 & \theta_2 & \theta_3}^\sfT$. Moreover, one introduces the basis that are useful for the definition of the design parameters of the mechanism:
\begin{itemize}
	\item $\calB_\frakb\bydef\left\{\bm{x}_\frakb,\bm{y}_\frakb,\bm{z}_\frakb\right\}$ related to the base of the robot; 
	\item $\calB_\star\bydef\left\{\bm{x}_\star,\bm{y}_\star,\bm{z}_\star\right\}$ related to the upper platform containing the sight device.
\end{itemize}
At home configuration, one obviously has $\calB_\frakb\equiv\calB_\star$. Furthermore, the axis $-\bm{x}_\star$ defines the LOS axis of the device given our conventions.

\begin{figure}[ht!]
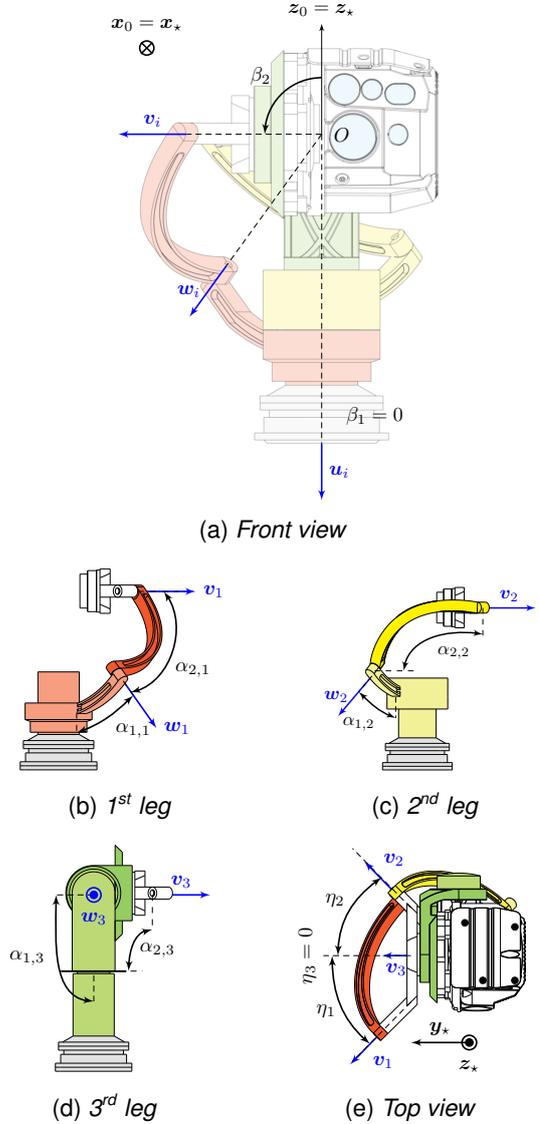

	\centering
	\begin{subfigure}{0.99\columnwidth}
		\centering
		\includestandalone[width=.575\columnwidth]{TikZ/spm_d1_beta_vec}
		\caption{Front view}
		\label{fig:spm_d1_front}
	\end{subfigure}
	\par\vspace*{0.5em}
	\begin{subfigure}{0.49\columnwidth}
		\centering
		\includestandalone[width=.75\columnwidth]{TikZ/spm_d1_jambe1}
		\caption{1\tss{st} leg}
		\label{fig:spm_d1_jambe1}
	\end{subfigure}
	\begin{subfigure}{0.49\columnwidth}
		\centering
		\includestandalone[width=.75\columnwidth]{TikZ/spm_d1_jambe2}
		\caption{2\tss{nd} leg}
		\label{fig:spm_d1_jambe2}
	\end{subfigure}
	\par\vspace*{0.5em}
	\begin{subfigure}{0.44\columnwidth}
		\centering
		\includestandalone[width=.8\columnwidth]{TikZ/spm_d1_jambe3}
		\caption{3\tss{rd} leg}
		\label{fig:spm_d1_jambe3}
	\end{subfigure}
	\begin{subfigure}{0.54\columnwidth}
		\centering
		\includestandalone[width=0.75\columnwidth]{TikZ/spm_d1_eta}
		\caption{Top view}
		\label{fig:spm_d1_top}
	\end{subfigure}
	\caption{Details on the design parameters and the vectors of the SPM with coaxial input shafts}
	\label{fig:spm_d1}
\end{figure}

\begin{table*}[ht!]
    \centering
    {\renewcommand{\arraystretch}{1.25} 
    \begin{tabular}{cccc}
    \hline\hline
    Design parameter 		& Notation 	& Value (rad) & Definition\\\hline\hline
	\multirow{3}{*}{\begin{tabular}[c]{@{}c@{}}Proximal link $\alpha_{1,i}$\\($i$\textsuperscript{th} leg)\end{tabular}}	
								& $\alpha_{1,1}$	& $\pi/4$	& $\angle\left(\bm{u}_1,\bm{w}_1\right)$\\
								& $\alpha_{1,2}$	& $\pi/4$ 	& $\angle\left(\bm{u}_2,\bm{w}_2\right)$\\
								& $\alpha_{1,3}$	& $\pi/2$ 	& $\angle\left(\bm{u}_3,\bm{w}_3\right)$ \\\hline
	\multirow{3}{*}{\begin{tabular}[c]{@{}c@{}}Distal link $\alpha_{2,i}$\\($i$\textsuperscript{th} leg)\end{tabular}}
								& $\alpha_{2,1}$	& $\pi/2$	& $\angle\left(\bm{w}_1,\bm{v}_1\right)$\\
								& $\alpha_{2,2}$	& $\pi/2$	& $\angle\left(\bm{w}_2,\bm{v}_2\right)$\\
								& $\alpha_{2,3}$	& $\pi/2$ 	& $\angle\left(\bm{w}_3,\bm{v}_3\right)$\\\hline
	\multirow{3}{*}{\begin{tabular}[c]{@{}c@{}}Pivot linkage disposition $\eta_i$\\($i$\textsuperscript{th} leg)\end{tabular}}
								& $\eta_1$			& $\pi/4$	& $\angle\left(\bm{y}_\star,\bm{v}_1\right)$\\
								& $\eta_2$			& $-\pi/4$	& $\angle\left(\bm{y}_\star,\bm{v}_2\right)$\\
								& $\eta_3$			& $0$		& $\angle\left(\bm{y}_\star,\bm{v}_3\right)$\\\hline
	Inner platform's geometry	& $\beta_1$			& $0$		& $\angle\left(\bm{u}_i,-\bm{z}_\frakb\right)$\\
	Upper platform's geometry	& $\beta_2$			& $\pi/2$	& $\angle\left(\bm{z}_\star,\bm{v}_i\right)$\\\hline\hline
    \end{tabular}
    }
	\caption{Design parameters of the SPM of interest}
	\label{tab:parameters}
\end{table*}

Figure \ref{fig:spm_d1} and Table \ref{tab:parameters} highlight the several design parameters of the robot. Each leg $i$ (3 in total, see Fig. \ref{fig:spm_d1_jambe1}--\ref{fig:spm_d1_jambe3}):
\begin{itemize}
	\item is attached to the base through an actuated revolute joint of rotation axis $\bm{u}_i$;
	\item has two bodies -- a proximal link (lighter shade) of angle $\alpha_{1,i}$ and a distal link (darker shade) of angle $\alpha_{2,i}$ -- that are connected through a passive revolute joint of rotation axis $\bm{w}_i$; \item is then linked to the platform containing the sight device by the means of another passive revolute joint of rotation axis $\bm{v}_i$.
\end{itemize}

Note that this robot is \emph{asymmetrical} in the sense that the proximal link can vary from one leg to another and that the pivot linkages of the upper platform are not regularly spaced (see Fig. \ref{fig:spm_d1_top}).

\section{Kinematic analysis}

\subsection{Geometric model}

Any SPM only makes spherical motions around its center of rotation $O$ that is also the center of mass of the sight device in our case (see Fig. \ref{fig:spm_d1_front}). Using this kinematic property, one can describe its geometry as in \cite{Le23} using vectors $\bm{u}_i$, $\bm{w}_i$ and $\bm{v}_i$. As stressed in Figure  \ref{fig:spm_d1}, all these vectors are concurrent in $O$. The reader may refer to Appendix \ref{a:gm} or \cite{Le23} detailing the expressions of these vectors. One can show that the \emph{geometric model} of a general SPM is established using the following kinematic closure.
\begin{equation}\label{eq:fgm_spm}
	\begin{aligned}
		\bm{f}\left(\bm{\theta},\bm{\chi}\right) &\bydef \mat{\bm{w}_1^{\sfT}(\theta_1)\,\bm{v}_1(\bm{\chi})-\cos{\left(\alpha_{2,1}\right)}\\
    	\bm{w}_2^{\sfT}(\theta_2)\,\bm{v}_2(\bm{\chi})-\cos{\left(\alpha_{2,2}\right)}\\
    	\bm{w}_3^{\sfT}(\theta_3)\,\bm{v}_3(\bm{\chi})-\cos{\left(\alpha_{2,3}\right)}}\\
		&= \bm{0}_{3\times1}
	\end{aligned}
\end{equation}

Appendix \ref{a:gm} shows the detailed geometric model of the SPM of interest. Solving \eqref{eq:fgm_spm} at home configuration, \textit{i.e.} for $\bm{\chi}=\bm{0}$ yields $\bm{\theta}=\frac{\pi}{2}\,\bm{1}_{3\times1}$ given the leaf of solution of interest.

\subsection{First order kinematic model}

Differentiating \eqref{eq:fgm_spm} w.r.t.~time provides the \emph{first order kinematic model}, \textit{i.e.}
\begin{equation}
	\dot{\bm{f}}\left(\bm{\theta},\bm{\chi}\right) = \bm{J}_1\dot{\bm{\chi}} + \bm{J}_2\dot{\bm{\theta}} = \bm{0}_{3\times1}
\end{equation}
where $\dot{\bm{\chi}}$ denotes the vector of the \emph{angular rates} and $\dot{\bm{\theta}}$ the vector of the \emph{joint velocities}. From this expression, one can deduce the first order \emph{Forward} \eqref{eq:fkm1} and \emph{Inverse} \eqref{eq:ikm1} kinematic models:
\begin{subequations}
	\begin{align}
		\dot{\bm{\chi}} &= -\bm{J}_1^{-1}\,\bm{J}_2\,\dot{\bm{\theta}} = \bm{J}\,\dot{\bm{\theta}}\label{eq:fkm1}\\
		\dot{\bm{\theta}} &= -\bm{J}_2^{-1}\,\bm{J}_1\,\dot{\bm{\chi}} = \bm{J}^{-1}\,\dot{\bm{\chi}}\label{eq:ikm1}
	\end{align}
\end{subequations}
with $\bm{J}=-\bm{J}_1^{-1}\bm{J}_2$ being the \emph{Jacobian} matrix of the SPM and $\bm{J}^{-1}=-\bm{J}_2^{-1}\bm{J}_1$ its inverse, both defined through $\bm{J}_1\bydef\partial\bm{f}/\partial\bm{\chi}$ and $\bm{J}_2\bydef\partial\bm{f}/\partial\bm{\theta}$. However, the speed control used for the LOS stabilization requires to express this kinematic model in function of the \emph{velocity} of the sight device. Such a value is expressed w.r.t.~the base in the LOS frame $\calF_\star$ and will be denoted in the sequel as $\rep{\bm{\Omega}_{\star/\frakb}}{\star}$. One obtains the latter through the following mapping
\begin{equation}\label{eq:mapping_Omega_dchi}
	\rep{\bm{\Omega}_{\star/\frakb}}{\star} = \bm{T}(\bm{\chi})\,\dot{\bm{\chi}}
\end{equation}
where
\begin{equation}
	\bm{T}(\bm{\chi}) = \mat{1 & 0 & -\sin(\chi_2)\\0 & \cos(\chi_1) & \sin(\chi_1)\cos(\chi_2)\\0 & -\sin(\chi_1) & \cos(\chi_1)\cos(\chi_2)}
\end{equation}

\begin{remark}
	It is worth stressing that $\bm{T}$ becomes singular if $\chi_2=\pm\,\pi/2$.
\end{remark}

For $\chi_2\neq\pm\,\pi/2$, the angular velocity vector $\rep{\bm{\Omega}_{\star/\frakb}}{\star}$ is thus linked to the joint velocities $\dot{\bm{\theta}}$ by
\begin{equation}
	\rep{\bm{\Omega}_{\star/\frakb}}{\star} = \bm{T}\bm{J}\,\dot{\bm{\theta}}
\end{equation}

Such a relationship will be later used in the speed loop control of the SPM.

\subsection{Coaxiality of the input shafts}

The next proposition states an important kinematic property of all coaxial SPMs (CoSPM).

\begin{proposition}\label{prop:invariance}
	For any 3-DOF 3-\underline{R}RR CoSPM, each actuated joint $\theta_i$ ($i\in\itvd{1,3}$) making the same displacement $-\epsilon$ generates a pure bearing motion $\epsilon$.
\end{proposition}

\begin{proof}
	Computing the \emph{general} geometric model \eqref{eq:fgm_spm} of 3-DOF 3-\underline{R}RR SPMs with $\theta'_i\byaff\theta_i+\epsilon$, $i\in\itvd{1,3}$ yields
	\begin{equation*}\label{eq:coaxial_psi}
		\bm{f}\left(\theta_1+\epsilon,\theta_2+\epsilon,\theta_3+\epsilon,\chi_1,\chi_2,\chi_3\right) = \bm{0}_{3\times1}
	\end{equation*}
	By expanding all the terms in $\cos{\left(\theta_i+\epsilon\right)}$ and $\sin{\left(\theta_i+\epsilon\right)}$ using the classical trigonometric identities while considering coaxiality ($\beta_1=0$), one can show that the expanded expressions are factorizable by $\cos{\left(\chi_3+\epsilon\right)}$ or $\sin{\left(\chi_3+\epsilon\right)}$, leading to
	\begin{equation}\label{eq:coaxial_chi}
		\bm{f}\left(\theta_1+\epsilon,\theta_2+\epsilon,\theta_3+\epsilon,\bm{\chi}\right) = \bm{f}\left(\bm{\theta},\chi_1,\chi_2,\chi_3+\epsilon\right)
	\end{equation}
	These factorizations cannot be achieved if $\beta_1\neq0$, \textit{i.e.} if the mechanism is not coaxial. Equation \eqref{eq:coaxial_chi} clearly shows that an $\epsilon$-displacement in bearing $\chi_3$ requires to move all the actuators $\theta_i$ by a quantity $-\epsilon$. The sign ``$-$'' appears because the motion of the $i$\textsuperscript{th} actuator is described by an angle $\theta_i$ defined with the counterclockwise direction w.r.t.~$\rep{\bm{u}_i}{\frakb}$ (and thus, with the clockwise direction w.r.t.~$\bm{z}_\frakb$).
\end{proof}

This leads to the following statement.

\begin{lemma}\label{lem:invariance}
	Any 3-DOF 3-\underline{R}RR SPM with coaxial input shafts is geometrically invariant w.r.t.~bearing $\chi_3$.
\end{lemma}

One will take advantage of such a consideration for the singularity analysis of the mechanism.

\subsection{Singularity analysis}\label{ss:sing_analysis}

\subsubsection{Issues \& strategies}

Singularity loci are problematical configurations in which the robot does not behave properly, namely in terms of DOF. The main reasons are a loss of at least one controllable DOF (Type-1 singularity) or the gain of at least one uncontrollable DOF (Type-2 singularity). From the first order kinematic model viewpoint, Type-1 singularity occurs when $\bm{J}_2(\bm{\theta},\bm{\chi})$ is no longer invertible whereas Type-2 singularity occurs when $\bm{J}_1(\bm{\theta},\bm{\chi})$ becomes singular. As a result, numerical instabilities may arise in the neighborhood of such areas jeopardizing the integrity of the device. The aim of this subsection is to study these singularities in the work- and joint spaces of interest. In particular, one ensures that both spaces are singularity-free in a certified manner for our mechanism, since singularity loci are intrinsically subject to the design parameters of the system. Such a work is crucial to prevent any unstable and undesirable behavior of the robot. In our very case, such a study can be simplified by:
\begin{itemize}
	\item taking into account Lem.~\ref{lem:invariance}, \textit{i.e.} the invariance of the SPM's geometry and thus its singularities w.r.t.~bearing $\chi_3$;
	\item turning the non-linear system $\bm{f}(\bm{\theta},\bm{\chi})$ into a polynomial one $\bm{F}(\bm{\Theta},\bm{X})$.
\end{itemize}

It has been shown in \cite{Le23} that the polynomial system $\bm{F}(\bm{\Theta},\bm{X})$ of any SPM can be written as
\begin{equation}\label{eq:polynomial_gm}
	\begin{aligned}
		\bm{F}(\bm{\Theta},\bm{X}) &\bydef \mat{a_1(\bm{X})\,\Theta_1^2 + b_1(\bm{X})\,\Theta_1 + c_1(\bm{X})\\
			a_2(\bm{X})\,\Theta_2^2 + b_2(\bm{X})\,\Theta_2 + c_2(\bm{X})\\
			a_3(\bm{X})\,\Theta_3^2 + b_3(\bm{X})\,\Theta_3 + c_3(\bm{X})}\\
			&= \bm{0}_{3\times1}
	\end{aligned}
\end{equation}

where the vectors $\bm{X}\bydef\mat{X_1 & X_2 & X_3}^\sfT$ and $\bm{\Theta}\bydef\mat{\Theta_1 & \Theta_2 & \Theta_3}^\sfT$ are defined by
\begin{equation}
	X_i\bydef\tan{\left(\dfrac{\chi_i}{2}\right)},\;\Theta_i\bydef\tan{\left(\dfrac{\theta_i}{2}\right)}, \quad\forall\,i\in\itvd{1,3}
\end{equation}

The coefficients $a_i(\bm{X})$, $b_i(\bm{X})$, and $c_i(\bm{X})$ are not shown in this article but can be easily computed.

\begin{remark}
	The reader may notice that each polynomial $F_i(\Theta_i,\bm{X})$ of \eqref{eq:polynomial_gm} is quadratic w.r.t.~$\Theta_i$, $i\in\itvd{1,3}$. Using such a formalism is particularly suitable for the computation of the \emph{inverse geometric model}, \textit{i.e.} finding the joint vector $\bm{\theta}$ from a given orientation vector $\bm{\chi}$.
\end{remark}

\subsubsection{Type-1 singularities}

\begin{figure}[htbp]
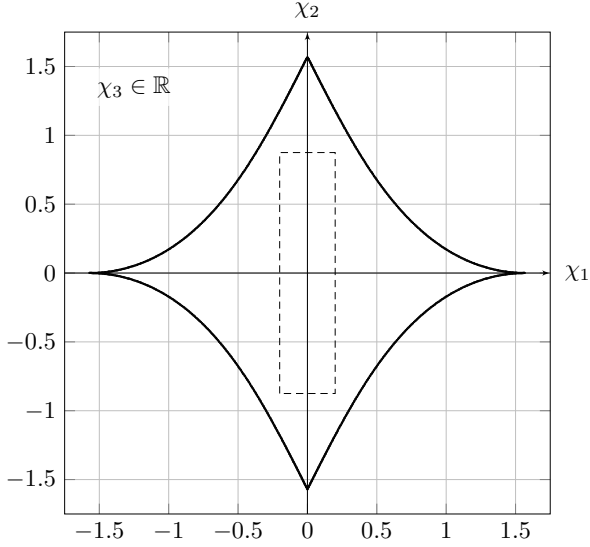

	\centering
	\includestandalone[width=8cm]{TikZ/spm_d1_st1}
	\caption{Prescribed workspace $\calW^\star$ (dashed) and Type-1 Singularity loci (solid) of the mechanism of interest in the $(\chi_1,\chi_2)$-plane}
	\label{fig:st1}
\end{figure}

As in \cite{Le23}, finding Type-1 singularity loci of any 3-DOF 3-\underline{R}RR SPM can be done by computing the \emph{critical points} belonging to the \emph{discriminant variety} \cite{LazRou07} of the inverse geometric model in its polynomial form, \textit{i.e.} $\calW_c(\bm{X})=\discrim{\left(\bm{F},\bm{X}\right)}$. One sets $X_3$ (or $\chi_3$) at an arbitrary real value (\textit{e.g.} $0$).
Figure \ref{fig:st1} shows the Type-1 singularity loci of the mechanism of interest in the $(\chi_1,\chi_2)$-plane and the prescribed workspace $\calW^\star$ defined earlier in Section~\ref{ss:pres}.

Physically speaking, Type-1 singularities are configurations where at least one leg of the SPM is folded or unfolded. Such a phenomenon never occurs in the prescribed workspace as $\calW^\star$ never meets $\calW_c(\bm{X})$.


\subsubsection{Type-2 singularities}\label{ss:spm_st2}

According to \cite{Mer06}, Type-2 singularities can be investigated with a similar problem called the \emph{path tracking} in orientation using the Kantorovich unicity operator \cite{Kan48}. Given the system $\bm{F}(\bm{\Theta},\bm{X})$, the goal is to ensure the uniqueness of the (orientation) solution $\bm{X}$ for the forward geometric model in a certain neighborhood of a known estimate $\bm{X}_0$, given a displacement in the (joint) parameter space $\bm{\Theta}$ and a certain computational precision. More precisely, the test guarantees that a classical Newton scheme will quadratically converge towards the desired solution with the initial known estimate. This strategy can be done iteratively by scanning $\calQ^\star$ being the image of the prescribed workspace $\calW^\star$ through the inverse geometric model. In our case, the path tracking in orientation was implemented using multiple-precision interval arithmetic \cite{Rou07}, taking into account all the possible uncertainties of the system, especially those on the design parameters of the SPM. A valid Kantorovich test for all values of $\calQ^\star$ means that the mechanism of interest is guaranteed to be far away from any numerical instabilities (including Type-2 singularities). Such a property is verified for this mechanism in its prescribed workspace. Hence, Type-2 singularities never occur in $\calW^\star$.

\section{Control law for the LOS stabilization}

\subsection{Principles \& Requirements}\label{ss:prin_req}

\begin{figure}[htbp]
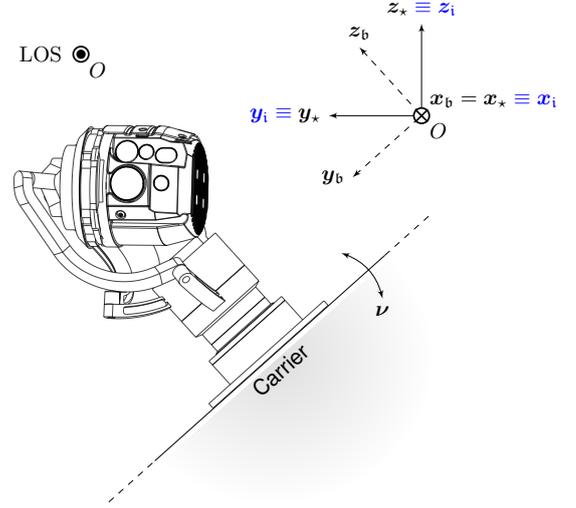

	\centering
	\includestandalone[width=\columnwidth]{TikZ/spm_stab_inertielle}
	\caption{Principle of inertial LOS stabilization (here w.r.t.~bank)}
	\label{fig:spm_stab_inertielle}
\end{figure}

As illustrated in Figure \ref{fig:spm_stab_inertielle}, the aim of inertial LOS stabilization is to maintain the direction of the LOS w.r.t.~the inertial frame $\calF_\fraki\bydef(O,\bm{x}_\fraki,\bm{y}_\fraki,\bm{z}_\fraki)$, despite disturbances. In our case study, one only focuses on the maritime environment where external disturbances are mostly waves. The latter make the carrier (and thus the base of the sight device) move w.r.t.~the inertial frame $\calF_\fraki$. In order to stabilize the LOS, the sight device needs to counteract the disturbances and therefore move w.r.t.~the carrier frame $\calF_\frakb$ where its base is attached, so that $\calF_\star\equiv\calF_\fraki$.

As our spherical parallel robot only provides rotational DOFs, the disturbance rejection is possible w.r.t.~the following angular motions:
\begin{itemize}
	\item \emph{roll} $\nu_1$ (angular disturbance w.r.t.~$x$-axis);
	\item \emph{pitch} $\nu_2$ (angular disturbance w.r.t.~$y$-axis);
	\item \emph{yaw} $\nu_3$ (angular disturbance w.r.t.~$z$-axis).
\end{itemize}
In the sequel, these angular disturbances are put into the vector $\bm{\nu}\bydef\mat{\nu_1 & \nu_2 & \nu_3}^\sfT$.

An appropriate way to evaluate the performance of the LOS stabilization is the consideration of the \emph{stabilization residual} vector $\bm{\varepsilon}$. The latter is defined as the difference between the desired position of the sight and its actual value. Thus, its components $\varepsilon_i$ should be as low as possible. A reasonable order of magnitude of $\varepsilon_i$ should not exceed $10^{-4}\,\text{rad}$. There are several strategies for LOS stabilization. In this article, we chose to implement a speed control loop.

\subsection{Speed control architecture}

\begin{figure*}[htbp]
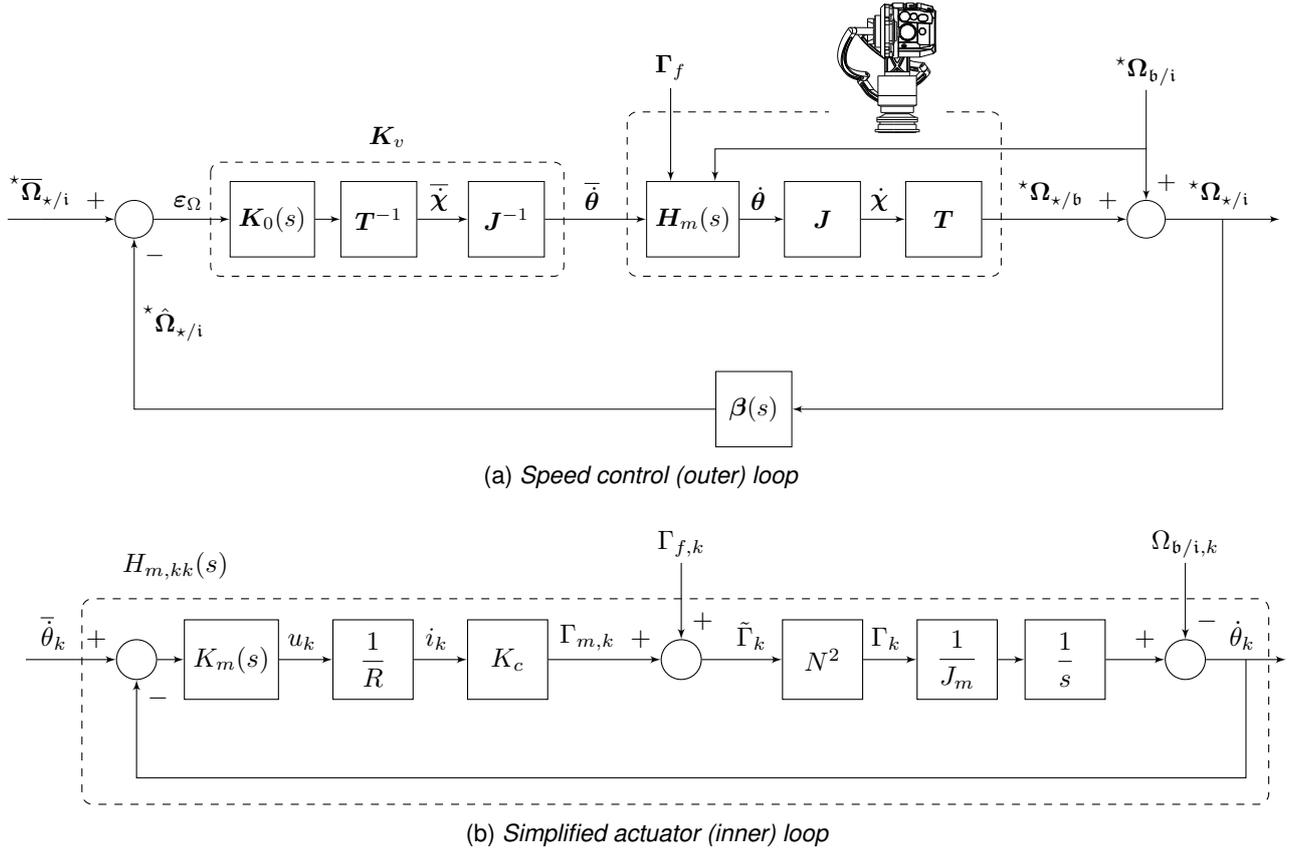

	\centering
	\begin{subfigure}{0.99\textwidth}
        \centering
        \includestandalone[width=\textwidth]{TikZ/spm_stab_inertielle_bd1}
		\caption{Speed control (outer) loop}
		\label{fig:spm_stab_inertielle_bd1}
    \end{subfigure}
    \hfill\vspace*{1em}
    \begin{subfigure}{0.99\textwidth}
        \centering
        \includestandalone[]{TikZ/spm_asv_moteur}
		\caption{Simplified actuator (inner) loop}
		\label{fig:spm_asv_moteur}
    \end{subfigure}
	\caption{Block diagrams considered for the inertial LOS stabilization}
	\label{fig:bd_gen}
\end{figure*}

Given Subsection~\ref{ss:prin_req}, using a speed control loop for LOS stabilization leads to set:
\begin{itemize}
	\item $\overline{\bm{\Omega}}_{\star/\fraki}$, the angular velocity of the sight w.r.t.~$\calF_\fraki$ as the reference signal;
	\item $\bm{\Omega}_{\frakb/\fraki}$, the angular velocity of the carrier w.r.t.~$\calF_\fraki$ as the external (non-measurable) disturbance signal.
\end{itemize}
Both values will be expressed in the LOS frame $\calF_\star$. It can then be shown that
\begin{equation}\label{eq:ext_dist_expr}
	\begin{aligned}
		\rep{\bm{\Omega}_{\frakb/\fraki}}{\star} &= \bm{R}_z(\chi_3)\bm{R}_y(\chi_2)\bm{R}_x(\chi_1)\,\rep{\bm{\Omega}_{\frakb/\fraki}}{\frakb}\\
		&= \bm{R}_z(\chi_3)\bm{R}_y(\chi_2)\bm{R}_x(\chi_1)\,\bm{T}'(\bm{\nu})\,\dot{\bm{\nu}}
	\end{aligned}
\end{equation}
where $\dot{\bm{\nu}}\bydef\mat{\dot{\nu}_1 & \dot{\nu}_2 & \dot{\nu}_3}^\sfT$ denotes the angular disturbances rate vector and 
\begin{equation}\label{eq:ext_dist_expr_detail}
	\bm{T}'(\bm{\nu}) = \mat{1 & 0 & -\sin(\nu_2)\\0 & \cos(\nu_1) & \sin(\nu_1)\cos(\nu_2)\\0 & -\sin(\nu_1) & \cos(\nu_1)\cos(\nu_2)}
\end{equation}
which is singular iff $\nu_2=\pm\,\pi/2$.

\subsubsection{Description}

Figure \ref{fig:spm_stab_inertielle_bd1} shows the block diagram of the speed control for the LOS stabilization. This control loop can be divided into three parts:
\begin{itemize}
	\item the system itself (also called \emph{plant}) which is modelled by its actuators $\bm{H}_m(s)$ generating the motions $\dot{\bm{\chi}}$ through $\dot{\bm{\theta}}$ and its kinematic Jacobian $\bm{J}$;
	\item the \emph{sensor} $\bm{\beta}(s)$ which is an Inertial Measurement Unit (IMU) and whose goal is to measure $\rep{\bm{\Omega}_{\star/\fraki}}{\star}$;
	\item the \emph{controller} part $\bm{K}_v$.
\end{itemize}

As the Jacobian matrices are defined through $\bm{\theta}$ and $\bm{\chi}$, one has at disposal a sensor measuring the current joint values $\hat{\bm{\theta}}$. The orientation vector $\bm{\chi}$ is then estimated through a Newton scheme solving $\hat{\bm{\chi}}=\bm{f}(\hat{\bm{\theta}})$ which does not bring instability given the analysis done in Subsection \ref{ss:spm_st2}.

The inertial angular velocity $\rep{\overline{\bm{\Omega}}_{\star/\fraki}}{\star}$ is given as the input reference signal. Its actual value is being estimated. The resulting velocity error estimation $\bm{\varepsilon}_\Omega$ defined by
\begin{equation}
	\bm{\varepsilon}_\Omega\bydef \rep{\overline{\bm{\Omega}}_{\star/\fraki}}{\star}-\rep{\hat{\bm{\Omega}}_{\star/\fraki}}{\star}
\end{equation}
is then given to the controller that computes the appropriate command $\overline{\dot{\bm{\theta}}}$ to the plant for the LOS stabilization despite the LOS disturbances $\rep{\bm{\Omega}_{\frakb/\fraki}}{\star}$. Each residual error signal $\varepsilon_i$, $i\in\itvd{1,3}$ can then be defined by
\begin{equation}
	\varepsilon_i = \int_{t_0=0}^t \varepsilon_{\Omega,i}(\tau) \dd\tau
\end{equation}
In the sequel, one considers the three identical actuators having a large reduction ratio $N$. As a result, the dynamics of the overall system can be considered as \emph{decoupled}. Moreover, the diagonal transfer function matrix $\bm{H}_m(s)$ must be regarded as a closed-loop transmittance matrix of the three actuators. The detailed block diagram of the $k\tss{th}$ diagonal component of this matrix is represented in Figure \ref{fig:spm_asv_moteur}. Such a model takes into account:
\begin{itemize}
	\item the \emph{electric resistance} $R$ [$\Omega$];
	\item the \emph{moment of inertia of the rotor} $J_m$ [$\text{kg}\cdot\text{m}^2$];
	\item the \emph{electromotive force} constant $K_c$ [$\text{Nm}/\text{A}$];
	\item the \emph{friction torque} $\Gamma_{f,k}$ [$\text{Nm}$] applied to the $k$\tss{th} actuator.
\end{itemize}

It is simplified in the sense that the electric inductance $L$ is neglected and the friction $f$ of the mechanical part is treated as an external step disturbance torque $\bm{\Gamma}_f \bydef \mat{\Gamma_{f,1} & \Gamma_{f,2} & \Gamma_{f,3}}^\sfT$.

In fact, each of its components $\Gamma_{f,k}$ acts as a \emph{Coulomb friction} torque defined as
\begin{equation}
	\Gamma_{f,k}\bydef f_k\sgn{\left(\dot{\theta}_k\right)}
\end{equation}
with $f_k$ being the friction coefficient applied to the $k$\tss{th} actuator.

Moreover, each actuator is controlled using a proportional controller $K_m(s)=K_m$ with a unitary feedback loop (assumed ideal). Under all these assumptions, $\bm{H}_m(s)=\diag\left(H_{m,11},H_{m,22},H_{m,33}\right)$ can be viewed as a diagonal matrix of first order transfer functions
\begin{equation}
	\bm{H}_m(s) = \dfrac{1}{1+\tau_m\,s}\id_3
\end{equation}
with $\tau_m=RJ_m/\left(N^2K_mK_c\right)$ being the \emph{time constant} of the actuators. In the sequel, one sets its value at $\tau_m=1.6\times10^{-3}\,\text{s}$. Note that the disturbance torque $\bm{\Gamma}_f$ can be treated as a step input disturbance by being brought at the upstream of $\bm{H}_m(s)$.

Furthermore, the sensor is modelled by a diagonal transfer matrix composed of pure delays $T_e$
\begin{equation}
	\bm{\beta}(s) = \e^{-T_es}\id_3
\end{equation}
where $T_e$ denotes the sampling period of the control loop. In our case $T_e=10^{-3}\,\text{s}$. Finally, the determination of $\bm{K}_v$ will be discussed in the next subsections.

\subsubsection{Synthesis of the controller}

As shown in Figure \ref{fig:spm_stab_inertielle_bd1}, the controller $\bm{K}_v$ is defined by
\begin{equation}
	\bm{K}_v = \bm{J}^{-1}\bm{T}^{-1}\,\bm{K}_0(s)
\end{equation}
where:
\begin{itemize}
	\item $\bm{J}^{-1}$ is the inverse kinematic Jacobian matrix;
	\item $\bm{T}^{-1}$ translating the Tait-Bryan angular rates $\dot{\bm{\chi}}$ into the velocity vector $\rep{\bm{\Omega}_{\star/\frakb}}{\star}$ through \eqref{eq:mapping_Omega_dchi};
	\item $\bm{K}_0(s)$ the linear part of the controller.
\end{itemize}	
Recall that the overall system is considered as decoupled. It can then be viewed as three independent and identical speed loops in $\rep{\dot{\Omega}_{\star/\fraki,j}}{\star}$, $j\in\itvd{1,3}$. As a result, the linear controller matrix $\bm{K}_0(s)$ is diagonal such that $\bm{K}_0(s)=K_0(s)\id_3$. In the sequel, $K_0(s)$ is defined as follows:
\begin{equation}
	K_0(s) = \overline{K_0}\,\dfrac{(s^2 + b_1s + b_2)\prod_{i=1}^3 (s+a_i)}{s^2 \prod_{i=1}^2 s^2 + c_is + d_i}
\end{equation}
with $\overline{K}_0 = 25884$, $a_1 = 4644$, $a_2 = 628.3$, $a_3 = 52.97$, $b_1 = 7356$, $b_2 = 2.584\times10^{7}$, $c_1 = 3.39\times10^{4}$, $d_1 = 2.943\times10^{8}$, $c_2 = 2899$, and $d_2 = 2.169\times10^{7}$.

\begin{figure}[htbp]
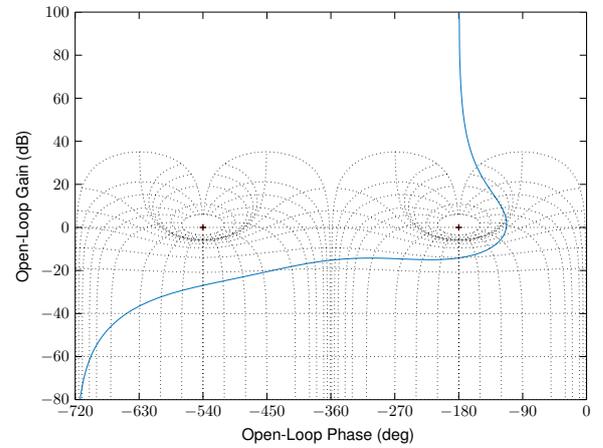

	\centering
	\includestandalone[width=\columnwidth]{TikZ/AsyCoSPM_1_open_loop_K0_black}
	\caption{Black-Nichols chart of the open-loop}
	\label{fig:black_K0BO_continu}
\end{figure}

\medskip
The open loop of each speed component $\rep{\dot{\Omega}_{\star/\fraki,j}}{\star}$, $j\in\itvd{1,3}$ is then defined by $K_0(s)H_m(s)\beta_{jj}(s)$. Figure \ref{fig:black_K0BO_continu} shows its Black-Nichols chart. The system has a phase margin of $\Delta\varphi=60^\circ$ and a gain margin of $\Delta G=14.2\,\text{dB}$. Such a controller will be discretized using the zero order hold method given $T_e$. 

Finally, the transfer between the speed error and the output disturbance for each loop $j$ denoted in the sequel as $D_j(s)$ is given by
\begin{equation}
	\begin{aligned}
		D_j(s)&\bydef\dfrac{\varepsilon_{\Omega,j}(s)}{\rep{\Omega_{\frakb/\fraki,j}}{\star}(s)} = \dfrac{\beta_{jj}(s)}{1+K_0(s)H_m(s)\beta_{jj}(s)}
	\end{aligned}
\end{equation}

\begin{figure}[htbp]
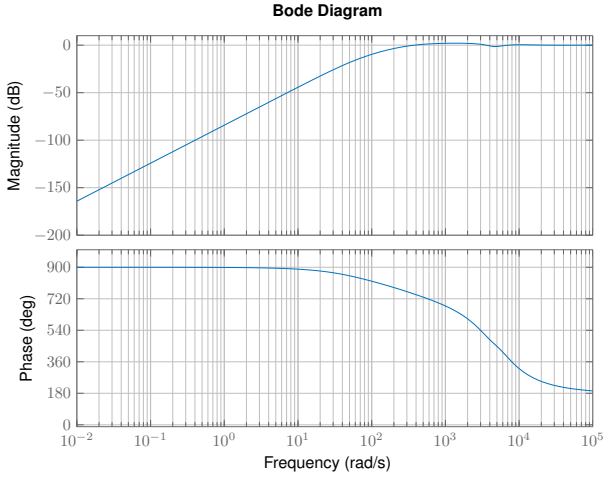

	\centering
	\includestandalone[width=\columnwidth]{TikZ/disturb_rej}
	\caption{Bode diagram of $D_j(s)$}
	\label{fig:bode_disturb_rej}
\end{figure}

Figure \ref{fig:bode_disturb_rej} displays the Bode diagram of $D_j(s)$.

\subsection{Simulations}

The following simulations are made using \textsc{Matlab} 2023b with Simulink\textsuperscript{\textregistered}.

\subsubsection{Inertial LOS stabilization}

\begin{figure}[htbp]
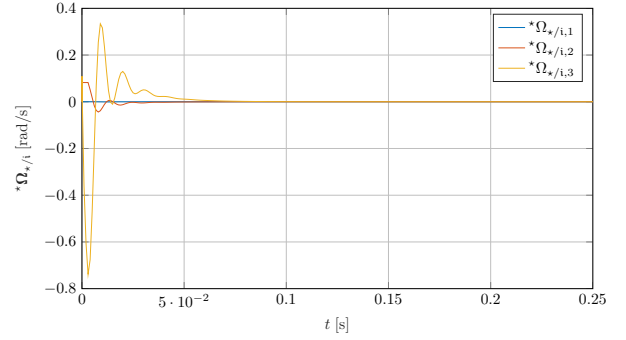
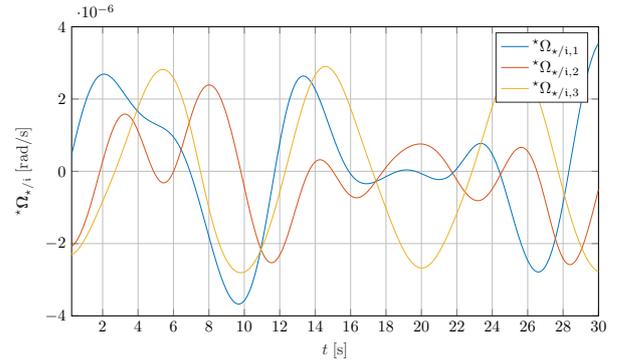

	\centering
	\begin{subfigure}{0.99\columnwidth}
		\centering
		\includestandalone[width=\columnwidth]{TikZ/AsyCoSPM_1_dot_CHI_reel_inertiel_trans}
		\caption{Transient state}
		\label{fig:s1_dot_chi_trans}
	\end{subfigure}
	\par\vspace*{0.75em}
	\begin{subfigure}{0.99\columnwidth}
		\centering
		\includestandalone[width=\columnwidth]{TikZ/AsyCoSPM_1_dot_CHI_reel_inertiel_perm}
		\caption{Steady state}
		\label{fig:s1_dot_chi_perm}
	\end{subfigure}
	\caption{Inertial velocity $\rep{\bm{\Omega}_{\star/\fraki}}{\star}$ of the sight device}
	\label{fig:AsyCoSPM_s1_dot_chi}
\end{figure}

\begin{figure}[htbp]
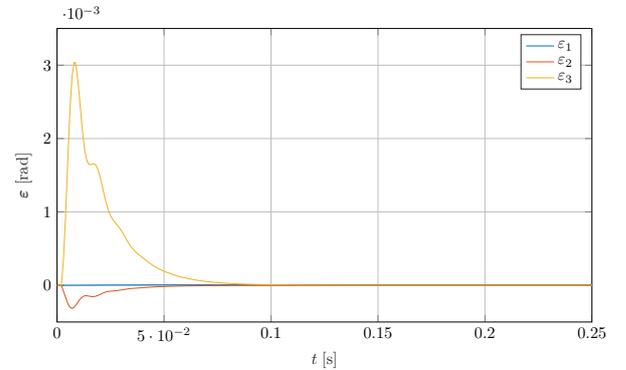
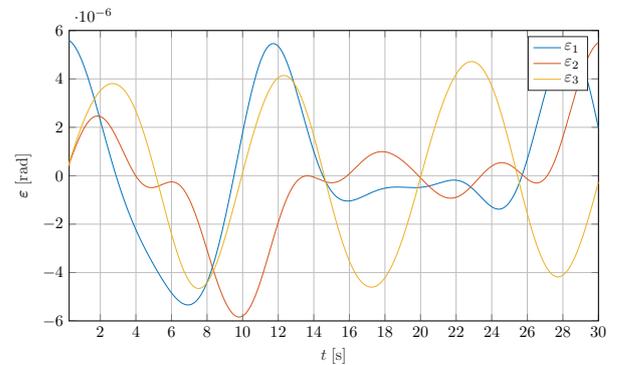

	\centering
	\begin{subfigure}{0.99\columnwidth}
		\centering
		\includestandalone[width=\columnwidth]{TikZ/AsyCoSPM_1_CHI_reel_inertiel_trans}
		\caption{Transient state}
		\label{fig:s1_chi_trans}
	\end{subfigure}
	\par\vspace*{0.75em}
	\begin{subfigure}{0.99\columnwidth}
		\centering
		\includestandalone[width=\columnwidth]{TikZ/AsyCoSPM_1_CHI_reel_inertiel_perm}
		\caption{Steady state}
		\label{fig:s1_chi_perm}
	\end{subfigure}
	\caption{Stabilization residual error $\bm{\varepsilon}$}
	\label{fig:AsyCoSPM_s1_chi}
\end{figure}

\begin{figure}[htbp]
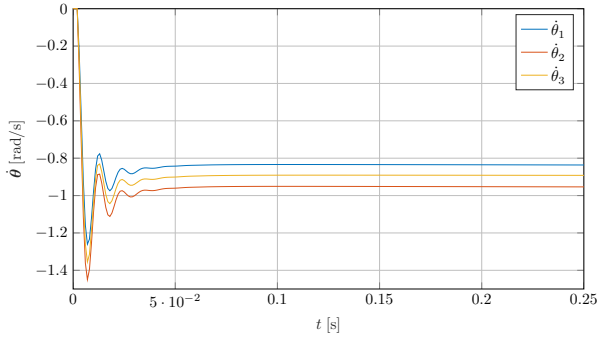
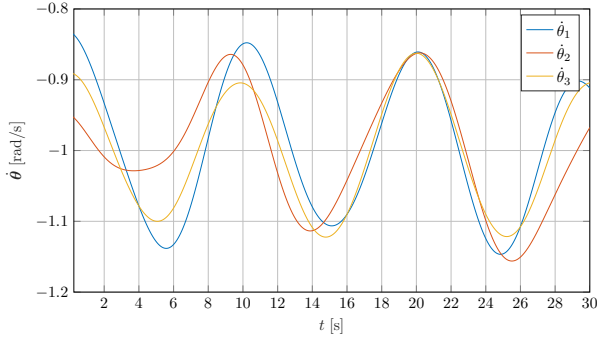

	\centering
	\begin{subfigure}{0.99\columnwidth}
		\centering
		\includestandalone[width=\columnwidth]{TikZ/AsyCoSPM_1_dot_THETA_trans}
		\caption{Transient state}
		\label{fig:s1_dot_theta_trans}
	\end{subfigure}
	\par\vspace*{0.75em}
	\begin{subfigure}{0.99\columnwidth}
		\centering
		\includestandalone[width=\columnwidth]{TikZ/AsyCoSPM_1_dot_THETA_perm}
		\caption{Steady state}
		\label{fig:s1_dot_theta_perm}
	\end{subfigure}
	\caption{Joint velocities}
	\label{fig:AsyCoSPM_s1_dot_theta}
\end{figure}

\begin{figure}[htbp]
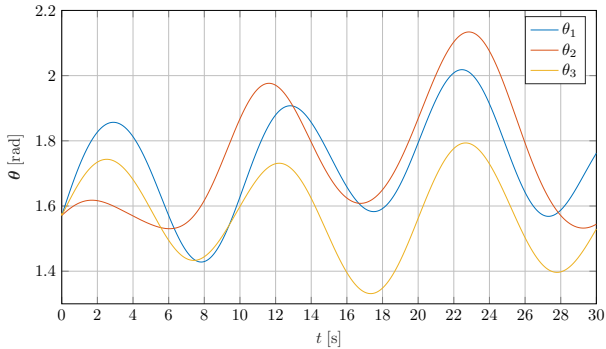

	\centering
	\includestandalone[width=\columnwidth]{TikZ/AsyCoSPM_1_theta_s1}
	\caption{Associated joint values}
	\label{fig:AsyCoSPM_s1_theta}
\end{figure}

In this subsection, the goal is to maintain the LOS at its home configuration despite the motion of the carrier subject to waves. For this purpose, the reference signal is defined by $\rep{\bm{\overline{\Omega}}_{\star/\fraki}}{\star}=\bm{0}$. One assumes that the angular disturbances are expressed through
\begin{equation}
	\forall\,i\in\itvd{1,3},\quad \nu_i(t) = \overline{\nu_i}\cos\left(2\pi\,f_i\right)
\end{equation}
We set the following parameters as follows:
\begin{itemize}
	\item $\overline{\nu_1}=\overline{\nu_2}=10^\circ$ and $\overline{\nu_3}=0$;
	\item $f_1=0.1\,\text{Hz}$ and $f_2=0.075\,\text{Hz}$; 
	\item input disturbance treated as a step of magnitude $1$.
\end{itemize}
The non-measurable external disturbance velocity $\rep{\bm{\Omega}_{\frakb/\fraki}}{\star}$ can be deduced from \eqref{eq:ext_dist_expr}, \eqref{eq:ext_dist_expr_detail} and the orientation vector $\bm{\chi}$ of the platform. The parallel robot is initially set in its home configuration, \textit{i.e.} $\bm{\chi}=\bm{0}$ and $\bm{\theta}=\frac{\pi}{2}\,\bm{1}_{3\times1}$. Given these conditions, a simulation spanning 30 seconds gives the following results:
\begin{itemize}
	\item Figure \ref{fig:AsyCoSPM_s1_dot_chi} shows the time response of $\rep{\bm{\Omega}_{\star/\fraki}}{\star}$ being the angular velocity of the sight in the transient (Fig.~\ref{fig:s1_dot_chi_trans}) and steady (Fig.~\ref{fig:s1_dot_chi_perm}) states.
	\item Figure \ref{fig:AsyCoSPM_s1_chi} plots the stabilization residual error $\bm{\varepsilon}$ of the sight device in the transient (Fig.~\ref{fig:s1_chi_trans}) and steady (Fig.~\ref{fig:s1_chi_perm}) states.
	\item Figure \ref{fig:AsyCoSPM_s1_dot_theta} depicts the joint values $\dot{\bm{\theta}}$ used as the input signal vector of the robot modelled by its Jacobian $\bm{J}$ (Fig.~\ref{fig:s1_dot_theta_trans}) and steady (Fig.~\ref{fig:s1_dot_theta_perm}) states.
	\item Figure \ref{fig:AsyCoSPM_s1_theta} depicts the joint values $\bm{\theta}$ used to counteract the disturbances and maintain the LOS.
\end{itemize}

On the one hand, one can see from Fig.~\ref{fig:AsyCoSPM_s1_dot_chi} and \ref{fig:AsyCoSPM_s1_chi} that the input disturbance torque is entirely rejected from the beginning of the simulation. This results also make sense regarding the Bode diagram of $D_j(s)$ (Figure \ref{fig:bode_disturb_rej}). For instance, one focuses on the first speed component $\rep{\Omega_{\star/\fraki,1}}{\star}$ (blue curve of Fig.~\ref{fig:AsyCoSPM_s1_dot_chi}). According to the Bode diagram of $D_j(s)$ at $\omega_1=2\pi f_1\simeq0.62\,\text{rad/s}$, one has $\abs{D_1(\im\omega_1)}\simeq-90\,\text{dB}$. This means that a unitary sine disturbance input $\rep{\Omega_{\frakb/\fraki,1}}{\star}$ with the same frequency $\omega_1$ will be attenuated by a factor $10^{5}$. Given the magnitude of $\rep{\Omega_{\frakb/\fraki}}{\star}$, this speed error in the steady state has an order of magnitude of $10^{-6}$, as confirmed by Fig.~\ref{fig:s1_dot_chi_perm}. Moreover, the magnitude of the residual stabilization error does not exceed $6\,\text{µrad}$. Such results are in accordance with the requirements explained in Subsection \ref{ss:prin_req}.

On the other hand, the motion of the actuators required to fulfill the inertial LOS stabilization is reasonable in the sense that the joint velocities (Figure \ref{fig:AsyCoSPM_s1_dot_theta}) do not have excessive values.

\subsubsection{Implementation using a digital model}

\begin{figure}[htbp]
	\centering
	\includegraphics[width=\columnwidth]{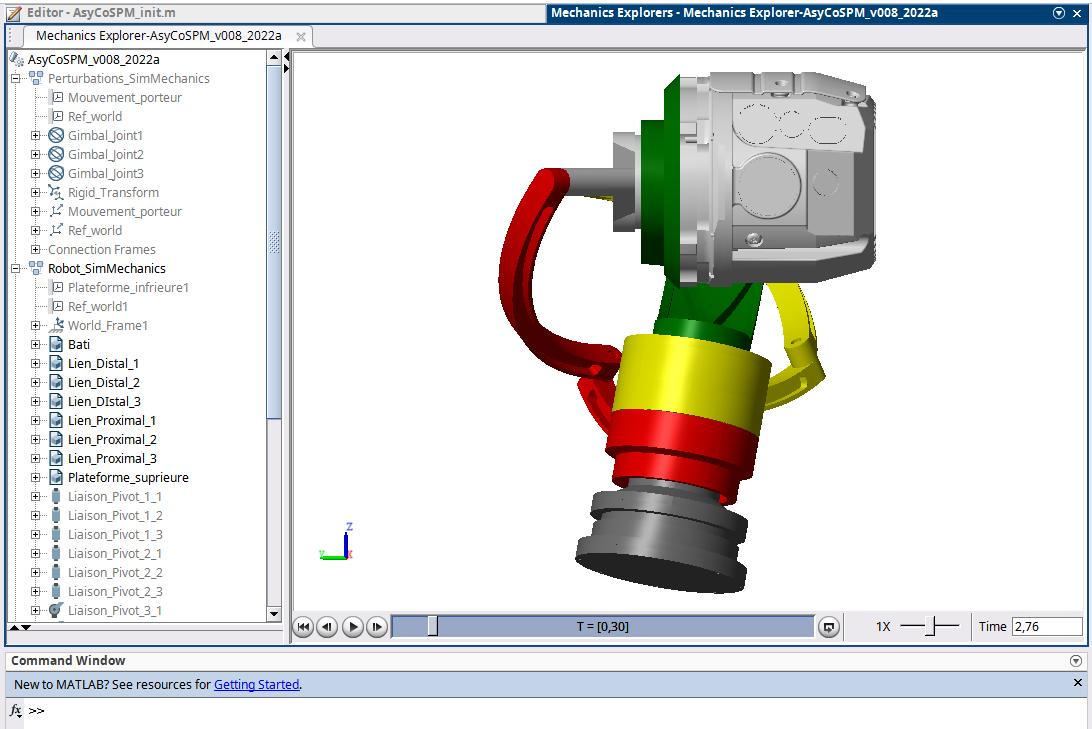}
	\caption{Digital model of the SPM}
	\label{fig:AsyCoSPM_SimMechanics}
\end{figure}

In this last simulation, we implemented a digital model of the spherical parallel robot and its speed control in a \textsc{Matlab} Simulink\textsuperscript{\textregistered} environment (here 2022a). The latter is interfaced with the \textsc{Catia} files of the mechanism via SimMechanics\texttrademark{}. Figure \ref{fig:AsyCoSPM_SimMechanics} shows the screenshot of the before-mentioned digital model.

\begin{figure}[htbp]
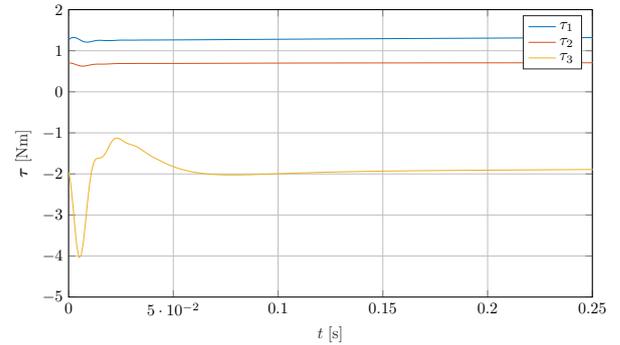
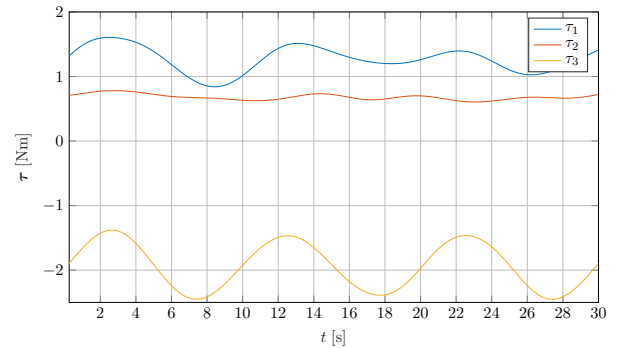

	\centering
	\begin{subfigure}{0.99\columnwidth}
		\centering
		\includestandalone[width=\columnwidth]{TikZ/AsyCoSPM_1_TAU_trans}
		\caption{Transient state}
		\label{fig:s1_tau_trans}
	\end{subfigure}
	\par\vspace*{0.75em}
	\begin{subfigure}{0.99\columnwidth}
		\centering
		\includestandalone[width=\columnwidth]{TikZ/AsyCoSPM_1_TAU_perm}
		\caption{Steady state}
		\label{fig:s1_tau_perm}
	\end{subfigure}
	\caption{Motor torque of the SPM}
	\label{fig:AsyCoSPM_s1_tau}
\end{figure}

Such a model allows us to have an overview of the dynamics of the robot. It is in this regard complementary to the kinematic study of the mechanism discussed in this article. For instance, one can evaluate the torque $\bm{\tau}\bydef\mat{\tau_1 & \tau_2 & \tau_3}^\sfT$ required to move the several actuators, as shown in Figure \ref{fig:AsyCoSPM_s1_tau}, where $\tau_i$ denotes the torque applied to the $i$\tss{th} actuator.

These curves show that the engine torques required to stabilize the LOS are reasonable: one obtains torques in absolute values that do not exceed $4\,\text{Nm}$ in the transient state and $2.5\,\text{Nm}$ in steady state. 

\section{Conclusion \& outlook}

This paper discussed on the use of a 3-DOF Spherical Parallel Manipulator with coaxial shafts that embeds a sight device for the inertial LOS stabilization. Having at disposal its kinematic model, it has been shown that a speed control loop achieves the disturbance rejection taking into account the motion of the carrier as well as the friction torque undergone by the actuators. Such results are also verified with the SimMechanics\texttrademark{} digital model of our robot simulating its dynamics.

Although the kinematics of the robot is certified for our regular workspace, the control strategy is not. Indeed, the inertial speed reference can in fact make the mechanism leave the safe region without knowing it. In this case, one cannot guaranty the good behavior of the device and its control. By the same logic, the current control loop does not take into account uncertainties on the system that can bring instability in the worst case. In this regard, further works will focus on the robustness of the system and a management of the joint stops to avoid the robot leaving the safe regions. Such aspects will bring the certification to the speed control.

As the article mainly focused on the kinematics of the robot, an interesting outlook could be diving into the dynamic model of the latter.

\begin{figure*}[ht!]
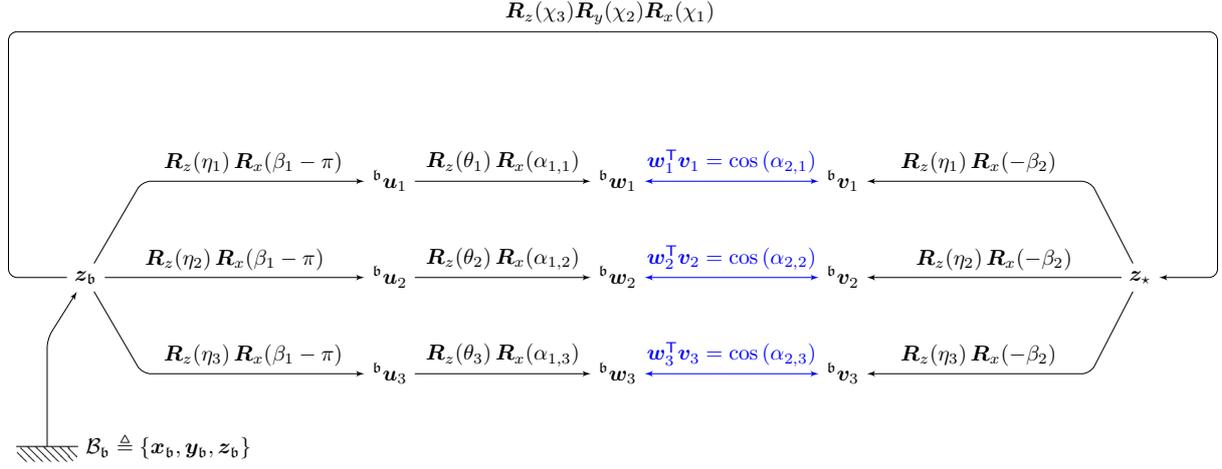

	\centering
	\includestandalone[width=2\columnwidth]{TikZ/spm_chaine_cinematique}
	\caption{Kinematic chain of a general SPM}
	\label{fig:spm_chaine_cinematique}
\end{figure*}

\section*{Acknowledgment}

The authors of this article want to thank Arnaud \textsc{Quadrat} for his careful review and relevant suggestions.

%% file: TikZ/spm_d1_legend.tex
	\def\ScaleFactor{1}

	\begin{tikzpicture}[>=latex',every text node part/.style={align=center}]
		
		\node[] at (0,0) (im) {\centering\includegraphics[scale=0.25]{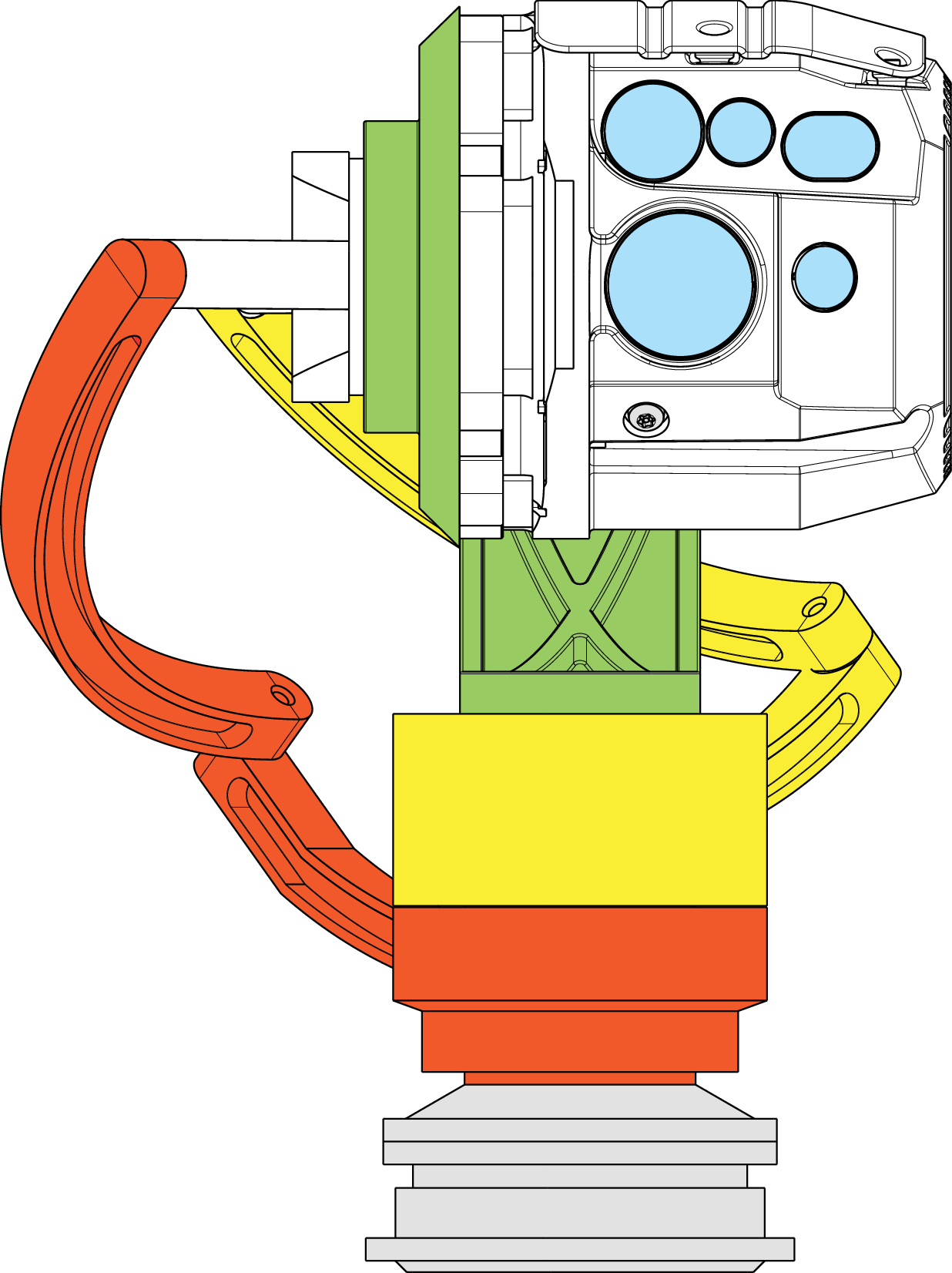}};
		
		
		\draw[<-,blue] (-1*\ScaleFactor,-1.25*\ScaleFactor)to[in=90,out=180](-2*\ScaleFactor,-2.5*\ScaleFactor)node[below]{a proximal link};
		\draw[->,blue] (-1.75*\ScaleFactor,3.5*\ScaleFactor)node[above]{a distal link}to[in=105,out=-135](-2.15*\ScaleFactor,1.65*\ScaleFactor);
		
		\node[rotate=90,blue] at (2*\ScaleFactor,-2.25*\ScaleFactor) {$\underbrace{\phantom{actuator actuar}}$};
		\node[blue] at (3.15*\ScaleFactor,-2.25*\ScaleFactor) {actuators};
		
		\draw[->,blue] (3.5*\ScaleFactor,0*\ScaleFactor)node[below]{a sight\\device}to[in=-85,out=90](2.25*\ScaleFactor,0.75*\ScaleFactor);
	\end{tikzpicture}

%% file: TikZ/rotation_order_zyx.tex
	\def\ScaleFactor{1}

	\begin{tikzpicture}[>=latex',every text node part/.style={align=center}]
		\draw[<->,black!50] (0*\ScaleFactor,2*\ScaleFactor)node[above]{$\bm{y}_0$}|-(2*\ScaleFactor,0*\ScaleFactor)node[right]{$\bm{x}_0$};
		\begin{scope}[rotate=30]
			\draw[<->] (0*\ScaleFactor,2*\ScaleFactor)node[above]{$\bm{y}_1$}|-(2*\ScaleFactor,0*\ScaleFactor)node[right]{$\bm{x}_1$};
		\end{scope}
		\draw[thick] (-0.5*\ScaleFactor,0*\ScaleFactor)node[below=0.5em]{${\color{black!50}\bm{z}_0}=\bm{z}_1$} circle (0.125*\ScaleFactor);
		\draw[draw=none,fill=black] (-0.5*\ScaleFactor,0*\ScaleFactor) circle (0.075*\ScaleFactor);
		
		\draw[->] (0*\ScaleFactor,0*\ScaleFactor)++(1*\ScaleFactor,0*\ScaleFactor) arc [start angle=0,end angle=30,x radius=1*\ScaleFactor,y radius=1*\ScaleFactor]node[right,midway]{$\chi_3$};
		\draw[->] (0*\ScaleFactor,0*\ScaleFactor)++(0*\ScaleFactor,1*\ScaleFactor) arc [start angle=90,end angle=120,x radius=1*\ScaleFactor,y radius=1*\ScaleFactor]node[above,midway]{$\chi_3$};
		
		\begin{scope}[xshift=5*\ScaleFactor cm]
			\draw[<->,black!50] (0*\ScaleFactor,2*\ScaleFactor)node[above]{$\bm{x}_1$}|-(2*\ScaleFactor,0*\ScaleFactor)node[right]{$\bm{z}_1$};
			\begin{scope}[rotate=30]
				\draw[<->] (0*\ScaleFactor,2*\ScaleFactor)node[above]{$\bm{x}_2$}|-(2*\ScaleFactor,0*\ScaleFactor)node[right]{$\bm{z}_2$};
			\end{scope}
			\draw[thick] (-0.5*\ScaleFactor,0*\ScaleFactor)node[below=0.5em]{${\color{black!50}\bm{y}_1}=\bm{y}_2$} circle (0.125*\ScaleFactor);
			\draw[draw=none,fill=black] (-0.5*\ScaleFactor,0*\ScaleFactor) circle (0.075*\ScaleFactor);
			
			\draw[->] (0*\ScaleFactor,0*\ScaleFactor)++(1*\ScaleFactor,0*\ScaleFactor) arc [start angle=0,end angle=30,x radius=1*\ScaleFactor,y radius=1*\ScaleFactor]node[right,midway]{$\chi_2$};
			\draw[->] (0*\ScaleFactor,0*\ScaleFactor)++(0*\ScaleFactor,1*\ScaleFactor) arc [start angle=90,end angle=120,x radius=1*\ScaleFactor,y radius=1*\ScaleFactor]node[above,midway]{$\chi_2$};
		\end{scope}
		
		\begin{scope}[xshift=10*\ScaleFactor cm]
			\draw[<->,black!50] (0*\ScaleFactor,2*\ScaleFactor)node[above]{$\bm{z}_2$}|-(2*\ScaleFactor,0*\ScaleFactor)node[right]{$\bm{y}_2$};
			\begin{scope}[rotate=30]
				\draw[<->] (0*\ScaleFactor,2*\ScaleFactor)node[above]{$\bm{z}_\star$}|-(2*\ScaleFactor,0*\ScaleFactor)node[right]{$\bm{y}_\star$};
			\end{scope}
			\draw[thick] (-0.5*\ScaleFactor,0*\ScaleFactor)node[below=0.5em]{${\color{black!50}\bm{x}_2}=\bm{x}_\star$} circle (0.125*\ScaleFactor);
			\draw[draw=none,fill=black] (-0.5*\ScaleFactor,0*\ScaleFactor) circle (0.075*\ScaleFactor);
			
			\draw[->] (0*\ScaleFactor,0*\ScaleFactor)++(1*\ScaleFactor,0*\ScaleFactor) arc [start angle=0,end angle=30,x radius=1*\ScaleFactor,y radius=1*\ScaleFactor]node[right,midway]{$\chi_1$};
			\draw[->] (0*\ScaleFactor,0*\ScaleFactor)++(0*\ScaleFactor,1*\ScaleFactor) arc [start angle=90,end angle=120,x radius=1*\ScaleFactor,y radius=1*\ScaleFactor]node[above,midway]{$\chi_1$};
		\end{scope}
	\end{tikzpicture}

%% file: TikZ/spm_d1_frames.tex
	\def\ScaleFactor{1}

	\begin{tikzpicture}[>=latex',every text node part/.style={align=center}]
		
		\node[opacity=0.15] at (0,0) (im) {\centering\includegraphics[scale=0.25]{pictures/SPM_D1/color_vue_devant}};

		\draw[black!15] (-2*\ScaleFactor,-3.5*\ScaleFactor)--(3*\ScaleFactor,-3.5*\ScaleFactor);
		\draw[draw=none,pattern=north west lines, pattern color=black!20] (-2*\ScaleFactor,-3.5*\ScaleFactor) rectangle (3*\ScaleFactor,-4*\ScaleFactor);
		
		
		\draw[->] ($(im.center)+(0.57*\ScaleFactor,-3.5*\ScaleFactor)$)node[below,right=0.25em]{$O'$}--($(im.center)+(0.57*\ScaleFactor,-2.5*\ScaleFactor)$)node[above]{$\bm{z}_\frakb$};
		\draw[->] ($(im.center)+(0.57*\ScaleFactor,-3.5*\ScaleFactor)$)--($(im.center)+(-0.57*\ScaleFactor,-3.5*\ScaleFactor)$)node[above]{$\bm{y}_\frakb$};

		\draw[thick] (1.25*\ScaleFactor,-3*\ScaleFactor)node[above=0.5em]{$\bm{x}_\frakb$} circle (0.125*\ScaleFactor);
		\draw[thick] (1.15*\ScaleFactor,-3.1*\ScaleFactor)--(1.35*\ScaleFactor,-2.9*\ScaleFactor);
		\draw[thick] (1.15*\ScaleFactor,-2.9*\ScaleFactor)--(1.35*\ScaleFactor,-3.1*\ScaleFactor);

		\draw[<->] ($(im.center)+(0.57*\ScaleFactor,2.975*\ScaleFactor)$)node[above]{$\bm{z}_\star$}|-node[below,right=0.25em]{$O$}($(im.center)+(-0.57*\ScaleFactor,1.975*\ScaleFactor)$)node[above]{$\bm{y}_\star$};

		\draw[thick] (-2.5*\ScaleFactor,2.5*\ScaleFactor)node[above=0.5em]{$\bm{x}_\star$} circle (0.125*\ScaleFactor);
		\draw[thick] (-2.6*\ScaleFactor,2.4*\ScaleFactor)--(-2.4*\ScaleFactor,2.6*\ScaleFactor);
		\draw[thick] (-2.6*\ScaleFactor,2.6*\ScaleFactor)--(-2.4*\ScaleFactor,2.4*\ScaleFactor);



		\draw[<->] (5*\ScaleFactor,3.5*\ScaleFactor)node[above]{$\bm{z}_\star$}|-node[midway,anchor=south east]{$O$}(4*\ScaleFactor,2.5*\ScaleFactor)node[left]{$\bm{y}_\star$};

		\draw[thick,fill=white] (5*\ScaleFactor,2.5*\ScaleFactor)node[below=0.5em]{$\bm{x}_\star=-\mathrm{LOS}$} circle (0.125*\ScaleFactor);
		\draw[thick] (4.9*\ScaleFactor,2.4*\ScaleFactor)--(5.1*\ScaleFactor,2.6*\ScaleFactor);
		\draw[thick] (4.9*\ScaleFactor,2.6*\ScaleFactor)--(5.1*\ScaleFactor,2.4*\ScaleFactor);


		\draw[dashed,rounded corners] (-1.15*\ScaleFactor,-0.25*\ScaleFactor) rectangle (2.25*\ScaleFactor,-3.75*\ScaleFactor);

		\draw[dashed,rounded corners] (4.5*\ScaleFactor,1.25*\ScaleFactor) rectangle (6.5*\ScaleFactor,-2.75*\ScaleFactor);

		\draw[draw=none,pattern=north west lines, pattern color=black!80] (5*\ScaleFactor,-2.25*\ScaleFactor) rectangle (6*\ScaleFactor,-2*\ScaleFactor);
		\draw[] (5*\ScaleFactor,-2*\ScaleFactor)--(6*\ScaleFactor,-2*\ScaleFactor);
		\draw[thick] (5.5*\ScaleFactor,-2*\ScaleFactor)--(5.5*\ScaleFactor,0.5*\ScaleFactor);

		\draw[red,fill=white] (5.35*\ScaleFactor,-1.75*\ScaleFactor) rectangle (5.65*\ScaleFactor,-1.15*\ScaleFactor);
		\draw[<-,thick] (5.15*\ScaleFactor,-1.6*\ScaleFactor)to[in=200,out=-20](5.85*\ScaleFactor,-1.6*\ScaleFactor)node[right]{$\theta_1$};
		\begin{scope}[yshift=0.75*\ScaleFactor cm]
			\draw[yellow!50!orange,fill=white] (5.35*\ScaleFactor,-1.75*\ScaleFactor) rectangle (5.65*\ScaleFactor,-1.15*\ScaleFactor);
			\draw[<-,thick] (5.15*\ScaleFactor,-1.6*\ScaleFactor)to[in=200,out=-20](5.85*\ScaleFactor,-1.6*\ScaleFactor)node[right]{$\theta_2$};
		\end{scope}
		\begin{scope}[yshift=1.5*\ScaleFactor cm]
			\draw[dgreen,fill=white] (5.35*\ScaleFactor,-1.75*\ScaleFactor) rectangle (5.65*\ScaleFactor,-1.15*\ScaleFactor);
			\draw[<-,thick] (5.15*\ScaleFactor,-1.6*\ScaleFactor)to[in=200,out=-20](5.85*\ScaleFactor,-1.6*\ScaleFactor)node[right]{$\theta_3$};
		\end{scope}

		\draw[red] (5.35*\ScaleFactor,-1.45*\ScaleFactor)--(4.5*\ScaleFactor,-0.75*\ScaleFactor);
		\draw[yellow!50!orange] (5.65*\ScaleFactor,-0.7*\ScaleFactor)--(6.5*\ScaleFactor,-0.25*\ScaleFactor);
		\draw[dgreen] (5.35*\ScaleFactor,0.05*\ScaleFactor)--++(-0.1*\ScaleFactor,0.1*\ScaleFactor)--(5.25*\ScaleFactor,1.25*\ScaleFactor);

		\draw[very thick] (2.25*\ScaleFactor,-2*\ScaleFactor)to[in=180,out=0](4.5*\ScaleFactor,-0.7*\ScaleFactor);
	\end{tikzpicture}

%% file: TikZ/spm_d1_beta_vec.tex
	\def\ScaleFactor{1}

	\begin{tikzpicture}[>=latex']
		
		\node[opacity=0.2] at (0,0) (im) {\centering\includegraphics[scale=0.25]{pictures/SPM_D1/color_vue_devant}};
		
		
		\draw[<-] ($(im.center)+(0.57*\ScaleFactor,3.935*\ScaleFactor)$)node[above]{$\bm{z}_0=\bm{z}_\star$}--++(0,-1.975*\ScaleFactor)node[at end,right=0.25em]{$O$};
		\draw[thick] (-2.5*\ScaleFactor,3.5*\ScaleFactor)node[above=0.5em]{$\bm{x}_0=\bm{x}_\star$} circle (0.125*\ScaleFactor);
		\draw[thick] (-2.6*\ScaleFactor,3.4*\ScaleFactor)--(-2.4*\ScaleFactor,3.6*\ScaleFactor);
		\draw[thick] (-2.6*\ScaleFactor,3.6*\ScaleFactor)--(-2.4*\ScaleFactor,3.4*\ScaleFactor);
		
		\draw[densely dashed] ($(im.center)+(0.57*\ScaleFactor,1.975*\ScaleFactor)$)--($(im.center)+(-1.8*\ScaleFactor,1.975*\ScaleFactor)$);
		\draw[thick,blue,->] ($(im.center)+(-1.8*\ScaleFactor,1.975*\ScaleFactor)$)--($(im.center)+(-3*\ScaleFactor,1.975*\ScaleFactor)$)node[midway,above]{$\bm{v}_i$};
		
		\draw[densely dashed] ($(im.center)+(0.57\ScaleFactor,1.975*\ScaleFactor)$)--($(im.center)+(0.57*\ScaleFactor,-3.5*\ScaleFactor)$);
		\draw[thick,blue,->] ($(im.center)+(0.57*\ScaleFactor,-3.5*\ScaleFactor)$)--($(im.center)+(0.57*\ScaleFactor,-4.5*\ScaleFactor)$)node[midway,right]{$\bm{u}_i$};
		
		\draw[densely dashed] ($(im.center)+(0.57*\ScaleFactor,1.975*\ScaleFactor)$)--($(im.center)+(-1.1*\ScaleFactor,-0.35*\ScaleFactor)$);
		\draw[thick,blue,->] ($(im.center)+(-1.1*\ScaleFactor,-0.35*\ScaleFactor)$)--($(im.center)+(-1.75*\ScaleFactor,-1.254940120*\ScaleFactor)$)node[midway,left]{$\bm{w}_i$};
		
		\draw[->,thick] ($(im.center)+(0.57\ScaleFactor,2.975*\ScaleFactor)$)to[in=90,out=180]($(im.center)+(-0.43*\ScaleFactor,1.975*\ScaleFactor)$);
		\node at (-.5*\ScaleFactor,3*\ScaleFactor) {$\beta_2$};
		\node at (1.5*\ScaleFactor,-3*\ScaleFactor) {$\beta_1=0$};
	\end{tikzpicture}

%% file: TikZ/spm_d1_jambe1.tex
	\def\ScaleFactor{1}

	\begin{tikzpicture}[>=latex']
		
		\node[] at (0,0) (im) {\centering\includegraphics[scale=0.14]{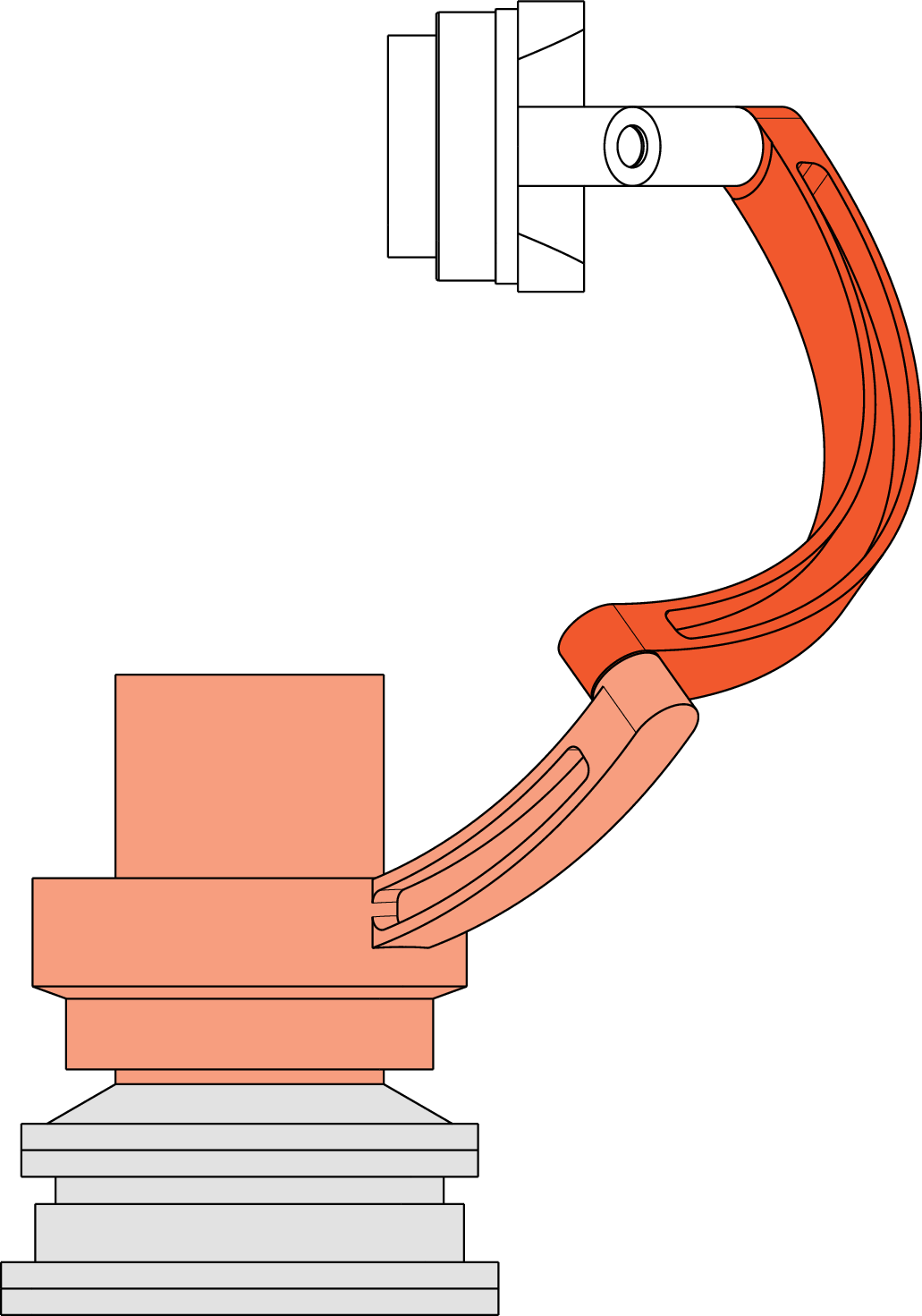}};
		
		
		\draw[dashed] ($(im.center)+(-0.25*\ScaleFactor,-0.75*\ScaleFactor)$)--($(im.center)+(-0.25*\ScaleFactor,-1.1*\ScaleFactor)$)node[near end](a){};
		\draw[dashed] ($(im.center)+(0.55*\ScaleFactor,-0.2*\ScaleFactor)$)--($(im.center)+(0.725*\ScaleFactor,-0.45*\ScaleFactor)$)node[near end](b){};
		\draw[dashed] ($(im.center)+(0.8*\ScaleFactor,1.35*\ScaleFactor)$)--($(im.center)+(1.5*\ScaleFactor,1.35*\ScaleFactor)$)node[near end](c){};
		
		\draw[<->,thick] ($(im.center)+(-0.25*\ScaleFactor,-1.1*\ScaleFactor)$)to[in=-135,out=20]($(im.center)+(0.725*\ScaleFactor,-0.45*\ScaleFactor)$)node[right,below=1em]{$\alpha_{1,1}$};
		\draw[<->,thick] ($(im.center)+(0.675*\ScaleFactor,-0.4*\ScaleFactor)$)to[in=-50,out=10]($(im.center)+(1.25*\ScaleFactor,1.35*\ScaleFactor)$);
		\node[anchor=center] at ($(im.center)+(1.75*\ScaleFactor,0*\ScaleFactor)$){$\alpha_{2,1}$};

		\begin{pgfonlayer}{background}
			\draw[thick,blue,->] ($(im.center)+(0.8*\ScaleFactor,1.35*\ScaleFactor)$)--++(1*\ScaleFactor,0*\ScaleFactor)node[right]{$\bm{v}_1$};
			\draw[thick,blue,->] ($(im.center)+(0.55*\ScaleFactor,-0.2*\ScaleFactor)$)--++(-55:1*\ScaleFactor)node[right]{$\bm{w}_1$};
		\end{pgfonlayer}
	\end{tikzpicture}

%% file: TikZ/spm_d1_jambe2.tex
	\def\ScaleFactor{1}

	\begin{tikzpicture}[>=latex']
		
		\node[] at (0,0) (im) {\centering\includegraphics[scale=0.14]{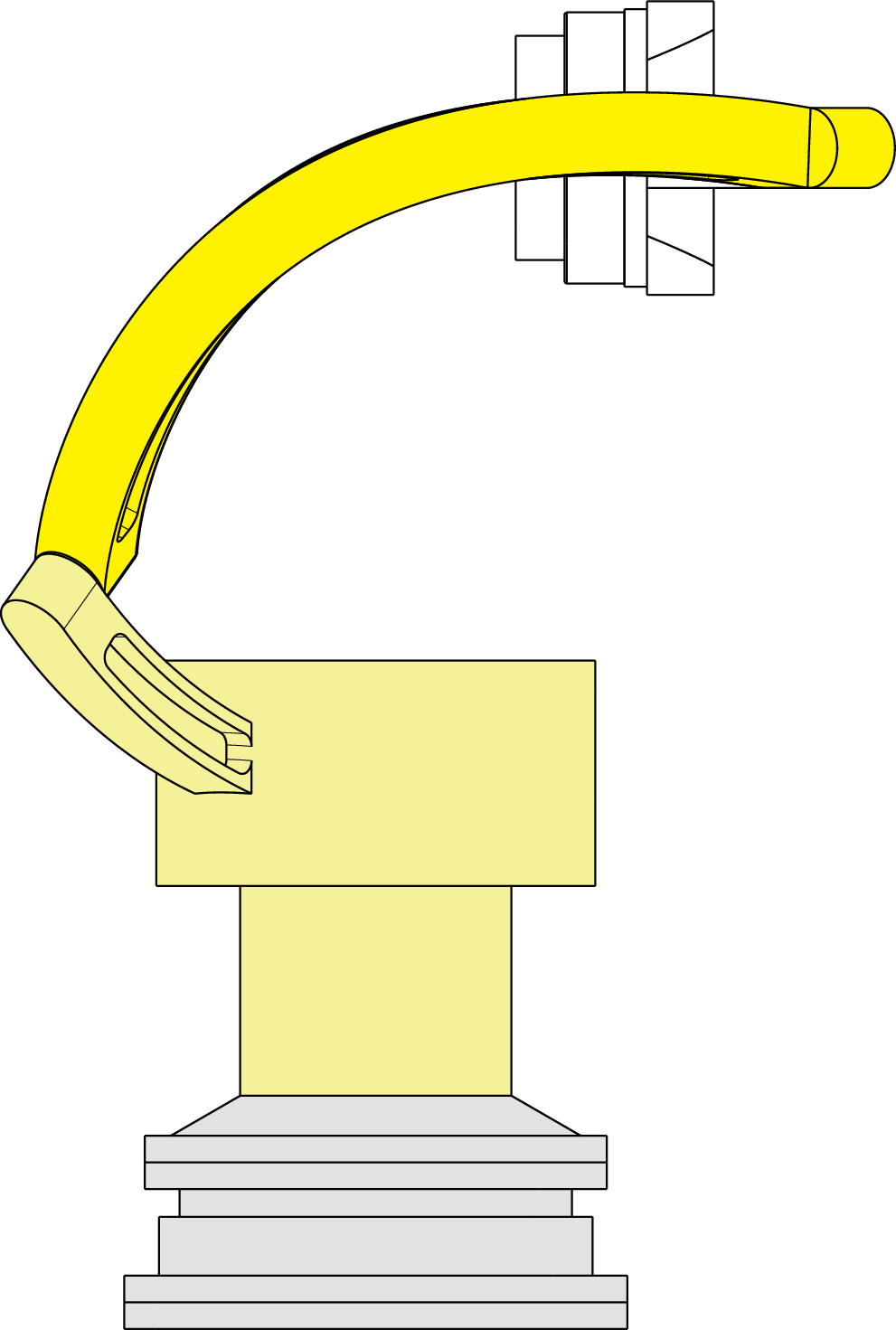}};
		
		
		\draw[densely dashed] ($(im.center)+(-0.6*\ScaleFactor,-0.35*\ScaleFactor)$)--($(im.center)+(-0.6*\ScaleFactor,-0.75*\ScaleFactor)$)node[near end](a){};
		\draw[densely dashed] ($(im.center)+(-1.05*\ScaleFactor,0.05*\ScaleFactor)$)--($(im.center)+(-1.3*\ScaleFactor,-0.25*\ScaleFactor)$)node[near end](b){};		
		\draw[dashed] ($(im.center)+(-1*\ScaleFactor,0.15*\ScaleFactor)$)--($(im.center)+(-0.25*\ScaleFactor,0.15*\ScaleFactor)$)node[near end](c){};
		\draw[dashed] ($(im.center)+(1.05*\ScaleFactor,1.35*\ScaleFactor)$)--($(im.center)+(1.05*\ScaleFactor,0.75*\ScaleFactor)$)node[near end](d){};
		
		\draw[<->,thick] ($(im.center)+(-0.6*\ScaleFactor,-0.75*\ScaleFactor)$)to[in=-45,out=160]($(im.center)+(-1.3*\ScaleFactor,-0.25*\ScaleFactor)$)node[right,below=1em]{$\alpha_{1,2}$};
		\draw[<->,thick] ($(im.center)+(-0.45*\ScaleFactor,0.15*\ScaleFactor)$)to[in=170,out=85]($(im.center)+(1.05*\ScaleFactor,0.85*\ScaleFactor)$);
		\node[anchor=center] at ($(im.center)+(0.5*\ScaleFactor,0.5*\ScaleFactor)$){$\alpha_{2,2}$};

		\begin{pgfonlayer}{background}
			\draw[thick,blue,->] ($(im.center)+(-1.05*\ScaleFactor,0.05*\ScaleFactor)$)--++(-130:1*\ScaleFactor)node[midway,left]{$\bm{w}_2$};
			\draw[thick,blue,->] ($(im.center)+(1.05*\ScaleFactor,1.35*\ScaleFactor)$)--++(0:1*\ScaleFactor)node[midway,above]{$\bm{v}_2$};
		\end{pgfonlayer}
	\end{tikzpicture}

%% file: TikZ/spm_d1_jambe3.tex
	\def\ScaleFactor{1}

	\begin{tikzpicture}[>=latex']
		
		\node[] at (0,0) (im) {\centering\includegraphics[scale=0.14]{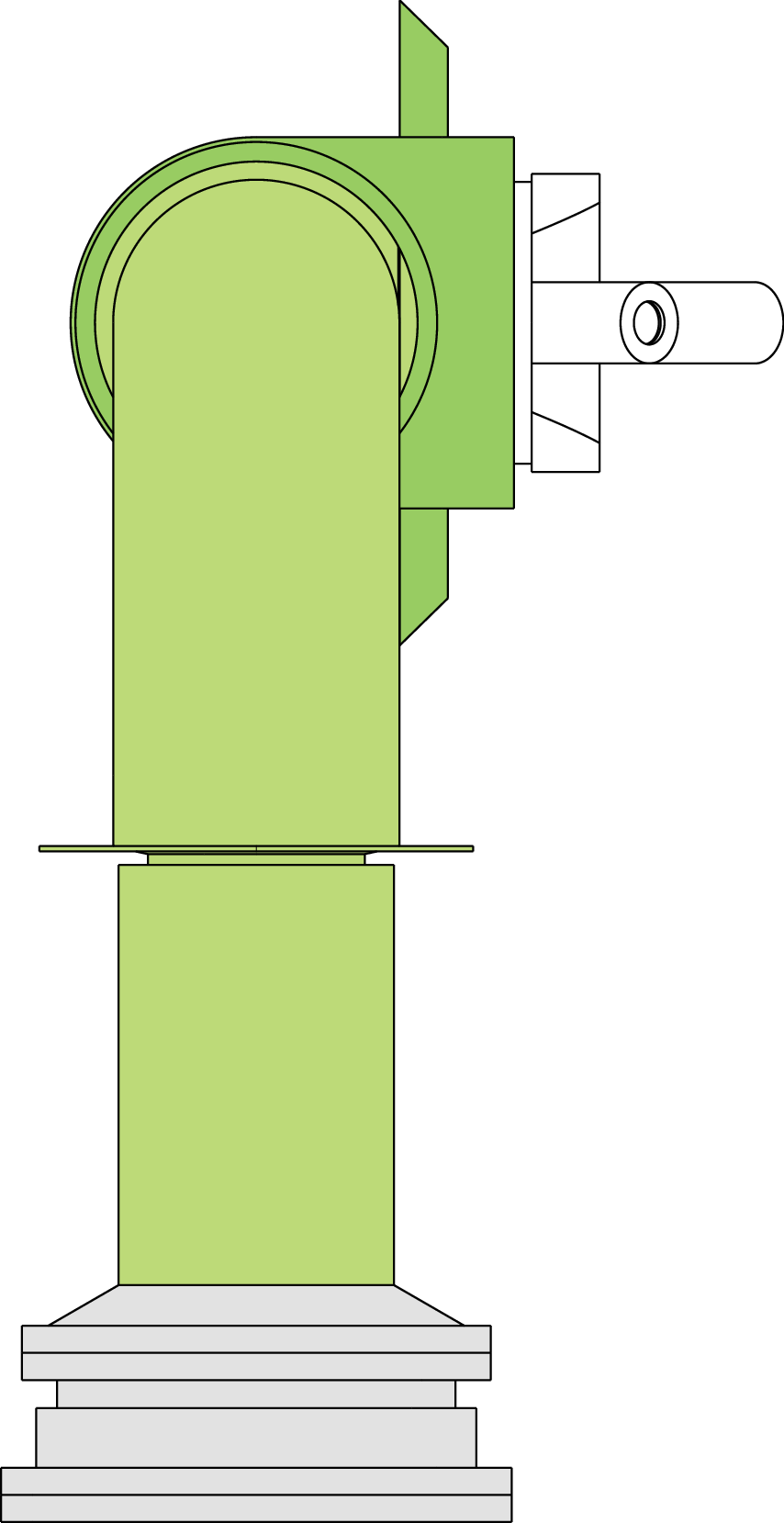}};
		
		
		\draw[densely dashed] ($(im.center)+(-0.35*\ScaleFactor,-0.3*\ScaleFactor)$)--($(im.center)+(-0.35*\ScaleFactor,-0.75*\ScaleFactor)$)node[near end](a){};
		\draw[densely dashed] ($(im.center)+(-0.35*\ScaleFactor,1.1*\ScaleFactor)$)--($(im.center)+(-1*\ScaleFactor,1.1*\ScaleFactor)$)node[near end](b){};	
		\draw[dashed] ($(im.center)+(-0.35*\ScaleFactor,-0.225*\ScaleFactor)$)--($(im.center)+(0.5*\ScaleFactor,-0.225*\ScaleFactor)$)node[near end](c){};
		\draw[dashed] ($(im.center)+(0.66*\ScaleFactor,1.1*\ScaleFactor)$)--($(im.center)+(0.66*\ScaleFactor,0.5*\ScaleFactor)$)node[near end](d){};
		
		\draw[<->,thick] ($(im.center)+(-0.35*\ScaleFactor,-0.75*\ScaleFactor)$)to[in=-90,out=180]($(im.center)+(-1*\ScaleFactor,1.1*\ScaleFactor)$);
		\node[anchor=center] at ($(im.center)+(-1.5*\ScaleFactor,0*\ScaleFactor)$){$\alpha_{1,3}$};
		\draw[<->,thick] ($(im.center)+(0.25*\ScaleFactor,-0.225*\ScaleFactor)$)to[in=180,out=90]($(im.center)+(0.66*\ScaleFactor,0.5*\ScaleFactor)$);
		\node[anchor=center] at ($(im.center)+(0.75*\ScaleFactor,0.15*\ScaleFactor)$){$\alpha_{2,3}$};

		\begin{pgfonlayer}{background}
			\draw[thick,blue,->] ($(im.center)+(0.66*\ScaleFactor,1.1*\ScaleFactor)$)--++(0:1*\ScaleFactor)node[midway,above]{$\bm{v}_3$};
		\end{pgfonlayer}

		\draw[thick,blue] ($(im.center)+(-0.35*\ScaleFactor,1.1*\ScaleFactor)$)node[below=0.5em]{$\bm{w}_3$} circle (0.125*\ScaleFactor);
		\draw[draw=none,fill=blue] ($(im.center)+(-0.35*\ScaleFactor,1.1*\ScaleFactor)$) circle (0.075*\ScaleFactor);
	\end{tikzpicture}

%% file: TikZ/spm_d1_eta.tex
	\def\ScaleFactor{1}

	\begin{tikzpicture}[>=latex']
		
		\node[] at (0,0) (im) {\centering\includegraphics[scale=0.14]{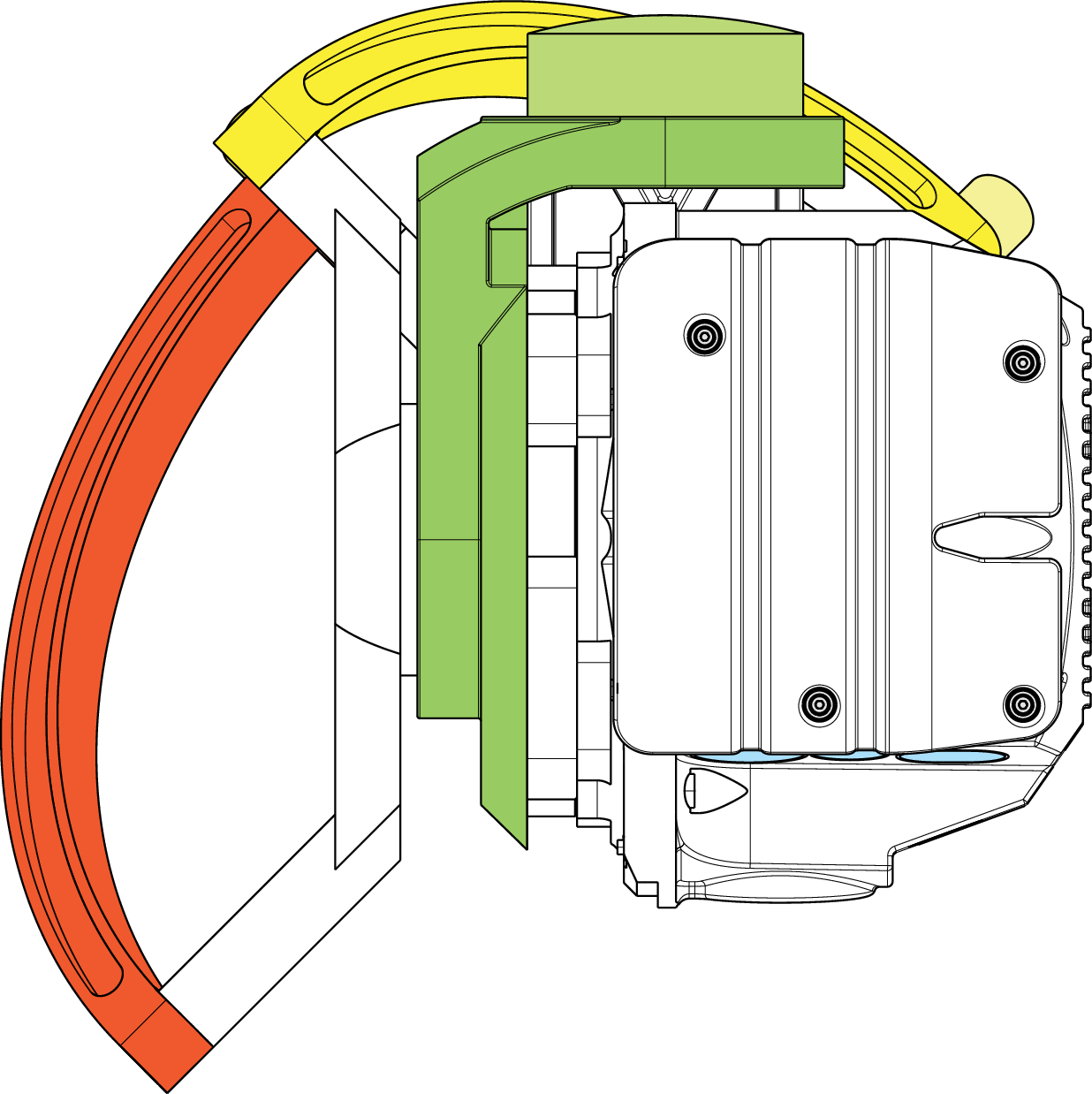}};
		
		
		\draw[densely dashed] ($(im.center)+(-0.5*\ScaleFactor,0*\ScaleFactor)$)--($(im.center)+(-2*\ScaleFactor,0*\ScaleFactor)$)node[above,rotate=90](a){$\eta_3=0$};
		\draw[densely dashed] ($(im.center)+(-0.5*\ScaleFactor,-0.8*\ScaleFactor)$)--($(im.center)+(-1.5*\ScaleFactor,-1.825*\ScaleFactor)$)node[near end](b){};
		\draw[densely dashed] ($(im.center)+(-0.5*\ScaleFactor,0.825*\ScaleFactor)$)--($(im.center)+(-1.5*\ScaleFactor,1.85*\ScaleFactor)$)node[near end](c){};
		
		\draw[<->,thick] ($(im.center)+(-1.85*\ScaleFactor,0*\ScaleFactor)$)to[in=140,out=-90]($(im.center)+(-1.2*\ScaleFactor,-1.525*\ScaleFactor)$);
		\node[anchor=center] at ($(im.center)+(-1.95*\ScaleFactor,-0.95*\ScaleFactor)$){$\eta_1$};
		\draw[<->,thick] ($(im.center)+(-1.75*\ScaleFactor,0*\ScaleFactor)$)to[in=-145,out=90]($(im.center)+(-1*\ScaleFactor,1.35*\ScaleFactor)$);
		\node[anchor=center] at ($(im.center)+(-1.8*\ScaleFactor,0.95*\ScaleFactor)$){$\eta_2$};

		\begin{pgfonlayer}{background}
			\draw[thick,blue,->] ($(im.center)+(-0.5*\ScaleFactor,-0.8*\ScaleFactor)$)--++(-134:1.5*\ScaleFactor)node[near end,right,anchor=north west]{$\bm{v}_1$};
			\draw[thick,blue,->] ($(im.center)+(-0.5*\ScaleFactor,0.8*\ScaleFactor)$)--++(134:1.15*\ScaleFactor)node[near end,right,anchor=south west]{$\bm{v}_2$};
			\draw[thick,blue,->] ($(im.center)+(-0.5*\ScaleFactor,0*\ScaleFactor)$)--++(180:0.5*\ScaleFactor)node[midway,below]{$\bm{v}_3$}node[midway,above,black]{}; 

			\draw[thick,->] ($(im.center)+(0.5*\ScaleFactor,-1.5*\ScaleFactor)$)--++(-1*\ScaleFactor,0*\ScaleFactor)node[above,midway]{$\bm{y}_\star$};
			\draw[thick] ($(im.center)+(0.5*\ScaleFactor,-1.5*\ScaleFactor)$)node[below=0.5em]{$\bm{z}_\star$} circle (0.125*\ScaleFactor);
			\draw[draw=none,fill=black] ($(im.center)+(0.5*\ScaleFactor,-1.5*\ScaleFactor)$) circle (0.075*\ScaleFactor);
		\end{pgfonlayer}
	\end{tikzpicture}

%% file: TikZ/spm_d1_st1.tex
	\def\ScaleFactor{2}
	\begin{tikzpicture}[>=latex',every text node part/.style={align=center}]
		\begin{axis}[anchor=origin, x=\ScaleFactor cm, y=\ScaleFactor cm, xmin=-1.75,
xmax=1.75, ymin=-1.75, ymax=1.75, xtick={-1.5,-1,-0.5,0,0.5,1,1.5}, ytick={-1.5,-1,-0.5,0,0.5,1,1.5}, xmajorgrids,ymajorgrids]
			\addplot[only marks, mark=*, mark options={}, mark size=0.25pt, color=black, fill=black!15] file[]{TikZ/SPM_D1_ST1.dat};
		\end{axis}
		\draw[very thin,draw=black,fill=none,densely dashed] (-0.2*\ScaleFactor,-0.875*\ScaleFactor)rectangle(0.2*\ScaleFactor,0.875*\ScaleFactor);
		\draw[->,very thin] (-1.75*\ScaleFactor,0)--(1.75*\ScaleFactor,0)node[right=0.25em]{$\chi_1$};
		\draw[->,very thin] (0,-1.75*\ScaleFactor)--(0,1.75*\ScaleFactor)node[above=0.25em]{$\chi_2$};
		\node[rounded corners, fill=white, fill opacity=0.75, text opacity=1] at (-1.25*\ScaleFactor,1.35*\ScaleFactor){$\chi_3\in\bbR$};
	\end{tikzpicture}

%% file: TikZ/spm_stab_inertielle.tex
	\def\ScaleFactor{1}
	\def\CoC{(-0.1*\ScaleFactor, 1.35*\ScaleFactor)}
	\def\CoC{(-0.1*\ScaleFactor, 1.35*\ScaleFactor)}
	\def\rotang{42}

	\begin{tikzpicture}[>=latex']
		
		\node at \CoC (CoC){};
		
		\begin{scope}[rotate around = (\rotang:(CoC))] 
			\node[transform shape,opacity=0.75] at (0,0) (im) {\centering\includegraphics[scale=0.25]{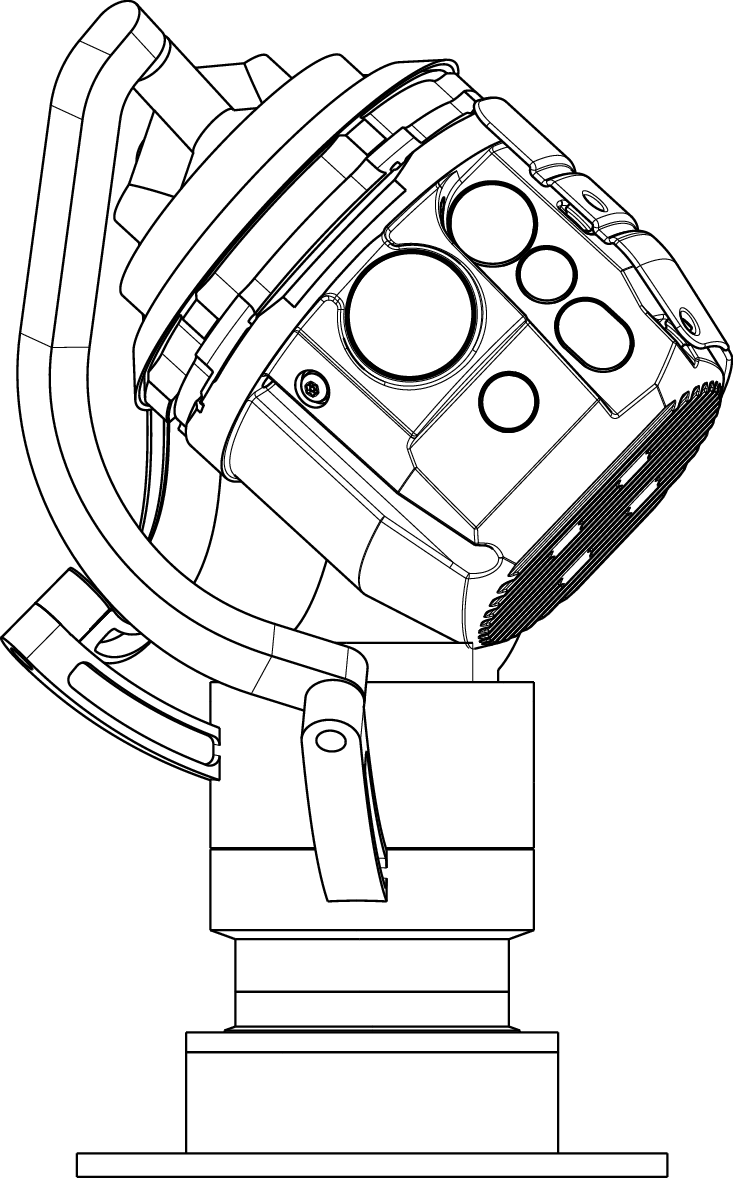}};
			\draw[] (-2.5*\ScaleFactor,-2.5*\ScaleFactor)--(2.5*\ScaleFactor,-2.5*\ScaleFactor)node[below,midway,transform shape]{Carrier};
			\draw[dashed] (-3.5*\ScaleFactor,-2.5*\ScaleFactor)node(a){}--(-2.5*\ScaleFactor,-2.5*\ScaleFactor)(2.5*\ScaleFactor,-2.5*\ScaleFactor)--(3.5*\ScaleFactor,-2.5*\ScaleFactor)node(b){};
			
			\draw[<->,transform shape] (2*\ScaleFactor,-3*\ScaleFactor)node[below,rotate=-\rotang]{$\bm{\nu}$}to[in=-60,out=60](2*\ScaleFactor,-2*\ScaleFactor);
		\end{scope}
		
		\draw[thick] (-0.5*\ScaleFactor,3.5*\ScaleFactor)node[left=0.5em]{$\mathrm{LOS}$}node[anchor=north west]{$O$} circle (0.125*\ScaleFactor);
		\draw[draw=none,fill=black] (-0.5*\ScaleFactor,3.5*\ScaleFactor) circle (0.075*\ScaleFactor);
		
		
		\begin{pgfonlayer}{background}
			\draw[draw=none,fill=white,outer color = white, inner color = black!50!white,fill opacity=0.25] (a)-|(b)--cycle;
		\end{pgfonlayer}
		
		\node at (5*\ScaleFactor,2.5*\ScaleFactor) (O_dep){};
		\begin{scope}[rotate around = (\rotang:(O_dep))]
			\draw[<->,dashed] (5*\ScaleFactor,4*\ScaleFactor)node[above]{$\bm{z}_\frakb$}|-(3.5*\ScaleFactor,2.5*\ScaleFactor)node[left]{$\bm{y}_\frakb$};
		\end{scope}
		
		\draw[<->] (5*\ScaleFactor,4*\ScaleFactor)node[above]{$\bm{z}_\star{\;\color{blue}\equiv\bm{z}_\fraki}$}|-(3.5*\ScaleFactor,2.5*\ScaleFactor)node[left]{${\color{blue}\bm{y}_\fraki\equiv\;}\bm{y}_\star$};
		\draw[thick,fill=white] (5*\ScaleFactor,2.5*\ScaleFactor)node[anchor=north west]{$O$}node[anchor=south west]{$\bm{x}_\frakb=\bm{x}_\star{\;\color{blue}\equiv\bm{x}_\fraki}$} circle (0.125*\ScaleFactor);
		\draw[thick] (4.9*\ScaleFactor,2.4*\ScaleFactor)--(5.1*\ScaleFactor,2.6*\ScaleFactor);
		\draw[thick] (4.9*\ScaleFactor,2.6*\ScaleFactor)--(5.1*\ScaleFactor,2.4*\ScaleFactor);
		
		
		\begin{scope}[rotate around = (\rotang:(O_dep))]
		\end{scope}
	\end{tikzpicture}

%% file: TikZ/spm_stab_inertielle_bd1.tex
    \begin{tikzpicture}[>=latex',every text node part/.style={align=center}]
        \sbEntree{E}
            \draw[] ($(E)+(0.35em,0)$)--($(E)-(1em,0)$);
        \sbComp*[4]{a}{E}
        \sbRelier[$\rep{\overline{\bm{\Omega}}_{\star/\fraki}}{\star}\hspace*{2.75em}$]{E}{a}
        \sbBloc[3]{b}{$\bm{K}_0(s)$}{a}
        \sbRelier[$\bm{\varepsilon}_\Omega$]{a}{b}
        \sbBloc[1]{b2}{$\bm{T}^{-1}$}{b}
        \sbRelier[]{b}{b2}
        \sbBloc[2]{b3}{$\bm{J}^{-1}$}{b2}
        \sbRelier[$\overline{\dot{\bm{\chi}}}$]{b2}{b3}

        \draw[dashed,rounded corners] ($(b.north west)+(-0.75em,0.75em)$) rectangle ($(b3.south east)+(0.75em,-0.75em)$)node[midway,above=2.5em]{$\bm{K}_v$};
        
        \sbBloc[4]{c}{$\bm{H}_m(s)$}{b3}
        \sbRelier[$\overline{\dot{\bm{\theta}}}$]{b3}{c}

        \draw[<-] (c.120)--++(0,1.3)node[above]{$\bm{\Gamma}_{f}$};

        \sbBloc[1.75]{c2}{$\bm{J}$}{c}
        \sbRelier[$\dot{\bm{\theta}}$]{c}{c2}
        \sbBloc[1.75]{c3}{$\bm{T}$}{c2}
        \sbRelier[$\dot{\bm{\chi}}$]{c2}{c3}
        \sbCompSum*[8]{d}{c3}{+}{}{+}{}

        \draw[dashed,rounded corners] ($(c.north west)+(-0.75em,2.75em)$) rectangle ($(c3.south east)+(0.75em,-0.75em)$)node[midway,below=2.5em]{};

        \sbDecaleNoeudy[-5.5]{d}{d1}
        \sbRelier[]{d1}{d}
        \node[] at (d1.north) {$\rep{\bm{\Omega}_{\frakb/\fraki}}{\star}$};

        \draw[<-] (c.60)--++(0,0.45)--++(5.99,0);

        \sbRelier[$\quad\rep{\bm{\Omega}_{\star/\frakb}}{\star}\quad$]{c3}{d}
        \sbSortie[4.5]{S}{d}
        \sbRelier[$\rep{\bm{\Omega}_{\star/\fraki}}{\star}$]{d}{S}

        \sbDecaleNoeudy[8.5]{d-S}{u}
        \sbBlocr[17]{cap1}{$\bm{\beta}(s)$}{u}
        \sbRelieryx{d-S}{cap1}
        \sbRelierxy[$\rep{\hat{\bm{\Omega}}_{\star/\fraki}}{\star}$]{cap1}{a}

        \node[fill=white,anchor=center] at ($(c.north west)+(9.5em,4.5em)$) (robot){\includegraphics[scale=0.065]{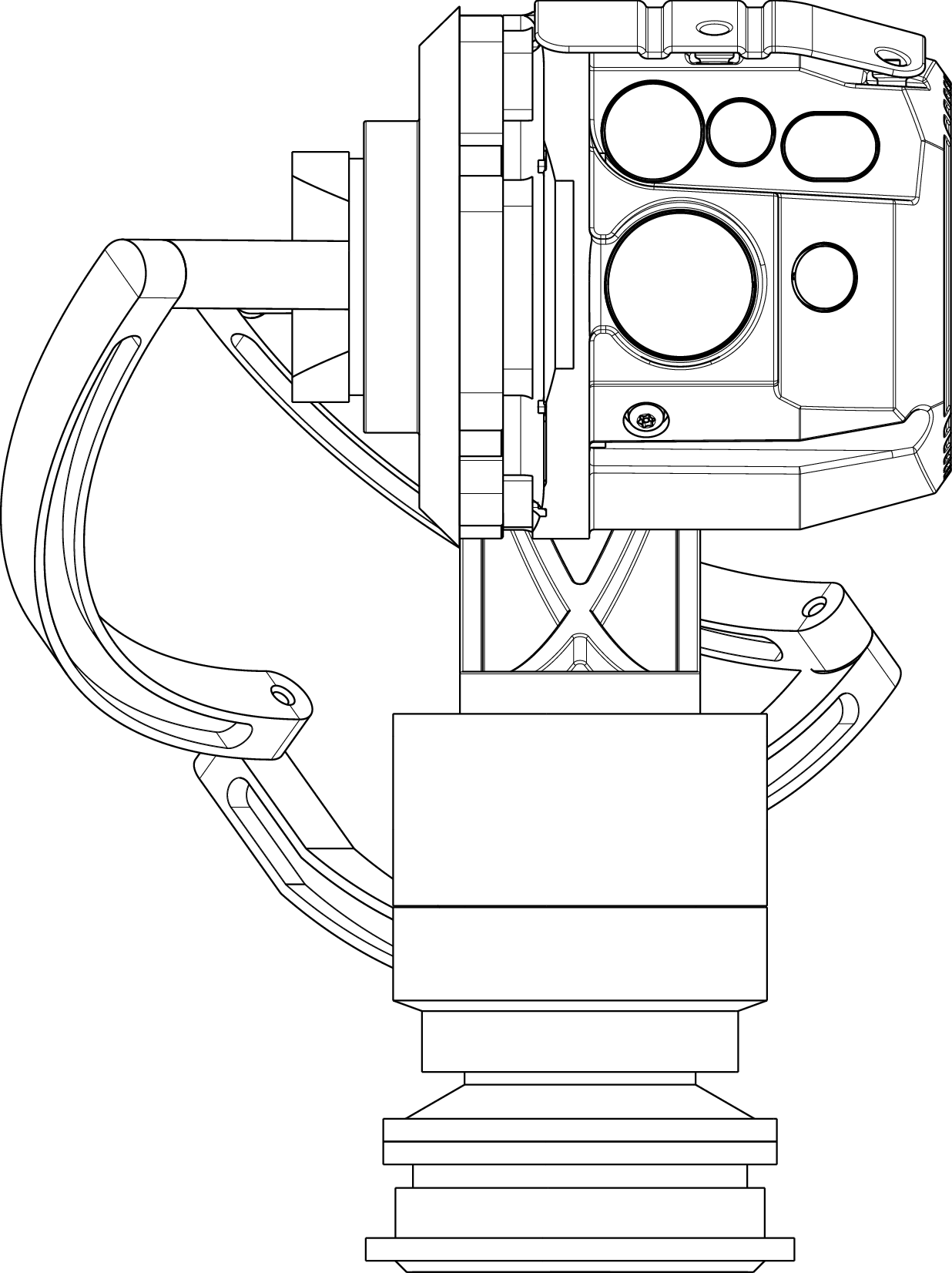}};
    \end{tikzpicture}

%% file: TikZ/spm_asv_moteur.tex
    \begin{tikzpicture}[>=latex,every text node part/.style={align=center}]
        \sbEntree{E}
        \sbComp*[4.5]{a}{E}
        \sbRelier[$\overline{\dot{\theta}}_k\quad$]{E}{a}
        \sbBloc[1]{a2}{$K_m(s)$}{a}
        \sbRelier[]{a}{a2}
        \sbBloc[2]{b}{$\dfrac{1}{R}$}{a2}
        \sbRelier[$u_k$]{a2}{b}
        \sbBloc[2]{b2}{$K_{c}$}{b}
        \sbRelier[$i_k$]{b}{b2}
        \sbCompSum*[6.5]{b3}{b2}{+}{}{+}{}
        \sbRelier[$\Gamma_{m,k}\quad$]{b2}{b3}
        \sbBloc[3]{c}{$N^2$}{b3}
        \sbRelier[$\quad\tilde{\Gamma}_{k}$]{b3}{c}
        \sbBloc[2]{c1}{$\dfrac{1}{J_m}$}{c}
        \sbRelier[$\Gamma_{k}$]{c}{c1}
        \sbBloc[1]{c2}{$\dfrac{1}{s}$}{c1}
        \sbRelier[]{c1}{c2}
        \sbCompSum*[4.5]{c3}{c2}{-}{}{+}{}
        \sbRelier[]{c2}{c3}
        \sbSortie[3]{S}{c3}
        \sbRelier[$\dot{\theta}_k$]{c3}{S}
        \sbRenvoi[5]{c3-S}{a}{}

        \sbDecaleNoeudy[-4]{b3}{bd}
        \sbRelier[]{bd}{b3}
        \node[] at (bd.north) {$\Gamma_{f,k}$};

        \sbDecaleNoeudy[-4]{c3}{cd}
        \sbRelier[]{cd}{c3}
        \node[] at (cd.north) {${\Omega}_{\frakb/\fraki,k}$};
        
        \sbDecaleNoeudy[5]{c3-S}{u}
        \draw[dashed,rounded corners] ($(a.north west)+(-1.5em,1.75em)$) rectangle ($(u.south east)+(0.5em,-1em)$);
        \node at ($(a.north west)+(2em,3em)$) {$H_{m,kk}(s)$};
    \end{tikzpicture}

%% file: TikZ/disturb_rej.tex
%
%
\definecolor{mycolor1}{rgb}{0.00000,0.44700,0.74100}%
\begin{tikzpicture}

\begin{axis}[%
width=4.396in,
height=1.713in,
at={(0.883in,2.125in)},
scale only axis,
separate axis lines,
every outer x axis line/.append style={white!40!black},
every x tick label/.append style={font=\color{white!40!black}},
every x tick/.append style={white!40!black},
xmode=log,
xmin=0.01,
xmax=100000,
xticklabels={\empty},
every outer y axis line/.append style={white!40!black},
every y tick label/.append style={font=\color{white!40!black}},
every y tick/.append style={white!40!black},
ymin=-200,
ymax=10,
ylabel={Magnitude (dB)},
axis background/.style={fill=white},
xmajorgrids,
xminorgrids,
ymajorgrids
]
\addplot [color=mycolor1, forget plot]
  table[row sep=crcr]{%
0.01	-164.187266702749\\
7.83846041097916	-48.4798078066846\\
14.5700666510016	-37.8588917595775\\
19.8644672947871	-32.6475363369516\\
27.082721744347	-27.5690861137116\\
36.9239107294943	-22.7061743596485\\
43.1137088539896	-20.3878395277986\\
50.3411435685681	-18.1630631413079\\
58.7801607227497	-16.0440136864375\\
68.6338658534042	-14.04016973561\\
80.1394124150505	-12.1574766133222\\
93.5737094563003	-10.3982822449203\\
109.26008611174	-8.76212963490198\\
127.57607330635	-7.24714601224258\\
148.962489958348	-5.85154321558562\\
173.934052361877	-4.57472582302316\\
203.09176209048	-3.41765854790009\\
237.137370566166	-2.38238409188295\\
276.890268419563	-1.47084154098613\\
323.307206124522	-0.683367401776366\\
377.50532053243	-0.0173956397137545\\
440.789021496211	0.533176860429478\\
514.681385675684	0.977744040866696\\
600.960813093519	1.32798441951789\\
701.703828670372	1.59645218071978\\
819.335058863593	1.79515050046282\\
956.685586218048	1.93432440484116\\
1117.06108627509	2.02149877553609\\
1304.321386719	2.06059291016734\\
1522.97336354776	2.05076682597175\\
1685.55746915114	2.01458686020723\\
1940.73257809935	1.91582899175751\\
2201.92639153178	1.76382238251068\\
2465.5895423043	1.55543588056787\\
2728.45067501463	1.28729308435265\\
2987.60402683936	0.959102493789942\\
3240.56027572779	0.577682174620378\\
3485.26607979408	0.160660512318202\\
4181.85359619385	-1.01547022596952\\
4440.20917602624	-1.29136818784943\\
4752.82805469714	-1.42323986398088\\
5134.39362633931	-1.32017327708971\\
5604.70737215873	-0.982508651143348\\
6190.90944056674	-0.524975440474805\\
6930.86054761015	-0.091165479908085\\
7878.40218814234	0.227886762746323\\
8376.77640068313	0.322278077315872\\
9781.03055009056	0.436296558807868\\
11420.6890629203	0.431065530559977\\
13335.2143216332	0.362279956809772\\
18180.9002735653	0.174624979582148\\
21228.6842214614	0.0958071667644163\\
24787.3882477536	0.0378298118239115\\
28942.6612471669	0.000381975218658681\\
33794.5100022479	-0.0203323882130348\\
39459.7060905639	-0.0292186886937031\\
46074.5962776266	-0.0307637816628983\\
62816.9611574505	-0.0242943786794569\\
100000	-0.0121883865128893\\
};
\end{axis}

\begin{axis}[%
width=4.396in,
height=1.519in,
at={(0.883in,0.481in)},
scale only axis,
separate axis lines,
every outer x axis line/.append style={white!40!black},
every x tick label/.append style={font=\color{white!40!black}},
every x tick/.append style={white!40!black},
xmode=log,
xmin=0.01,
xmax=100000,
every outer y axis line/.append style={white!40!black},
every y tick label/.append style={font=\color{white!40!black}},
every y tick/.append style={white!40!black},
ymin=-10.8,
ymax=1000,
ytick={   0,  180,  360,  540,  720,  900},
ylabel={Phase (deg)},
axis background/.style={fill=white},
xmajorgrids,
xminorgrids,
ymajorgrids
]
\addplot [color=mycolor1, forget plot]
  table[row sep=crcr]{%
0.01	899.989039420941\\
0.0159192673694222	899.982551561301\\
0.0253423073579151	899.972223364552\\
0.0345510729459229	899.962130025774\\
0.0471060754198309	899.948369020047\\
0.0642232542222871	899.929607611105\\
0.0875603910141324	899.904028783831\\
0.102238715529896	899.88794051568\\
0.119377664171448	899.869155276126\\
0.139389727552476	899.847220960526\\
0.162756544802672	899.821609678768\\
0.190040495388238	899.791705053094\\
0.221898234145879	899.756787387926\\
0.259096495283708	899.716016355714\\
0.302530545710282	899.668410783688\\
0.353245731817091	899.612825057497\\
0.41246263829016	899.547921577789\\
0.481606464457793	899.472138613121\\
0.562341325190349	899.38365278551\\
0.656610301883835	899.280335301685\\
0.766682207454575	899.159700901991\\
0.895206190858038	899.018848338624\\
1.04527549532031	898.854391014858\\
1.22050190478482	898.662376218044\\
1.42510267030311	898.438191164916\\
1.66400200847092	898.176453856097\\
1.94294961471587	897.870886523099\\
2.26865904374328	897.5141692732\\
2.64896928761042	897.097771442659\\
3.09303344019778	896.611758235514\\
3.61153898873976	896.044570581204\\
4.21696503428582	895.382776993088\\
4.92388263170723	894.610797869436\\
5.74930547768411	893.710605636065\\
6.71309938683008	892.661409112115\\
7.83846041097813	891.439338533725\\
9.15247310877371	890.017160220018\\
10.6867624016452	888.364068684218\\
12.4782547046992	886.445630909238\\
14.5700666510016	884.223993542144\\
17.0125427985279	881.658507151535\\
19.8644672947858	878.706964122209\\
23.1944786607596	875.327667472995\\
27.0827217443434	871.482507150161\\
31.6227766016838	867.141058626836\\
36.9239107294991	862.285375540096\\
43.1137088539868	856.914619393619\\
50.3411435685714	851.048100173626\\
58.780160722742	844.725047757994\\
68.6338658534042	837.999944809788\\
80.1394124150558	830.93369350216\\
93.5737094562941	823.582748700328\\
127.576073306367	808.177363021908\\
173.934052361889	791.901319193082\\
237.137370566166	774.684799373113\\
323.307206124501	756.484147169925\\
440.789021496153	737.318706498395\\
514.681385675684	727.337237096071\\
600.960813093558	717.014414136012\\
701.703828670326	706.244149852366\\
819.335058863593	694.876747224581\\
956.685586218173	682.718549095383\\
1117.06108627502	669.532287270773\\
1304.32138671908	655.036977793003\\
1522.97336354756	638.907458884895\\
1685.55746915092	627.291753026558\\
1940.73257809935	609.516606240359\\
2201.92639153207	591.762536725878\\
2465.58954230446	574.233705581806\\
2728.45067501427	557.180866470471\\
3240.56027572737	525.657320355015\\
3938.53055266398	488.874669724285\\
4440.20917602595	468.204971259952\\
5134.39362633898	444.048766193437\\
5604.70737215836	428.130800222885\\
8376.77640068258	350.164538006793\\
9781.0305500912	324.435343462309\\
11420.6890629189	302.341017094377\\
13335.2143216332	283.654385265212\\
15570.6840475389	267.96988001056\\
18180.9002735641	254.841214368026\\
21228.6842214628	243.840412810698\\
24787.3882477504	234.58665692307\\
28942.6612471669	226.759897082094\\
33794.5100022501	220.102832607373\\
39459.7060905613	214.414126996235\\
46074.5962776266	209.537163263434\\
53798.3840344434	205.348471123315\\
62816.9611574464	201.74827132321\\
73347.3817081724	198.653786257548\\
85643.0859486936	195.994901473103\\
100000	193.711442601125\\
};
\end{axis}

\begin{axis}[%
width=4.521in,
height=3.566in,
at={(0.758in,0.281in)},
scale only axis,
xmin=0,
xmax=1,
xtick={\empty},
xlabel={Frequency  (rad/s)},
ymin=0,
ymax=1,
ytick={\empty},
axis line style={draw=none},
ticks=none,
title style={font=\bfseries},
title={Bode Diagram},
axis x line*=bottom,
axis y line*=left
]
\end{axis}
\end{tikzpicture}%

%% file: TikZ/AsyCoSPM_1_dot_CHI_reel_inertiel_trans.tex
%
%
\definecolor{mycolor1}{rgb}{0.00000,0.44700,0.74100}%
\definecolor{mycolor2}{rgb}{0.85000,0.32500,0.09800}%
\definecolor{mycolor3}{rgb}{0.92900,0.69400,0.12500}%
\begin{tikzpicture}

\begin{axis}[%
width=4.521in,
height=2.5in,
at={(0.758in,0.481in)},
scale only axis,
xmin=0,
xmax=0.25,
xtick={0,0.05,0.1,0.15,0.2,0.25},
xlabel style={font=\color{white!15!black}},
xlabel={$t$ $\mathrm{[s]}$},
ymin=-0.8,
ymax=0.4,
ylabel style={font=\color{white!15!black}},
ylabel={$\rep{\bm{\Omega}_{\star/\fraki}}{\star}$ $\mathrm{[rad/s]}$},
axis background/.style={fill=white},
xmajorgrids,
ymajorgrids,
legend style={legend cell align=left, align=left, draw=white!15!black}
]
\addplot [color=mycolor1]
  table[row sep=crcr]{%
  0	0\\
  0.002	5.28123410974413e-05\\
  0.005	0.000136548008645043\\
  0.00700000000000001	-3.03058938967427e-05\\
  0.011	-0.000524327871742836\\
  0.013	-0.000511943141332738\\
  0.018	-9.74869950637891e-05\\
  0.026	-0.000143777322980987\\
  0.03	-5.12829585757046e-05\\
  0.037	-4.96156375431323e-05\\
  0.044	-2.4502150795247e-05\\
  0.297	5.61296523649002e-07\\
};
\addlegendentry{$\rep{\Omega_{\star/\fraki,1}}{\star}$}

\addplot [color=mycolor2]
  table[row sep=crcr]{%
  0	0\\
  3.1554436208836e-05	0.0822467033424113\\
  0.003	0.0822465781792698\\
  0.004	0.0515423368672068\\
  0.00600000000000001	-0.0162528410989545\\
  0.00700000000000001	-0.0370979854427983\\
  0.00800000000000001	-0.0438994494523052\\
  0.00900000000000001	-0.0383932119841919\\
  0.01	-0.0256213855762789\\
  0.011	-0.0112836682333203\\
  0.012	-4.88147337799893e-05\\
  0.013	0.0055903135104407\\
  0.014	0.00546015154302437\\
  0.015	0.00115967821747792\\
  0.016	-0.0048763366228014\\
  0.017	-0.0103147566755731\\
  0.018	-0.0135911706750881\\
  0.019	-0.0141775395489374\\
  0.02	-0.012474827245492\\
  0.022	-0.00624410342317538\\
  0.023	-0.00372435783905917\\
  0.024	-0.00236230736210674\\
  0.025	-0.00216569634951924\\
  0.026	-0.0028013718467535\\
  0.028	-0.00464445232561522\\
  0.029	-0.00508882916480397\\
  0.03	-0.00501675615245561\\
  0.031	-0.00451255847748128\\
  0.033	-0.00300946161998111\\
  0.034	-0.00240123763348388\\
  0.035	-0.00203232015651117\\
  0.036	-0.0018997579227128\\
  0.038	-0.00203988667549299\\
  0.04	-0.00213019808617276\\
  0.042	-0.00186858196905992\\
  0.045	-0.00127522045614908\\
  0.047	-0.00108021325632623\\
  0.055	-0.000699894419142089\\
  0.059	-0.000524835159692816\\
  0.077	-0.000164629885838374\\
  0.095	-5.1509089820323e-05\\
  0.136	-5.36581652865742e-06\\
  0.297	-2.0616044061339e-06\\
};
\addlegendentry{$\rep{\Omega_{\star/\fraki,2}}{\star}$}

\addplot [color=mycolor3]
  table[row sep=crcr]{%
  0	0\\
  3.1554436208836e-05	0.109662271123215\\
  0.00020095091452077	-0.00895311423279255\\
  0.001	-0.356844402978298\\
  0.002	-0.605712529810613\\
  0.003	-0.738486799781578\\
  0.004	-0.676105856769936\\
  0.005	-0.46901590387887\\
  0.00700000000000001	0.0704470180227381\\
  0.00800000000000001	0.254513365682064\\
  0.00900000000000001	0.333662951294379\\
  0.01	0.316537590740019\\
  0.011	0.236069962891606\\
  0.012	0.134154154979439\\
  0.013	0.0472916543123091\\
  0.014	-0.00250872872037922\\
  0.015	-0.010317904229307\\
  0.016	0.0147872311754562\\
  0.018	0.0962496087520969\\
  0.019	0.122558054976839\\
  0.02	0.129622636480219\\
  0.021	0.119080178277481\\
  0.022	0.0973454533965972\\
  0.023	0.0725126530366696\\
  0.024	0.0515498513901163\\
  0.025	0.0385402376755081\\
  0.026	0.0342217709734693\\
  0.027	0.0366292417054047\\
  0.029	0.0479796679049243\\
  0.03	0.0509811791526403\\
  0.031	0.0503430115757031\\
  0.032	0.0464204416186973\\
  0.035	0.0287198071247022\\
  0.036	0.0249879672272587\\
  0.037	0.0230562508182218\\
  0.038	0.0225092728337054\\
  0.04	0.0228312282936897\\
  0.041	0.0225305793048761\\
  0.042	0.0215657008655973\\
  0.043	0.0200132115306983\\
  0.045	0.0162186434434322\\
  0.046	0.0145453052379499\\
  0.047	0.0132502481905479\\
  0.048	0.0123466740562568\\
  0.049	0.0117463335758542\\
  0.052	0.0104269998895171\\
  0.054	0.00917567630989979\\
  0.056	0.00777416472706993\\
  0.0580000000000001	0.00662401192886986\\
  0.0600000000000001	0.00583304567937548\\
  0.0670000000000001	0.00374000936915986\\
  0.0700000000000001	0.00303054381874057\\
  0.074	0.00236138775153771\\
  0.079	0.00168065840400078\\
  0.084	0.00120992136308362\\
  0.091	0.00075774910964066\\
  0.099	0.000445820652911988\\
  0.109	0.000228534918591983\\
  0.124	8.3001290269058e-05\\
  0.152	1.09934924219512e-05\\
  0.267	-2.30508905652016e-06\\
  0.297	-2.30211013318993e-06\\
};
\addlegendentry{$\rep{\Omega_{\star/\fraki,3}}{\star}$}

\end{axis}
\end{tikzpicture}%

%% file: TikZ/AsyCoSPM_1_dot_CHI_reel_inertiel_perm.tex
%
%
\definecolor{mycolor1}{rgb}{0.00000,0.44700,0.74100}%
\definecolor{mycolor2}{rgb}{0.85000,0.32500,0.09800}%
\definecolor{mycolor3}{rgb}{0.92900,0.69400,0.12500}%
\begin{tikzpicture}

\begin{axis}[%
width=4.521in,
height=2.5in,
at={(0.758in,0.481in)},
scale only axis,
xmin=0.25,
xmax=30,
xlabel style={font=\color{white!15!black}},
xlabel={$t$ $\mathrm{[s]}$},
ymin=-4e-06,
ymax=4e-06,
ylabel style={font=\color{white!15!black}},
ylabel={$\rep{\bm{\Omega}_{\star/\fraki}}{\star}$ $\mathrm{[rad/s]}$},
axis background/.style={fill=white},
xmajorgrids,
ymajorgrids,
legend style={legend cell align=left, align=left, draw=white!15!black}
]
\addplot [color=mycolor1]
  table[row sep=crcr]{%
  0.196999999999999	3.55300741716746e-07\\
  0.219000000000001	3.97938812568555e-07\\
  0.273	5.10797775632454e-07\\
  0.408999999999999	7.94196072462228e-07\\
  0.513000000000002	1.00478304787543e-06\\
  0.606000000000002	1.18709607122014e-06\\
  0.692	1.34964050957365e-06\\
  0.773	1.49665146409461e-06\\
  0.850000000000001	1.63031911526446e-06\\
  0.923999999999999	1.75269276070367e-06\\
  0.995000000000001	1.86409222990847e-06\\
  1.065	1.96781465788831e-06\\
  1.133	2.06248165213196e-06\\
  1.2	2.14964757105918e-06\\
  1.266	2.22939006633283e-06\\
  1.331	2.30183124827477e-06\\
  1.395	2.36713351142726e-06\\
  1.459	2.42637382186217e-06\\
  1.523	2.47949902387745e-06\\
  1.587	2.52648685261647e-06\\
  1.651	2.56734602288589e-06\\
  1.715	2.60211608349437e-06\\
  1.78	2.63126897337429e-06\\
  1.845	2.6543178250904e-06\\
  1.911	2.67161478717526e-06\\
  1.978	2.68305654316237e-06\\
  2.046	2.68858297758356e-06\\
  2.116	2.68813118253775e-06\\
  2.188	2.68148498605569e-06\\
  2.262	2.66849914964951e-06\\
  2.339	2.64881330380717e-06\\
  2.419	2.62222513569554e-06\\
  2.504	2.58777640738117e-06\\
  2.594	2.54513523856303e-06\\
  2.692	2.4925026664846e-06\\
  2.801	2.42771620762028e-06\\
  2.928	2.34590844172544e-06\\
  3.093	2.23309049474096e-06\\
  3.568	1.90474158046072e-06\\
  3.718	1.80932655524657e-06\\
  3.854	1.7288518137093e-06\\
  3.984	1.65801678519983e-06\\
  4.111	1.59492762108471e-06\\
  4.238	1.5379790134773e-06\\
  4.367	1.48628035390175e-06\\
  4.501	1.43872908964227e-06\\
  4.645	1.39382640540475e-06\\
  4.809	1.3489856023341e-06\\
  5.035	1.29392811842877e-06\\
  5.339	1.21921246076795e-06\\
  5.486	1.1766781398137e-06\\
  5.608	1.13541211632651e-06\\
  5.717	1.09253964808431e-06\\
  5.817	1.04722569815863e-06\\
  5.912	9.98117421602274e-07\\
  6.002	9.45568142896036e-07\\
  6.089	8.88749269734035e-07\\
  6.174	8.27174197581826e-07\\
  6.257	7.61009097516308e-07\\
  6.339	6.8960125432227e-07\\
  6.421	6.12077432293745e-07\\
  6.503	5.28390071252716e-07\\
  6.585	4.38555648685224e-07\\
  6.667	3.42653137863635e-07\\
  6.751	2.38265052132647e-07\\
  6.836	1.26515956822004e-07\\
  6.923	6.05310646051294e-09\\
  7.013	-1.24663010581116e-07\\
  7.107	-2.6732191926726e-07\\
  7.206	-4.23711572494767e-07\\
  7.312	-5.97309210093044e-07\\
  7.429	-7.95136916309502e-07\\
  7.564	-1.02969020332466e-06\\
  7.744	-1.34903618231874e-06\\
  8.089	-1.9621109323964e-06\\
  8.222	-2.19147493751848e-06\\
  8.335	-2.3803575039949e-06\\
  8.436	-2.54317892967038e-06\\
  8.529	-2.68710697781671e-06\\
  8.616	-2.81577005978306e-06\\
  8.699	-2.93248675831137e-06\\
  8.778	-3.03756521091714e-06\\
  8.854	-3.13264903084587e-06\\
  8.928	-3.21915571177556e-06\\
  9	-3.29719616587454e-06\\
  9.07	-3.36693596025839e-06\\
  9.138	-3.42858977830929e-06\\
  9.204	-3.48241558612017e-06\\
  9.269	-3.52938536352099e-06\\
  9.333	-3.56954394575837e-06\\
  9.396	-3.60295878110151e-06\\
  9.458	-3.62971852752025e-06\\
  9.519	-3.64993164936323e-06\\
  9.579	-3.66372510640645e-06\\
  9.638	-3.67124381028816e-06\\
  9.697	-3.67262027722859e-06\\
  9.755	-3.66785048555585e-06\\
  9.812	-3.65712530836504e-06\\
  9.869	-3.64030208288568e-06\\
  9.925	-3.61773480506145e-06\\
  9.981	-3.5890918006487e-06\\
  10.037	-3.55429356346804e-06\\
  10.092	-3.51405919829517e-06\\
  10.147	-3.46776835868923e-06\\
  10.202	-3.41538081016779e-06\\
  10.257	-3.35687036923105e-06\\
  10.312	-3.29222644523952e-06\\
  10.367	-3.22145509912275e-06\\
  10.422	-3.14458040051591e-06\\
  10.477	-3.06164612950965e-06\\
  10.533	-2.9710448288256e-06\\
  10.589	-2.87431964096641e-06\\
  10.646	-2.76969834089869e-06\\
  10.703	-2.65900267137908e-06\\
  10.761	-2.54031409241406e-06\\
  10.82	-2.41354202401567e-06\\
  10.881	-2.27635077365562e-06\\
  10.944	-2.12847632852231e-06\\
  11.009	-1.96974839283826e-06\\
  11.077	-1.79753491380552e-06\\
  11.149	-1.60901088008814e-06\\
  11.227	-1.39852309288813e-06\\
  11.313	-1.16020510532167e-06\\
  11.415	-8.71190199092098e-07\\
  11.565	-4.39300404053711e-07\\
  11.748	8.64033502523398e-08\\
  11.848	3.67387809063757e-07\\
  11.932	5.97394478774049e-07\\
  12.007	7.96771036704058e-07\\
  12.077	9.76805210939347e-07\\
  12.143	1.14047087507174e-06\\
  12.206	1.29061159270805e-06\\
  12.267	1.42984670503665e-06\\
  12.326	1.55837913951018e-06\\
  12.383	1.67650314608636e-06\\
  12.439	1.78649796822583e-06\\
  12.494	1.88846488313743e-06\\
  12.548	1.98255087369148e-06\\
  12.602	2.07051531475599e-06\\
  12.655	2.15077982801404e-06\\
  12.708	2.2249381288475e-06\\
  12.761	2.29292276898718e-06\\
  12.813	2.35358184141887e-06\\
  12.865	2.40823487729358e-06\\
  12.917	2.4568814858128e-06\\
  12.97	2.50030266712997e-06\\
  13.023	2.53754423695796e-06\\
  13.076	2.56866524139809e-06\\
  13.129	2.59374288091863e-06\\
  13.183	2.61317549643536e-06\\
  13.237	2.62654836902243e-06\\
  13.292	2.63407632061785e-06\\
  13.348	2.63558321833557e-06\\
  13.405	2.63091970964524e-06\\
  13.463	2.6199652403136e-06\\
  13.522	2.60262978457604e-06\\
  13.582	2.57885559307169e-06\\
  13.643	2.54861863879796e-06\\
  13.706	2.51129070250045e-06\\
  13.771	2.46664414760289e-06\\
  13.838	2.41451191129727e-06\\
  13.908	2.3538855948857e-06\\
  13.981	2.28449305339495e-06\\
  14.058	2.20510274573371e-06\\
  14.139	2.1154588267791e-06\\
  14.227	2.01185519088654e-06\\
  14.323	1.89261186989143e-06\\
  14.432	1.75095096466293e-06\\
  14.564	1.57303694336974e-06\\
  14.782	1.27201038324642e-06\\
  14.989	9.88442419469493e-07\\
  15.121	8.13684160050343e-07\\
  15.234	6.7003040982172e-07\\
  15.337	5.45067763368934e-07\\
  15.434	4.33423601009508e-07\\
  15.526	3.33544928565743e-07\\
  15.615	2.42938394734438e-07\\
  15.702	1.60420388084503e-07\\
  15.788	8.49807619829335e-08\\
  15.872	1.73478937881555e-08\\
  15.956	-4.42004868261847e-08\\
  16.04	-9.96221700688693e-08\\
  16.124	-1.48932901566923e-07\\
  16.209	-1.92682239941178e-07\\
  16.295	-2.30780859311608e-07\\
  16.382	-2.63197836858353e-07\\
  16.471	-2.90231348287762e-07\\
  16.562	-3.1177101789126e-07\\
  16.656	-3.27920439247009e-07\\
  16.754	-3.38615137707166e-07\\
  16.856	-3.4363494805234e-07\\
  16.964	-3.42814406195657e-07\\
  17.079	-3.35797768258317e-07\\
  17.204	-3.21970617278566e-07\\
  17.342	-3.00461060476209e-07\\
  17.499	-2.69719137691027e-07\\
  17.691	-2.25719670510216e-07\\
  18.002	-1.47286534968316e-07\\
  18.311	-7.14613186403312e-08\\
  18.502	-3.07153769085744e-08\\
  18.665	-1.93753990629375e-09\\
  18.814	1.83223356486906e-08\\
  18.954	3.13227666026705e-08\\
  19.09	3.78638844722445e-08\\
  19.224	3.818657745569e-08\\
  19.358	3.23995763551466e-08\\
  19.495	2.03738679260823e-08\\
  19.639	1.59097623964044e-09\\
  19.797	-2.52502587727577e-08\\
  19.985	-6.35643928603713e-08\\
  20.593	-1.92541406818236e-07\\
  20.732	-2.12703991309127e-07\\
  20.855	-2.24551293115383e-07\\
  20.969	-2.29473130275437e-07\\
  21.076	-2.28047380090857e-07\\
  21.179	-2.20594518651751e-07\\
  21.279	-2.0725117622078e-07\\
  21.377	-1.88062262651556e-07\\
  21.474	-1.62952535021077e-07\\
  21.571	-1.31713612461226e-07\\
  21.669	-9.40068147770035e-08\\
  21.769	-4.93840666138112e-08\\
  21.873	3.20354587302063e-09\\
  21.983	6.50125002721325e-08\\
  22.104	1.39227985584967e-07\\
  22.249	2.34559372813692e-07\\
  22.72	5.49397647375827e-07\\
  22.825	6.10288765301448e-07\\
  22.918	6.58220155713707e-07\\
  23.003	6.95997684374561e-07\\
  23.082	7.25120756328579e-07\\
  23.157	7.46785385530302e-07\\
  23.229	7.615800470262e-07\\
  23.299	7.69897930297248e-07\\
  23.367	7.71902985974293e-07\\
  23.434	7.67737649454148e-07\\
  23.499	7.57649814886463e-07\\
  23.564	7.41434522666395e-07\\
  23.628	7.19359089629279e-07\\
  23.692	6.91143004161177e-07\\
  23.756	6.56751769412267e-07\\
  23.82	6.16194423486149e-07\\
  23.884	5.6952327653903e-07\\
  23.949	5.15962749858545e-07\\
  24.014	4.56336483267705e-07\\
  24.081	3.88722678934528e-07\\
  24.149	3.13964129361466e-07\\
  24.219	2.30867126305156e-07\\
  24.291	1.39287653411202e-07\\
  24.366	3.77878777158003e-08\\
  24.445	-7.525780887363e-08\\
  24.529	-2.01592598614297e-07\\
  24.621	-3.46185569100044e-07\\
  24.724	-5.14333571288716e-07\\
  24.848	-7.23142672853783e-07\\
  25.048	-1.06710274394572e-06\\
  25.241	-1.3968662813113e-06\\
  25.361	-1.59579975900215e-06\\
  25.463	-1.75886374620404e-06\\
  25.554	-1.89834728558935e-06\\
  25.638	-2.02114134495446e-06\\
  25.718	-2.13202024568204e-06\\
  25.794	-2.23125580944838e-06\\
  25.867	-2.32045574222184e-06\\
  25.937	-2.39991352657398e-06\\
  26.005	-2.47101478834111e-06\\
  26.071	-2.53394507865323e-06\\
  26.135	-2.58894561255829e-06\\
  26.198	-2.63702088432183e-06\\
  26.26	-2.67820781729711e-06\\
  26.321	-2.71257248840584e-06\\
  26.381	-2.74020878876513e-06\\
  26.44	-2.76123617837243e-06\\
  26.498	-2.77579801988281e-06\\
  26.555	-2.78406006160026e-06\\
  26.612	-2.78619208771147e-06\\
  26.668	-2.78219962623893e-06\\
  26.724	-2.77206974885758e-06\\
  26.779	-2.75605849608951e-06\\
  26.834	-2.73396494776534e-06\\
  26.889	-2.70572807536951e-06\\
  26.944	-2.67130034714569e-06\\
  26.998	-2.63144356793532e-06\\
  27.052	-2.5855701366595e-06\\
  27.106	-2.53367779379232e-06\\
  27.16	-2.4757797874031e-06\\
  27.215	-2.41066634387721e-06\\
  27.27	-2.33940156846302e-06\\
  27.325	-2.26205266784518e-06\\
  27.381	-2.17713533245956e-06\\
  27.437	-2.08611168162065e-06\\
  27.494	-1.98732923806233e-06\\
  27.552	-1.88063058814691e-06\\
  27.611	-1.7658959450273e-06\\
  27.671	-1.64304871930199e-06\\
  27.732	-1.51206116782987e-06\\
  27.795	-1.37066906802374e-06\\
  27.86	-1.21868735192265e-06\\
  27.928	-1.05355735158241e-06\\
  28	-8.72504791971096e-07\\
  28.076	-6.75256487170373e-07\\
  28.159	-4.53697634128503e-07\\
  28.254	-1.93812258686421e-07\\
  28.372	1.3547854749163e-07\\
  28.737	1.15854153648343e-06\\
  28.831	1.4131262275896e-06\\
  28.913	1.62925076807596e-06\\
  28.989	1.82354216349268e-06\\
  29.06	1.99903664110934e-06\\
  29.128	2.16105058825633e-06\\
  29.193	2.30987436822261e-06\\
  29.256	2.44807537796987e-06\\
  29.317	2.57588572338818e-06\\
  29.377	2.6955559278008e-06\\
  29.436	2.80714601430532e-06\\
  29.494	2.9107607950607e-06\\
  29.551	3.00654398444067e-06\\
  29.608	3.09619198546329e-06\\
  29.664	3.17817554318367e-06\\
  29.72	3.25402530521046e-06\\
  29.775	3.32247589085455e-06\\
  29.83	3.38487906503815e-06\\
  29.885	3.44119327877479e-06\\
  29.94	3.49139290278799e-06\\
  29.995	3.53546563403029e-06\\
  30	3.5391684711783e-06\\
};
\addlegendentry{$\rep{\Omega_{\star/\fraki,1}}{\star}$}

\addplot [color=mycolor2]
  table[row sep=crcr]{%
  0.196999999999999	-2.15350971188855e-06\\
  0.202000000000002	-2.1360192050679e-06\\
  0.207999999999998	-2.12085893736003e-06\\
  0.216000000000001	-2.10716075699224e-06\\
  0.227	-2.09536280948441e-06\\
  0.244	-2.08444894411741e-06\\
  0.285	-2.06668764590745e-06\\
  0.353999999999999	-2.03483727645448e-06\\
  0.417999999999999	-1.99904913600335e-06\\
  0.481999999999999	-1.95721234064195e-06\\
  0.547000000000001	-1.90865084803704e-06\\
  0.613	-1.85322411994093e-06\\
  0.68	-1.79081685303117e-06\\
  0.748000000000001	-1.72134253517697e-06\\
  0.817	-1.6447477584336e-06\\
  0.888000000000002	-1.55977820526232e-06\\
  0.960999999999999	-1.46621306740258e-06\\
  1.036	-1.36390016791665e-06\\
  1.114	-1.2512881895077e-06\\
  1.195	-1.12818677777682e-06\\
  1.281	-9.91306116304713e-07\\
  1.374	-8.37040015966295e-07\\
  1.477	-6.59928385715602e-07\\
  1.6	-4.42031666381126e-07\\
  1.795	-8.94532838913165e-08\\
  1.977	2.37411530434883e-07\\
  2.09	4.34243279556767e-07\\
  2.184	5.9204046465311e-07\\
  2.269	7.28709071751155e-07\\
  2.347	8.48105422335266e-07\\
  2.42	9.5387365206534e-07\\
  2.49	1.04925426924751e-06\\
  2.557	1.13449998195847e-06\\
  2.622	1.21112065798457e-06\\
  2.685	1.27930887927619e-06\\
  2.746	1.33933407298059e-06\\
  2.806	1.39236431806466e-06\\
  2.865	1.43848440359307e-06\\
  2.923	1.47782159132248e-06\\
  2.981	1.51106254620004e-06\\
  3.038	1.53767640398428e-06\\
  3.095	1.55820297109699e-06\\
  3.152	1.57258421751294e-06\\
  3.209	1.58079021872481e-06\\
  3.266	1.58281978102082e-06\\
  3.323	1.5787008642576e-06\\
  3.38	1.56849115029445e-06\\
  3.438	1.55194088335975e-06\\
  3.496	1.52929938934676e-06\\
  3.555	1.50017962141646e-06\\
  3.615	1.4644739607661e-06\\
  3.676	1.42212746467862e-06\\
  3.739	1.37230727048632e-06\\
  3.804	1.31481184340032e-06\\
  3.872	1.24852585159374e-06\\
  3.944	1.17212704253689e-06\\
  4.021	1.08417543032147e-06\\
  4.105	9.81996318927258e-07\\
  4.202	8.57666183406991e-07\\
  4.328	6.8959498733534e-07\\
  4.598	3.27758996832017e-07\\
  4.694	2.06430282645442e-07\\
  4.776	1.08792690411974e-07\\
  4.85	2.66824997652293e-08\\
  4.919	-4.38358753740431e-08\\
  4.984	-1.0422938956367e-07\\
  5.046	-1.55822174718878e-07\\
  5.106	-1.99707614001454e-07\\
  5.164	-2.36111056040045e-07\\
  5.221	-2.65816609612557e-07\\
  5.277	-2.88890110056172e-07\\
  5.332	-3.05455351679029e-07\\
  5.386	-3.15687660901176e-07\\
  5.44	-3.19828501460506e-07\\
  5.494	-3.17800292748416e-07\\
  5.547	-3.0977074771954e-07\\
  5.6	-2.95754944090731e-07\\
  5.654	-2.75349957945537e-07\\
  5.708	-2.48832044746905e-07\\
  5.763	-2.15649176737998e-07\\
  5.818	-1.76385249517352e-07\\
  5.874	-1.30353832616947e-07\\
  5.932	-7.64898722138696e-08\\
  5.991	-1.55468988793928e-08\\
  6.052	5.36012336738168e-08\\
  6.115	1.3110290097984e-07\\
  6.182	2.19720032390569e-07\\
  6.253	3.1984235349114e-07\\
  6.33	4.34656190151372e-07\\
  6.416	5.69150774509808e-07\\
  6.519	7.36628493314129e-07\\
  6.675	9.97250356959967e-07\\
  6.855	1.29649898283901e-06\\
  6.958	1.46155228009093e-06\\
  7.046	1.59653707143548e-06\\
  7.125	1.71170188423275e-06\\
  7.199	1.81352120520728e-06\\
  7.269	1.90377627973248e-06\\
  7.336	1.98412536889236e-06\\
  7.401	2.05601694247548e-06\\
  7.464	2.11966784036122e-06\\
  7.526	2.17624056375598e-06\\
  7.587	2.22580136721717e-06\\
  7.647	2.26846527695557e-06\\
  7.706	2.30438867632188e-06\\
  7.765	2.3342171857621e-06\\
  7.823	2.35751162236397e-06\\
  7.881	2.37476429987282e-06\\
  7.939	2.38593185386549e-06\\
  7.997	2.39099011523081e-06\\
  8.055	2.3899331935695e-06\\
  8.113	2.38277188913116e-06\\
  8.171	2.36953336596457e-06\\
  8.229	2.35025931871746e-06\\
  8.288	2.32451799320188e-06\\
  8.347	2.29266144202711e-06\\
  8.406	2.25477522874939e-06\\
  8.466	2.21016426493748e-06\\
  8.527	2.1586403669005e-06\\
  8.588	2.10102976438975e-06\\
  8.65	2.03638867546374e-06\\
  8.713	1.96459105694657e-06\\
  8.777	1.88553601532249e-06\\
  8.842	1.7991509118076e-06\\
  8.909	1.70392751286386e-06\\
  8.977	1.60114719349735e-06\\
  9.047	1.48920729259316e-06\\
  9.119	1.36794132288287e-06\\
  9.194	1.23544774055517e-06\\
  9.272	1.09148385618596e-06\\
  9.354	9.33976462391684e-07\\
  9.441	7.6073646582131e-07\\
  9.536	5.65389431272934e-07\\
  9.643	3.39137045557436e-07\\
  9.775	5.35985904548397e-08\\
  10.196	-8.61899398074684e-07\\
  10.296	-1.07033315543958e-06\\
  10.384	-1.24777381316221e-06\\
  10.464	-1.40310043050818e-06\\
  10.539	-1.54267307550526e-06\\
  10.61	-1.66870360018834e-06\\
  10.677	-1.7816176409724e-06\\
  10.742	-1.8850876841725e-06\\
  10.805	-1.97924356726276e-06\\
  10.866	-2.0642973410645e-06\\
  10.925	-2.14053355307442e-06\\
  10.983	-2.20943421069819e-06\\
  11.04	-2.2710814349125e-06\\
  11.096	-2.32559748170047e-06\\
  11.151	-2.37314142026435e-06\\
  11.206	-2.4146043031692e-06\\
  11.26	-2.44928826376167e-06\\
  11.314	-2.47791697560729e-06\\
  11.368	-2.50042895899583e-06\\
  11.422	-2.51678524065824e-06\\
  11.476	-2.52697009628378e-06\\
  11.53	-2.53099140579138e-06\\
  11.584	-2.52888039398158e-06\\
  11.638	-2.52069120421083e-06\\
  11.693	-2.50618247932266e-06\\
  11.748	-2.48555429038788e-06\\
  11.804	-2.45839743229226e-06\\
  11.861	-2.42454472854092e-06\\
  11.918	-2.38462016355356e-06\\
  11.977	-2.33713400632496e-06\\
  12.037	-2.28269162860784e-06\\
  12.099	-2.2202467597765e-06\\
  12.163	-2.14958220823291e-06\\
  12.229	-2.07056854861776e-06\\
  12.298	-1.9818501684199e-06\\
  12.371	-1.88186823280034e-06\\
  12.45	-1.76746227964486e-06\\
  12.537	-1.63520812535012e-06\\
  12.637	-1.47688347595931e-06\\
  12.767	-1.2645137275058e-06\\
  13.069	-7.6897302747625e-07\\
  13.171	-6.09142805529928e-07\\
  13.259	-4.77225771788881e-07\\
  13.339	-3.63284154758503e-07\\
  13.414	-2.6248761031411e-07\\
  13.485	-1.73081627252714e-07\\
  13.553	-9.34391515272637e-08\\
  13.619	-2.2138340938227e-08\\
  13.684	4.19820160857398e-08\\
  13.748	9.89408945883952e-08\\
  13.81	1.48080232520442e-07\\
  13.872	1.91129078785934e-07\\
  13.933	2.2744056948909e-07\\
  13.994	2.57702588157827e-07\\
  14.055	2.8190019207841e-07\\
  14.116	3.00059458169244e-07\\
  14.178	3.1239736841826e-07\\
  14.24	3.18676931954087e-07\\
  14.303	3.19008602644999e-07\\
  14.367	3.13307229049542e-07\\
  14.433	3.01318980433507e-07\\
  14.501	2.82816042584955e-07\\
  14.571	2.5766066968913e-07\\
  14.644	2.25339793047397e-07\\
  14.721	1.85157446708217e-07\\
  14.804	1.35684970814509e-07\\
  14.895	7.5249324993365e-08\\
  14.999	-4.97131225074554e-11\\
  15.132	-1.02811092972388e-07\\
  15.5	-3.90464119703893e-07\\
  15.609	-4.67413514115833e-07\\
  15.705	-5.29191165554721e-07\\
  15.794	-5.80436893216074e-07\\
  15.878	-6.22789617210628e-07\\
  15.959	-6.57600992326479e-07\\
  16.038	-6.85494708108081e-07\\
  16.116	-7.06919049520138e-07\\
  16.193	-7.21955181859357e-07\\
  16.27	-7.30848153551733e-07\\
  16.347	-7.33607485869925e-07\\
  16.425	-7.3023588953447e-07\\
  16.504	-7.20659397046575e-07\\
  16.584	-7.04886847557873e-07\\
  16.667	-6.82399512896836e-07\\
  16.753	-6.52966019032419e-07\\
  16.843	-6.16053601021349e-07\\
  16.939	-5.70540439781553e-07\\
  17.043	-5.15073349305339e-07\\
  17.159	-4.47019701255158e-07\\
  17.297	-3.5970512257677e-07\\
  17.484	-2.34759966133424e-07\\
  17.899	4.40344294361239e-08\\
  18.066	1.49005764171761e-07\\
  18.219	2.39126379142363e-07\\
  18.364	3.1848150072733e-07\\
  18.507	3.90624101953563e-07\\
  18.649	4.56107589030808e-07\\
  18.791	5.1542443912922e-07\\
  18.931	5.67875080292879e-07\\
  19.073	6.14944713817067e-07\\
  19.213	6.55236441104989e-07\\
  19.351	6.88857372921348e-07\\
  19.485	7.15481697000087e-07\\
  19.615	7.35346322500163e-07\\
  19.742	7.48737686961931e-07\\
  19.865	7.55678879471589e-07\\
  19.984	7.56388971012711e-07\\
  20.101	7.50978383479151e-07\\
  20.215	7.39557968643112e-07\\
  20.327	7.22140132580762e-07\\
  20.437	6.98812705479668e-07\\
  20.545	6.69725107371733e-07\\
  20.651	6.35093872602965e-07\\
  20.758	5.93963832784539e-07\\
  20.865	5.46648887933543e-07\\
  20.973	4.92725369838354e-07\\
  21.083	4.31650455823274e-07\\
  21.196	3.62800633979532e-07\\
  21.315	2.84158456764771e-07\\
  21.443	1.93417257321471e-07\\
  21.589	8.3642756720792e-08\\
  21.782	-6.81423486526e-08\\
  22.132	-3.44153654907586e-07\\
  22.268	-4.44511048414142e-07\\
  22.384	-5.24138290103338e-07\\
  22.488	-5.89530873895683e-07\\
  22.584	-6.43924799703655e-07\\
  22.675	-6.89484423332942e-07\\
  22.762	-7.27024296764966e-07\\
  22.846	-7.57247889282553e-07\\
  22.928	-7.8069456677099e-07\\
  23.008	-7.97528223017707e-07\\
  23.087	-8.08086340953196e-07\\
  23.165	-8.12446216968965e-07\\
  23.243	-8.10680628404725e-07\\
  23.32	-8.02881693573454e-07\\
  23.398	-7.88848648625162e-07\\
  23.476	-7.68695699093769e-07\\
  23.555	-7.42152682420283e-07\\
  23.635	-7.09149542643672e-07\\
  23.717	-6.69157568466971e-07\\
  23.801	-6.22031475217e-07\\
  23.888	-5.67060961742527e-07\\
  23.979	-5.03394709028271e-07\\
  24.076	-4.29317669414786e-07\\
  24.181	-3.42929734387099e-07\\
  24.301	-2.37910239064831e-07\\
  24.454	-9.75241079004263e-08\\
  24.798	2.20129649619594e-07\\
  24.913	3.18749648897665e-07\\
  25.011	3.96700926330595e-07\\
  25.098	4.59906477345839e-07\\
  25.179	5.12719950052087e-07\\
  25.255	5.5622745875894e-07\\
  25.327	5.9143304298459e-07\\
  25.397	6.1953158692063e-07\\
  25.464	6.40330686252355e-07\\
  25.529	6.54443407910321e-07\\
  25.593	6.62184525879184e-07\\
  25.655	6.63595685068685e-07\\
  25.716	6.58914988349579e-07\\
  25.776	6.482564636201e-07\\
  25.835	6.3177609632703e-07\\
  25.894	6.09234312776152e-07\\
  25.953	5.80550985063155e-07\\
  26.012	5.45677259822241e-07\\
  26.07	5.05348715762466e-07\\
  26.129	4.58194943320223e-07\\
  26.188	4.04917788188186e-07\\
  26.247	3.45612889418589e-07\\
  26.307	2.79260518709634e-07\\
  26.368	2.05743621251031e-07\\
  26.431	1.23654903205761e-07\\
  26.495	3.41700676642631e-08\\
  26.561	-6.41430943915111e-08\\
  26.63	-1.72962760558448e-07\\
  26.704	-2.95843037179111e-07\\
  26.784	-4.34930978343573e-07\\
  26.874	-5.97752119801953e-07\\
  26.982	-7.99567295928227e-07\\
  27.188	-1.19230276851567e-06\\
  27.315	-1.4307515705525e-06\\
  27.408	-1.59942295141491e-06\\
  27.488	-1.73855481477858e-06\\
  27.561	-1.8594686324036e-06\\
  27.628	-1.96449006040211e-06\\
  27.692	-2.05880365911071e-06\\
  27.753	-2.1426932583779e-06\\
  27.812	-2.21781585807435e-06\\
  27.869	-2.2844135649791e-06\\
  27.925	-2.34382246233622e-06\\
  27.98	-2.39611598118472e-06\\
  28.034	-2.44142213645659e-06\\
  28.088	-2.48058375618143e-06\\
  28.141	-2.51292062714015e-06\\
  28.194	-2.53912334713391e-06\\
  28.246	-2.55881098709665e-06\\
  28.298	-2.57250392010633e-06\\
  28.351	-2.58028792998743e-06\\
  28.404	-2.58186388180093e-06\\
  28.457	-2.57728144248404e-06\\
  28.51	-2.56661493125421e-06\\
  28.564	-2.54958950662854e-06\\
  28.618	-2.52647410192708e-06\\
  28.673	-2.49682147668295e-06\\
  28.729	-2.46047363461344e-06\\
  28.786	-2.41730554506603e-06\\
  28.844	-2.3672264006791e-06\\
  28.903	-2.31017896012986e-06\\
  28.964	-2.24502499079904e-06\\
  29.026	-2.17270422453453e-06\\
  29.09	-2.09197028766539e-06\\
  29.157	-2.00125531790718e-06\\
  29.226	-1.90167705227395e-06\\
  29.298	-1.79163672342497e-06\\
  29.374	-1.66929179101771e-06\\
  29.454	-1.5343298400694e-06\\
  29.539	-1.38479409983461e-06\\
  29.631	-1.21678383990798e-06\\
  29.733	-1.02426414372303e-06\\
  29.849	-7.990356856169e-07\\
  29.992	-5.14981689292426e-07\\
  30	-4.98947812843653e-07\\
};
\addlegendentry{$\rep{\Omega_{\star/\fraki,2}}{\star}$}

\addplot [color=mycolor3]
  table[row sep=crcr]{%
  0.196999999999999	-1.65385233330539e-06\\
  0.199000000000002	-1.73719667273531e-06\\
  0.201000000000001	-1.81013467326352e-06\\
  0.202999999999999	-1.87395838224802e-06\\
  0.204999999999998	-1.92979948465677e-06\\
  0.207000000000001	-1.97864909168288e-06\\
  0.209	-2.02137511351452e-06\\
  0.210999999999999	-2.05873756442543e-06\\
  0.213000000000001	-2.09140214124659e-06\\
  0.216000000000001	-2.13284582528672e-06\\
  0.219000000000001	-2.16668585650837e-06\\
  0.222000000000001	-2.19429229630919e-06\\
  0.225999999999999	-2.22331506805062e-06\\
  0.23	-2.24534743864524e-06\\
  0.234999999999999	-2.2654956772783e-06\\
  0.239999999999998	-2.27959925069854e-06\\
  0.245999999999999	-2.29090094450157e-06\\
  0.254000000000001	-2.29963847431236e-06\\
  0.265000000000001	-2.30470774553737e-06\\
  0.280999999999999	-2.30512401699912e-06\\
  0.309999999999999	-2.29862186884588e-06\\
  0.372	-2.27760604332161e-06\\
  0.451000000000001	-2.24497468792606e-06\\
  0.532	-2.20546859353021e-06\\
  0.616	-2.15834141670257e-06\\
  0.702000000000002	-2.10396437339e-06\\
  0.791	-2.04160251016106e-06\\
  0.884	-1.97034898619108e-06\\
  0.983000000000001	-1.88830814806806e-06\\
  1.088	-1.79510598030674e-06\\
  1.201	-1.68864493588217e-06\\
  1.326	-1.56464921019506e-06\\
  1.466	-1.41953408316908e-06\\
  1.628	-1.24534399148502e-06\\
  1.82	-1.0325830963609e-06\\
  2.04	-7.8258394253794e-07\\
  2.261	-5.25439464382771e-07\\
  2.459	-2.89112318085927e-07\\
  2.637	-7.06582738985162e-08\\
  2.803	1.3912014651396e-07\\
  2.965	3.4995762376866e-07\\
  3.131	5.72171586554759e-07\\
  3.316	8.26099803674651e-07\\
  3.616	1.24518832222975e-06\\
  3.83	1.5411641669516e-06\\
  3.972	1.73153661719994e-06\\
  4.091	1.88506546194844e-06\\
  4.197	2.01581427816677e-06\\
  4.295	2.13065414556013e-06\\
  4.387	2.23241165286936e-06\\
  4.475	2.32364665819773e-06\\
  4.559	2.40467999645944e-06\\
  4.641	2.47767095729046e-06\\
  4.72	2.54194616644554e-06\\
  4.798	2.59930552459764e-06\\
  4.874	2.64912707592657e-06\\
  4.949	2.69222284998705e-06\\
  5.023	2.72866306616493e-06\\
  5.096	2.75854459275138e-06\\
  5.168	2.78198546865838e-06\\
  5.24	2.79931845525994e-06\\
  5.311	2.81032142623872e-06\\
  5.381	2.81514972400032e-06\\
  5.451	2.81390457956832e-06\\
  5.52	2.80664054486124e-06\\
  5.589	2.79329192665045e-06\\
  5.657	2.774095136715e-06\\
  5.725	2.74881411854722e-06\\
  5.792	2.71787226324705e-06\\
  5.859	2.68086482435592e-06\\
  5.926	2.63772065878243e-06\\
  5.992	2.58915960671402e-06\\
  6.058	2.53453433884943e-06\\
  6.124	2.47381092677301e-06\\
  6.19	2.40697308839799e-06\\
  6.256	2.33402602134447e-06\\
  6.322	2.25499979933375e-06\\
  6.388	2.16995286805854e-06\\
  6.455	2.07755080694483e-06\\
  6.523	1.97765246312542e-06\\
  6.592	1.87016517827487e-06\\
  6.662	1.75505336486026e-06\\
  6.734	1.63057712043724e-06\\
  6.809	1.49475155808432e-06\\
  6.887	1.34732285772543e-06\\
  6.969	1.18619800915098e-06\\
  7.058	1.00508336586813e-06\\
  7.156	7.9939712804844e-07\\
  7.271	5.51683054794694e-07\\
  7.432	1.9813908735955e-07\\
  7.681	-3.48483801815291e-07\\
  7.799	-6.01009212175541e-07\\
  7.9	-8.11149021728852e-07\\
  7.991	-9.94503412954373e-07\\
  8.077	-1.1617169413114e-06\\
  8.158	-1.31319880480874e-06\\
  8.237	-1.45484885649694e-06\\
  8.313	-1.5850825718644e-06\\
  8.388	-1.70751929218227e-06\\
  8.462	-1.82218667177381e-06\\
  8.535	-1.92917314478791e-06\\
  8.607	-2.02861767917284e-06\\
  8.679	-2.12195526216874e-06\\
  8.751	-2.2091460998297e-06\\
  8.823	-2.29017954467281e-06\\
  8.895	-2.36506994966135e-06\\
  8.967	-2.43385174414357e-06\\
  9.039	-2.49657635009726e-06\\
  9.112	-2.55405365479078e-06\\
  9.185	-2.60544576846655e-06\\
  9.259	-2.65141387245649e-06\\
  9.333	-2.6912988957406e-06\\
  9.408	-2.72560300373925e-06\\
  9.483	-2.75383059289425e-06\\
  9.559	-2.77631649581167e-06\\
  9.635	-2.79271651137947e-06\\
  9.711	-2.80309473765783e-06\\
  9.788	-2.80752332315615e-06\\
  9.865	-2.80587232026619e-06\\
  9.943	-2.79803983715965e-06\\
  10.02	-2.78425797617388e-06\\
  10.098	-2.76419523714821e-06\\
  10.176	-2.73801498806847e-06\\
  10.254	-2.70574276939328e-06\\
  10.332	-2.66740640952889e-06\\
  10.41	-2.62304109099887e-06\\
  10.489	-2.57200613162922e-06\\
  10.568	-2.51488884117634e-06\\
  10.648	-2.45092319772766e-06\\
  10.728	-2.38088388826441e-06\\
  10.809	-2.30388976518725e-06\\
  10.891	-2.21984575077272e-06\\
  10.974	-2.12868120996745e-06\\
  11.059	-2.02914556624023e-06\\
  11.145	-1.92228857898158e-06\\
  11.233	-1.80678019034985e-06\\
  11.323	-1.68246469200994e-06\\
  11.415	-1.54922933504054e-06\\
  11.509	-1.4070077511974e-06\\
  11.606	-1.25418085872298e-06\\
  11.707	-1.088955659867e-06\\
  11.813	-9.0940891794844e-07\\
  11.925	-7.13548011788134e-07\\
  12.046	-4.95756513174683e-07\\
  12.181	-2.46513042867491e-07\\
  12.345	6.27144380871414e-08\\
  12.873	1.06333606098019e-06\\
  12.993	1.2815559387036e-06\\
  13.099	1.46832742942138e-06\\
  13.196	1.63319538160067e-06\\
  13.286	1.78014748897226e-06\\
  13.371	1.91294073559334e-06\\
  13.452	2.03351521577133e-06\\
  13.531	2.14502915341086e-06\\
  13.607	2.24623998690276e-06\\
  13.681	2.3387280485565e-06\\
  13.753	2.42268719219396e-06\\
  13.824	2.49940394780879e-06\\
  13.894	2.5689146561092e-06\\
  13.963	2.63129001609741e-06\\
  14.031	2.68663569613636e-06\\
  14.098	2.73508635828534e-06\\
  14.164	2.77680528881774e-06\\
  14.23	2.81247408651097e-06\\
  14.296	2.84202680589374e-06\\
  14.362	2.86541784078054e-06\\
  14.427	2.88240717694066e-06\\
  14.493	2.89351269344706e-06\\
  14.559	2.89844219736324e-06\\
  14.625	2.897233560617e-06\\
  14.691	2.88994542785304e-06\\
  14.758	2.87641103113856e-06\\
  14.825	2.85680063782934e-06\\
  14.893	2.83081751462078e-06\\
  14.962	2.79835225569514e-06\\
  15.032	2.75932998405892e-06\\
  15.103	2.71371425242251e-06\\
  15.176	2.66074077615031e-06\\
  15.252	2.59936918567405e-06\\
  15.329	2.5310877163065e-06\\
  15.409	2.45406983623297e-06\\
  15.493	2.3670508539908e-06\\
  15.581	2.26974228212384e-06\\
  15.675	2.15960080396371e-06\\
  15.776	2.03504595930326e-06\\
  15.887	1.89192131472282e-06\\
  16.013	1.72314610935587e-06\\
  16.163	1.51586003127591e-06\\
  16.369	1.22453848661053e-06\\
  16.97	3.71534454757239e-07\\
  17.231	9.35695965154082e-09\\
  17.51	-3.71577403512902e-07\\
  17.789	-7.46534677631416e-07\\
  18.022	-1.05381522530479e-06\\
  18.208	-1.29318537389622e-06\\
  18.364	-1.48797721166716e-06\\
  18.501	-1.65301643306748e-06\\
  18.623	-1.79404166189556e-06\\
  18.736	-1.9186958475359e-06\\
  18.842	-2.02960400130792e-06\\
  18.942	-2.128213875352e-06\\
  19.037	-2.21591397675525e-06\\
  19.129	-2.29480607671917e-06\\
  19.218	-2.36505493589334e-06\\
  19.304	-2.42691087137814e-06\\
  19.389	-2.48192201368624e-06\\
  19.472	-2.52950203716296e-06\\
  19.553	-2.56987381419549e-06\\
  19.633	-2.60369334981192e-06\\
  19.712	-2.63104548636761e-06\\
  19.791	-2.65228105433835e-06\\
  19.869	-2.66716984853588e-06\\
  19.947	-2.67597347658466e-06\\
  20.025	-2.67867450176595e-06\\
  20.103	-2.67528782771365e-06\\
  20.182	-2.66569734463928e-06\\
  20.261	-2.64998471521949e-06\\
  20.341	-2.62794419114698e-06\\
  20.422	-2.59949018044381e-06\\
  20.504	-2.56457599689952e-06\\
  20.588	-2.52266188027761e-06\\
  20.674	-2.47357284521854e-06\\
  20.762	-2.41719221349967e-06\\
  20.853	-2.35272288406918e-06\\
  20.947	-2.27998041779642e-06\\
  21.045	-2.1979961211116e-06\\
  21.147	-2.10657914223589e-06\\
  21.255	-2.00368248215455e-06\\
  21.371	-1.88694454905658e-06\\
  21.494	-1.75698582793871e-06\\
  21.627	-1.61023971045893e-06\\
  21.769	-1.44735199825163e-06\\
  21.919	-1.26910218156695e-06\\
  22.072	-1.08120670816447e-06\\
  22.225	-8.87245676040038e-07\\
  22.374	-6.92272536184646e-07\\
  22.517	-4.99090820227366e-07\\
  22.656	-3.05204761019695e-07\\
  22.792	-1.0937577243908e-07\\
  22.928	9.26004695145366e-08\\
  23.067	3.05150098967033e-07\\
  23.218	5.42275657267055e-07\\
  23.399	8.32921887905513e-07\\
  23.837	1.53903259914046e-06\\
  23.963	1.73414463233712e-06\\
  24.07	1.89387226612325e-06\\
  24.167	2.03267519793826e-06\\
  24.257	2.15542822701309e-06\\
  24.342	2.26530067237718e-06\\
  24.423	2.3639413058163e-06\\
  24.501	2.45286241806753e-06\\
  24.576	2.53236427383285e-06\\
  24.65	2.60471596291723e-06\\
  24.722	2.66903637324845e-06\\
  24.793	2.72637038989387e-06\\
  24.863	2.77678767446332e-06\\
  24.932	2.82039826160485e-06\\
  25	2.85734555660611e-06\\
  25.068	2.88821158989094e-06\\
  25.136	2.9129252574478e-06\\
  25.204	2.93143512664074e-06\\
  25.271	2.94357484875718e-06\\
  25.338	2.94964867819658e-06\\
  25.405	2.94965833802507e-06\\
  25.472	2.94362066100007e-06\\
  25.539	2.93156548281104e-06\\
  25.607	2.91321847001313e-06\\
  25.675	2.88876984910758e-06\\
  25.743	2.85828512858188e-06\\
  25.812	2.82125883899198e-06\\
  25.881	2.77818386251738e-06\\
  25.951	2.72840273041197e-06\\
  26.022	2.67176986668005e-06\\
  26.093	2.60907982152503e-06\\
  26.166	2.53845464470714e-06\\
  26.239	2.46173795304117e-06\\
  26.313	2.37793539881181e-06\\
  26.389	2.28575342475779e-06\\
  26.466	2.18627902981439e-06\\
  26.545	2.07813849328886e-06\\
  26.626	1.96118638129406e-06\\
  26.71	1.83379049900623e-06\\
  26.797	1.69574331110312e-06\\
  26.889	1.54357854142972e-06\\
  26.986	1.3769737812197e-06\\
  27.091	1.19042753965459e-06\\
  27.207	9.78138739782253e-07\\
  27.344	7.21098675882104e-07\\
  27.536	3.54159471527282e-07\\
  27.84	-2.26654329082976e-07\\
  27.983	-4.93281593350048e-07\\
  28.106	-7.16591237903685e-07\\
  28.217	-9.12120288631968e-07\\
  28.321	-1.0893331179318e-06\\
  28.421	-1.25364771363934e-06\\
  28.517	-1.40529801129219e-06\\
  28.61	-1.54613356428968e-06\\
  28.7	-1.67642478743346e-06\\
  28.789	-1.79920350973362e-06\\
  28.877	-1.91444939545704e-06\\
  28.963	-2.02097454504724e-06\\
  29.048	-2.12017119238794e-06\\
  29.133	-2.21316622628365e-06\\
  29.216	-2.29786326855219e-06\\
  29.298	-2.37549988213459e-06\\
  29.38	-2.44703351626185e-06\\
  29.461	-2.51161171860304e-06\\
  29.541	-2.56937810405589e-06\\
  29.621	-2.62109047710624e-06\\
  29.7	-2.66614522104192e-06\\
  29.779	-2.70515957367934e-06\\
  29.858	-2.73806445960645e-06\\
  29.936	-2.76449451774852e-06\\
  30	-2.78163732048142e-06\\
};
\addlegendentry{$\rep{\Omega_{\star/\fraki,3}}{\star}$}

\end{axis}
\end{tikzpicture}%

%% file: TikZ/AsyCoSPM_1_CHI_reel_inertiel_trans.tex
%
%
\definecolor{mycolor1}{rgb}{0.00000,0.44700,0.74100}%
\definecolor{mycolor2}{rgb}{0.85000,0.32500,0.09800}%
\definecolor{mycolor3}{rgb}{0.92900,0.69400,0.12500}%
\begin{tikzpicture}

\begin{axis}[%
width=4.521in,
height=2.5in,
at={(0.758in,0.481in)},
scale only axis,
xmin=0,
xmax=0.25,
xtick={0,0.05,0.1,0.15,0.2,0.25},
xlabel style={font=\color{white!15!black}},
xlabel={$t$ $\mathrm{[s]}$},
ymin=-0.0005,
ymax=0.0035,
ylabel style={font=\color{white!15!black}},
ylabel={$\bm{\varepsilon}$ $\mathrm{[rad]}$},
axis background/.style={fill=white},
xmajorgrids,
ymajorgrids,
legend style={legend cell align=left, align=left, draw=white!15!black}
]
\addplot [color=mycolor1]
  table[row sep=crcr]{%
  0	0\\
  0.014	1.47805650324573e-06\\
  0.02	3.13652841998913e-06\\
  0.056	5.68653985127954e-06\\
  0.299	5.56129503204161e-06\\
};
\addlegendentry{$\varepsilon_1$}

\addplot [color=mycolor2]
  table[row sep=crcr]{%
  0	0\\
  0.002	0\\
  0.005	-0.00024673994439478\\
  0.00600000000000001	-0.000298282281261997\\
  0.00700000000000001	-0.000314799059291826\\
  0.00800000000000001	-0.000298546218192919\\
  0.00900000000000001	-0.000261448232750094\\
  0.01	-0.000217548783297772\\
  0.011	-0.000179155571313594\\
  0.012	-0.000153534185737325\\
  0.013	-0.000142250517503995\\
  0.014	-0.000142201702770239\\
  0.016	-0.000153252167823692\\
  0.017	-0.000154411846041147\\
  0.018	-0.000149535509418375\\
  0.019	-0.000139220752742808\\
  0.022	-9.89772152732438e-05\\
  0.023	-8.95188671070635e-05\\
  0.024	-8.3274763683916e-05\\
  0.026	-7.71880984827278e-05\\
  0.028	-7.2221030286479e-05\\
  0.03	-6.37956578786181e-05\\
  0.033	-4.91775140838424e-05\\
  0.035	-4.23974742140842e-05\\
  0.039	-3.41293494975359e-05\\
  0.046	-2.08207450352327e-05\\
  0.053	-1.3287862107636e-05\\
  0.06	-8.28491765830774e-06\\
  0.071	-3.92630530937632e-06\\
  0.087	-1.25414559570824e-06\\
  0.117	1.44358476839024e-08\\
  0.256	4.9721142841852e-07\\
  0.299	5.86239476796369e-07\\
};
\addlegendentry{$\varepsilon_2$}

\addplot [color=mycolor3]
  table[row sep=crcr]{%
  0	0\\
  0.002	0\\
  0.003	0.000356844402978285\\
  0.004	0.000962556932788905\\
  0.005	0.00170104373257046\\
  0.00600000000000001	0.00237714958934043\\
  0.00700000000000001	0.00284616549321931\\
  0.00800000000000001	0.00303758977760776\\
  0.00850000000000001	0.00303758977760776\\ %
  0.00900000000000001	0.00296714275958498\\
  0.01	0.00271262939390293\\
  0.011	0.00237896644260854\\
  0.012	0.00206242885186853\\
  0.013	0.00182635888897692\\
  0.014	0.0016922047339975\\
  0.015	0.00164491307968517\\
  0.016	0.00164742180840555\\
  0.017	0.00165773971263489\\
  0.018	0.0016429524814594\\
  0.019	0.00158685261196112\\
  0.02	0.00149060300320902\\
  0.021	0.00136804494823223\\
  0.022	0.00123842231175197\\
  0.023	0.00111934213347448\\
  0.024	0.00102199668007791\\
  0.025	0.000949484027041259\\
  0.026	0.000897934175651094\\
  0.027	0.000859393937975628\\
  0.028	0.000825172167002119\\
  0.029	0.000788542925296731\\
  0.03	0.000746175336077293\\
  0.031	0.000698195668172363\\
  0.033	0.000596871477444028\\
  0.034	0.00055045103582535\\
  0.035	0.000509959148051187\\
  0.036	0.000475815780263733\\
  0.037	0.000447095973139078\\
  0.038	0.000422108005911803\\
  0.042	0.000331048736174844\\
  0.044	0.000286952456004352\\
  0.045	0.000266939244473685\\
  0.046	0.000248811748514877\\
  0.047	0.000232593105071455\\
  0.049	0.000204797551642932\\
  0.051	0.000180704544010812\\
  0.054	0.000148064622001898\\
  0.056	0.000129042714823557\\
  0.058	0.000112803487095747\\
  0.061	9.28367563514665e-05\\
  0.064	7.62518894437125e-05\\
  0.067	6.23476233176934e-05\\
  0.07	5.11039557908322e-05\\
  0.074	3.9315598377887e-05\\
  0.079	2.82274580931663e-05\\
  0.084	2.03299519538347e-05\\
  0.09	1.37138378661361e-05\\
  0.098	8.15798142150825e-06\\
  0.109	4.05547082077407e-06\\
  0.125	1.57953195717431e-06\\
  0.156	4.92069050672672e-07\\
  0.299	6.53719997578683e-07\\
};
\addlegendentry{$\varepsilon_3$}

\end{axis}
\end{tikzpicture}%

%% file: TikZ/AsyCoSPM_1_CHI_reel_inertiel_perm.tex
%
%
\definecolor{mycolor1}{rgb}{0.00000,0.44700,0.74100}%
\definecolor{mycolor2}{rgb}{0.85000,0.32500,0.09800}%
\definecolor{mycolor3}{rgb}{0.92900,0.69400,0.12500}%
\begin{tikzpicture}

\begin{axis}[%
width=4.521in,
height=2.5in,
at={(0.758in,0.481in)},
scale only axis,
xmin=0.25,
xmax=30,
xlabel style={font=\color{white!15!black}},
xlabel={$t$ $\mathrm{[s]}$},
ymin=-6e-06,
ymax=6e-06,
ylabel style={font=\color{white!15!black}},
ylabel={$\bm{\varepsilon}$ $\mathrm{[rad]}$},
axis background/.style={fill=white},
xmajorgrids,
ymajorgrids,
legend style={legend cell align=left, align=left, draw=white!15!black}
]
\addplot [color=mycolor1]
  table[row sep=crcr]{%
  0.199000000000002	5.60706052965543e-06\\
  0.266999999999999	5.57821306301776e-06\\
  0.332999999999998	5.5409629808878e-06\\
  0.399000000000001	5.49461607590729e-06\\
  0.466000000000001	5.43839251321288e-06\\
  0.533000000000001	5.37308804382519e-06\\
  0.600999999999999	5.2977212909866e-06\\
  0.670000000000002	5.21212434989593e-06\\
  0.739999999999998	5.11617525589259e-06\\
  0.812000000000001	5.00824026872237e-06\\
  0.885000000000002	4.88961617506334e-06\\
  0.960000000000001	4.75854767145734e-06\\
  1.037	4.61480137303738e-06\\
  1.117	4.4562027987638e-06\\
  1.2	4.28239907535044e-06\\
  1.287	4.09094174713687e-06\\
  1.379	3.87917156174922e-06\\
  1.477	3.64431875610194e-06\\
  1.584	3.37858961074744e-06\\
  1.704	3.07122423137685e-06\\
  1.848	2.69282249121261e-06\\
  2.057	2.13338692134357e-06\\
  2.364	1.31246095591564e-06\\
  2.522	8.99637665696673e-07\\
  2.659	5.50681900080008e-07\\
  2.786	2.36293690392131e-07\\
  2.907	-5.40827578277003e-08\\
  3.024	-3.25715372895274e-07\\
  3.139	-5.83554601973901e-07\\
  3.253	-8.30007991226012e-07\\
  3.367	-1.06733033078399e-06\\
  3.482	-1.2975994572173e-06\\
  3.599	-1.52270904152374e-06\\
  3.719	-1.74437451505582e-06\\
  3.842	-1.96241355610027e-06\\
  3.97	-2.18012228359044e-06\\
  4.104	-2.39880738917009e-06\\
  4.245	-2.619708709517e-06\\
  4.395	-2.8455108633807e-06\\
  4.557	-3.08008928584513e-06\\
  4.731	-3.32279820725034e-06\\
  4.918	-3.57443158804926e-06\\
  5.114	-3.8289993078422e-06\\
  5.308	-4.07191126683415e-06\\
  5.49	-4.29079546648836e-06\\
  5.655	-4.48023211419013e-06\\
  5.803	-4.64118704712746e-06\\
  5.938	-4.77896699990765e-06\\
  6.061	-4.89551477755867e-06\\
  6.176	-4.99541268794701e-06\\
  6.283	-5.07934305105096e-06\\
  6.385	-5.15024219538418e-06\\
  6.482	-5.20852461249888e-06\\
  6.575	-5.25524802341693e-06\\
  6.664	-5.29088257650301e-06\\
  6.75	-5.31627851785288e-06\\
  6.834	-5.33197299645849e-06\\
  6.916	-5.33813162206798e-06\\
  6.996	-5.33499458299502e-06\\
  7.074	-5.32287060295289e-06\\
  7.151	-5.30179711688561e-06\\
  7.227	-5.27183933485276e-06\\
  7.302	-5.23310544053857e-06\\
  7.376	-5.18574407237793e-06\\
  7.449	-5.12994157375601e-06\\
  7.522	-5.06496640539922e-06\\
  7.594	-4.99177306423348e-06\\
  7.666	-4.9094384344528e-06\\
  7.738	-4.81789348683037e-06\\
  7.81	-4.71709608618198e-06\\
  7.882	-4.60703151006214e-06\\
  7.954	-4.48771289995875e-06\\
  8.026	-4.35918162011717e-06\\
  8.098	-4.22150758083717e-06\\
  8.171	-4.07268857216536e-06\\
  8.244	-3.9147057044886e-06\\
  8.318	-3.74537368585948e-06\\
  8.393	-3.56452462568768e-06\\
  8.469	-3.37203679023901e-06\\
  8.546	-3.16783953380195e-06\\
  8.625	-2.94909255771358e-06\\
  8.705	-2.71844583110692e-06\\
  8.788	-2.46993266372897e-06\\
  8.873	-2.20628592373373e-06\\
  8.962	-1.92102577756259e-06\\
  9.055	-1.61375374219119e-06\\
  9.154	-1.27744268141328e-06\\
  9.262	-9.01209151749072e-07\\
  9.383	-4.70260633989028e-07\\
  9.53	6.2923181332053e-08\\
  10.036	1.90669116406639e-06\\
  10.147	2.29681182517538e-06\\
  10.245	2.6321751782632e-06\\
  10.334	2.92770145193799e-06\\
  10.417	3.19425018702191e-06\\
  10.495	3.43572240879553e-06\\
  10.57	3.65876705643586e-06\\
  10.642	3.86368402516268e-06\\
  10.711	4.05093363653464e-06\\
  10.778	4.22362331775616e-06\\
  10.843	4.38204979502643e-06\\
  10.906	4.52659840632919e-06\\
  10.968	4.65980792085929e-06\\
  11.029	4.78176437468392e-06\\
  11.089	4.8926047071518e-06\\
  11.148	4.99251352081842e-06\\
  11.206	5.08171958557568e-06\\
  11.264	5.16179431997443e-06\\
  11.321	5.23142622554928e-06\\
  11.378	5.29193794207572e-06\\
  11.435	5.34322083822758e-06\\
  11.491	5.38453857146237e-06\\
  11.547	5.41682135235533e-06\\
  11.603	5.44004720026692e-06\\
  11.659	5.45422173203747e-06\\
  11.715	5.45937831120114e-06\\
  11.772	5.45542921059905e-06\\
  11.829	5.44229409271679e-06\\
  11.886	5.42009550841271e-06\\
  11.944	5.38836003727283e-06\\
  12.003	5.34680379971064e-06\\
  12.062	5.29612182020855e-06\\
  12.122	5.23548725439582e-06\\
  12.183	5.16474909417752e-06\\
  12.246	5.08243117636198e-06\\
  12.31	4.98958888428547e-06\\
  12.376	4.88459840397582e-06\\
  12.444	4.76716420649836e-06\\
  12.514	4.63709701037374e-06\\
  12.587	4.49228183541095e-06\\
  12.664	4.33027106083728e-06\\
  12.745	4.15064508985097e-06\\
  12.832	3.94851776164273e-06\\
  12.928	3.71614126137843e-06\\
  13.037	3.44282906183935e-06\\
  13.169	3.10227841637811e-06\\
  13.393	2.51344273038967e-06\\
  13.592	1.99413281620764e-06\\
  13.723	1.66135245649457e-06\\
  13.836	1.38328771015495e-06\\
  13.939	1.13886346753134e-06\\
  14.036	9.17790227816795e-07\\
  14.128	7.17186726006958e-07\\
  14.217	5.32224969873596e-07\\
  14.303	3.62531451969517e-07\\
  14.388	2.03953408828284e-07\\
  14.471	5.81948178535185e-08\\
  14.553	-7.67539489743285e-08\\
  14.635	-2.0253740018461e-07\\
  14.716	-3.17687018025481e-07\\
  14.797	-4.23749877853652e-07\\
  14.878	-5.20738883125205e-07\\
  14.96	-6.09750333779857e-07\\
  15.042	-6.89651507457256e-07\\
  15.125	-7.614251735788e-07\\
  15.209	-8.24972143931291e-07\\
  15.295	-8.80852628171169e-07\\
  15.383	-9.28784327669518e-07\\
  15.473	-9.68570031290028e-07\\
  15.565	-1.00010544912266e-06\\
  15.661	-1.02378754363031e-06\\
  15.761	-1.03918157989824e-06\\
  15.866	-1.04605204498398e-06\\
  15.977	-1.04404000822456e-06\\
  16.096	-1.03261799466736e-06\\
  16.226	-1.01086219927993e-06\\
  16.373	-9.76865830892848e-07\\
  16.547	-9.27127807415218e-07\\
  16.785	-8.49169090599844e-07\\
  17.33	-6.68030036621303e-07\\
  17.549	-6.06296115535088e-07\\
  17.747	-5.5953253053076e-07\\
  17.938	-5.23560732546002e-07\\
  18.127	-4.97110889341457e-07\\
  18.321	-4.79176751611021e-07\\
  18.526	-4.69479516596039e-07\\
  18.757	-4.67978559015592e-07\\
  19.058	-4.7591396779012e-07\\
  19.63	-4.92438026356012e-07\\
  19.864	-4.88605657267271e-07\\
  20.073	-4.76088786172113e-07\\
  20.274	-4.54901886826065e-07\\
  20.48	-4.2391930321628e-07\\
  20.708	-3.80179368875133e-07\\
  21.035	-3.07092321349955e-07\\
  21.366	-2.35714637142337e-07\\
  21.553	-2.04566642736381e-07\\
  21.709	-1.87516423721945e-07\\
  21.849	-1.81238558383257e-07\\
  21.979	-1.84496748545371e-07\\
  22.102	-1.96659776463548e-07\\
  22.221	-2.17548613790086e-07\\
  22.337	-2.47005885967155e-07\\
  22.452	-2.85300504998531e-07\\
  22.568	-3.33091406901076e-07\\
  22.686	-3.90875147360248e-07\\
  22.809	-4.60335204621742e-07\\
  22.94	-5.43558630994312e-07\\
  23.088	-6.47004863907341e-07\\
  23.278	-7.89647412346994e-07\\
  23.651	-1.07153024231366e-06\\
  23.787	-1.16360475033161e-06\\
  23.902	-1.23252459260925e-06\\
  24.006	-1.28579952018981e-06\\
  24.102	-1.32589964252361e-06\\
  24.192	-1.35447714200154e-06\\
  24.278	-1.37275602440923e-06\\
  24.361	-1.38131578708567e-06\\
  24.441	-1.38054171827662e-06\\
  24.519	-1.37078039585958e-06\\
  24.596	-1.35202448348082e-06\\
  24.672	-1.3242991414586e-06\\
  24.747	-1.28770211560436e-06\\
  24.821	-1.24239724996755e-06\\
  24.894	-1.18860780773389e-06\\
  24.967	-1.12568578458649e-06\\
  25.04	-1.05358040869419e-06\\
  25.113	-9.72286880340789e-07\\
  25.186	-8.81845814149074e-07\\
  25.259	-7.82342539196179e-07\\
  25.333	-6.72359554698687e-07\\
  25.408	-5.51708325247091e-07\\
  25.484	-4.20257467936835e-07\\
  25.561	-2.77938340786932e-07\\
  25.64	-1.22729584717263e-07\\
  25.721	4.56274626969844e-08\\
  25.804	2.27283109666132e-07\\
  25.891	4.2695360491507e-07\\
  25.982	6.451063221391e-07\\
  26.078	8.84490688690676e-07\\
  26.182	1.15311832260545e-06\\
  26.298	1.46210978613226e-06\\
  26.437	1.84192505514602e-06\\
  26.671	2.49233408311511e-06\\
  26.86	3.01360415377872e-06\\
  26.984	3.34660054335245e-06\\
  27.089	3.61957038208516e-06\\
  27.183	3.85486280407576e-06\\
  27.269	4.06110535422499e-06\\
  27.35	4.24627627282348e-06\\
  27.427	4.41315011201482e-06\\
  27.5	4.56229791367946e-06\\
  27.571	4.69819325843446e-06\\
  27.639	4.81926624473772e-06\\
  27.705	4.92775688343272e-06\\
  27.77	5.02545124447806e-06\\
  27.833	5.11106517109283e-06\\
  27.895	5.18624062451067e-06\\
  27.956	5.25109838989124e-06\\
  28.016	5.30580742719167e-06\\
  28.075	5.35058109107922e-06\\
  28.134	5.38619961076847e-06\\
  28.192	5.41211588611645e-06\\
  28.25	5.42886447618685e-06\\
  28.307	5.43627594140617e-06\\
  28.364	5.43462816082751e-06\\
  28.421	5.42385746982177e-06\\
  28.478	5.40392580106186e-06\\
  28.535	5.3748208728166e-06\\
  28.592	5.33655623513596e-06\\
  28.649	5.28917114550609e-06\\
  28.706	5.23273030950122e-06\\
  28.764	5.16609659229061e-06\\
  28.822	5.09030012452172e-06\\
  28.881	5.00394655489345e-06\\
  28.94	4.90844259459777e-06\\
  29	4.80214721321204e-06\\
  29.061	4.68484536852998e-06\\
  29.123	4.55636537566306e-06\\
  29.186	4.41658248107046e-06\\
  29.25	4.26542220566262e-06\\
  29.316	4.10029376141097e-06\\
  29.383	3.92353691935909e-06\\
  29.452	3.73243062412598e-06\\
  29.524	3.52382422619257e-06\\
  29.599	3.29727426517934e-06\\
  29.677	3.05249587384537e-06\\
  29.76	2.78278316301339e-06\\
  29.849	2.48424681004167e-06\\
  29.946	2.14950659582769e-06\\
  30	1.95958112314543e-06\\
};
\addlegendentry{$\varepsilon_1$}

\addplot [color=mycolor2]
  table[row sep=crcr]{%
  0.199000000000002	3.77400795059657e-07\\
  0.280000000000001	5.46996563599578e-07\\
  0.419	8.30096880122255e-07\\
  0.536000000000001	1.05948249284893e-06\\
  0.641000000000002	1.2563727018744e-06\\
  0.738	1.42926920432274e-06\\
  0.830000000000002	1.58413759265841e-06\\
  0.917000000000002	1.72151754895822e-06\\
  1.001	1.8450402841097e-06\\
  1.082	1.95503760025417e-06\\
  1.161	2.05316589685367e-06\\
  1.238	2.13966018591805e-06\\
  1.313	2.21484680906769e-06\\
  1.387	2.27995009538517e-06\\
  1.46	2.33507293856405e-06\\
  1.532	2.38037516808731e-06\\
  1.604	2.41650794308157e-06\\
  1.675	2.4430402199016e-06\\
  1.746	2.46046271001887e-06\\
  1.817	2.46874543918807e-06\\
  1.888	2.46790403579666e-06\\
  1.959	2.45800076470459e-06\\
  2.031	2.43881649453215e-06\\
  2.104	2.41013601609552e-06\\
  2.178	2.37180701745388e-06\\
  2.253	2.3237476050042e-06\\
  2.33	2.26513411050178e-06\\
  2.409	2.19570397064217e-06\\
  2.49	2.11531756022509e-06\\
  2.574	2.02282392791631e-06\\
  2.663	1.91558312678808e-06\\
  2.758	1.79179457404643e-06\\
  2.861	1.64828809445794e-06\\
  2.978	1.47587257259829e-06\\
  3.123	1.25250456761705e-06\\
  3.594	5.1952733670646e-07\\
  3.711	3.51130037046232e-07\\
  3.815	2.10476358120104e-07\\
  3.911	8.9644341727535e-08\\
  4.002	-1.58763207025459e-08\\
  4.09	-1.0879631773264e-07\\
  4.175	-1.89465882982631e-07\\
  4.259	-2.60009684893703e-07\\
  4.342	-3.20486108762452e-07\\
  4.424	-3.71091172013394e-07\\
  4.506	-4.12598623711347e-07\\
  4.589	-4.45464490184122e-07\\
  4.673	-4.69609442887986e-07\\
  4.76	-4.8536634977836e-07\\
  4.85	-4.92380209493604e-07\\
  4.944	-4.90497182425997e-07\\
  5.045	-4.79242050488438e-07\\
  5.158	-4.57244684781699e-07\\
  5.293	-4.21370003778065e-07\\
  5.529	-3.47414669477075e-07\\
  5.706	-2.96510886954593e-07\\
  5.827	-2.70725578843667e-07\\
  5.931	-2.5757610089272e-07\\
  6.025	-2.54713551584018e-07\\
  6.113	-2.61103036081067e-07\\
  6.197	-2.76364115592287e-07\\
  6.277	-2.99957733318479e-07\\
  6.355	-3.32040322348348e-07\\
  6.431	-3.72353145650095e-07\\
  6.506	-4.2121521204308e-07\\
  6.58	-4.78487290678231e-07\\
  6.654	-5.44895307541537e-07\\
  6.728	-6.20492205172241e-07\\
  6.802	-7.05250705834715e-07\\
  6.877	-8.00394424516071e-07\\
  6.952	-9.04651930255795e-07\\
  7.029	-1.02089332898458e-06\\
  7.107	-1.14778132598303e-06\\
  7.188	-1.28879602812049e-06\\
  7.271	-1.44245408506549e-06\\
  7.358	-1.61269580800649e-06\\
  7.45	-1.80194663812472e-06\\
  7.549	-2.01487574358339e-06\\
  7.658	-2.25862197567039e-06\\
  7.785	-2.55209873500917e-06\\
  7.955	-2.95489402546423e-06\\
  8.282	-3.73119663521493e-06\\
  8.411	-4.02694706025386e-06\\
  8.521	-4.27010316528253e-06\\
  8.62	-4.47988031382351e-06\\
  8.712	-4.66570736890048e-06\\
  8.798	-4.83038132159663e-06\\
  8.88	-4.9783857214436e-06\\
  8.959	-5.1119215775941e-06\\
  9.035	-5.2313824134842e-06\\
  9.109	-5.33868417207373e-06\\
  9.181	-5.43409097630843e-06\\
  9.252	-5.51908310697513e-06\\
  9.322	-5.59368776009705e-06\\
  9.391	-5.65798568175069e-06\\
  9.459	-5.71210733113503e-06\\
  9.526	-5.75622903298267e-06\\
  9.592	-5.79056914062903e-06\\
  9.658	-5.8156955127231e-06\\
  9.723	-5.83130960762901e-06\\
  9.788	-5.8377655918207e-06\\
  9.853	-5.83499669915e-06\\
  9.918	-5.82296592455123e-06\\
  9.983	-5.80166684116534e-06\\
  10.048	-5.77112433930438e-06\\
  10.113	-5.73139530146705e-06\\
  10.178	-5.68256916011478e-06\\
  10.244	-5.62380980895227e-06\\
  10.31	-5.5559609819511e-06\\
  10.377	-5.47799173133967e-06\\
  10.445	-5.38974335739795e-06\\
  10.514	-5.29111402514104e-06\\
  10.585	-5.18044348396529e-06\\
  10.658	-5.05739123468629e-06\\
  10.733	-4.92172097210641e-06\\
  10.811	-4.77133256993056e-06\\
  10.892	-4.605920114642e-06\\
  10.978	-4.42098061270713e-06\\
  11.07	-4.21378384984905e-06\\
  11.17	-3.97928913642431e-06\\
  11.285	-3.7001822228433e-06\\
  11.43	-3.33853626344194e-06\\
  11.839	-2.31302024644719e-06\\
  11.955	-2.03461255310344e-06\\
  12.057	-1.79876786887689e-06\\
  12.151	-1.59046607706159e-06\\
  12.239	-1.40448965524342e-06\\
  12.323	-1.23599760470938e-06\\
  12.404	-1.08256570641174e-06\\
  12.483	-9.42018267835465e-07\\
  12.56	-8.1409623220452e-07\\
  12.636	-6.96938084132626e-07\\
  12.711	-5.90424889423957e-07\\
  12.786	-4.93116633748514e-07\\
  12.86	-4.06214148540585e-07\\
  12.934	-3.28375072911058e-07\\
  13.009	-2.58678117148747e-07\\
  13.084	-1.98112054761168e-07\\
  13.16	-1.45853203292745e-07\\
  13.237	-1.019852255979e-07\\
  13.316	-6.61042740546236e-08\\
  13.397	-3.84381095841491e-08\\
  13.481	-1.89020674667972e-08\\
  13.569	-7.66726060419387e-09\\
  13.662	-5.07814590378075e-09\\
  13.762	-1.16471383648786e-08\\
  13.871	-2.81054290951488e-08\\
  13.997	-5.65930200480125e-08\\
  14.157	-1.02504749577292e-07\\
  14.613	-2.38727260182259e-07\\
  14.749	-2.66849038865757e-07\\
  14.871	-2.8299636767315e-07\\
  14.985	-2.89010117171529e-07\\
  15.095	-2.85692912882496e-07\\
  15.203	-2.73256844707248e-07\\
  15.31	-2.51778928372914e-07\\
  15.418	-2.20917673487975e-07\\
  15.528	-1.80310667730055e-07\\
  15.642	-1.29046355823448e-07\\
  15.763	-6.53641691883422e-08\\
  15.894	1.28675843313886e-08\\
  16.044	1.11866150120932e-07\\
  16.238	2.49711348487835e-07\\
  16.694	5.77021783243481e-07\\
  16.851	6.78331705472601e-07\\
  16.989	7.58407786349835e-07\\
  17.117	8.23683013351229e-07\\
  17.24	8.772804953594e-07\\
  17.359	9.19991567371881e-07\\
  17.476	9.52831559430933e-07\\
  17.592	9.76233742733257e-07\\
  17.708	9.90475157891524e-07\\
  17.825	9.95649841684099e-07\\
  17.943	9.91728327903729e-07\\
  18.064	9.78492938230602e-07\\
  18.188	9.55673083780084e-07\\
  18.315	9.23108864014921e-07\\
  18.447	8.80026494343156e-07\\
  18.584	8.26081631544184e-07\\
  18.727	7.60581876591004e-07\\
  18.878	6.82200884227768e-07\\
  19.039	5.89364198333442e-07\\
  19.213	4.79729397540041e-07\\
  19.406	3.48771639124834e-07\\
  19.633	1.85255149887098e-07\\
  19.973	-7.00550195631422e-08\\
  20.335	-3.39376342139985e-07\\
  20.538	-4.81149715625406e-07\\
  20.706	-5.89517188842592e-07\\
  20.855	-6.76632179619219e-07\\
  20.992	-7.47672320500214e-07\\
  21.12	-8.04980043511705e-07\\
  21.242	-8.50478894420803e-07\\
  21.359	-8.84975779769093e-07\\
  21.472	-9.09191609110849e-07\\
  21.582	-9.23706764410781e-07\\
  21.691	-9.28927718746309e-07\\
  21.798	-9.24953749148472e-07\\
  21.905	-9.11851170570799e-07\\
  22.012	-8.89624928390731e-07\\
  22.12	-8.58071011577977e-07\\
  22.23	-8.16790230118158e-07\\
  22.343	-7.65224154264388e-07\\
  22.461	-7.0214062830587e-07\\
  22.586	-6.26007164328257e-07\\
  22.721	-5.34480189884334e-07\\
  22.875	-4.20624189700902e-07\\
  23.071	-2.65964576584565e-07\\
  23.532	1.01231819371606e-07\\
  23.683	2.10090338015334e-07\\
  23.814	2.95555391716107e-07\\
  23.934	3.64774638939025e-07\\
  24.046	4.2033303060407e-07\\
  24.153	4.64357491125611e-07\\
  24.257	4.98000865434278e-07\\
  24.358	5.21562313338109e-07\\
  24.458	5.3571844915723e-07\\
  24.557	5.40560488815345e-07\\
  24.656	5.36233372372408e-07\\
  24.756	5.22657330748189e-07\\
  24.857	4.99817865318164e-07\\
  24.961	4.67169282813984e-07\\
  25.07	4.23721125741849e-07\\
  25.186	3.68194982058867e-07\\
  25.314	2.97540591276402e-07\\
  25.465	2.04672179648924e-07\\
  25.726	3.29786864483594e-08\\
  25.917	-8.80965771443698e-08\\
  26.046	-1.60868889764743e-07\\
  26.155	-2.13414011795976e-07\\
  26.253	-2.51654341809626e-07\\
  26.344	-2.7805939240011e-07\\
  26.429	-2.93676322371539e-07\\
  26.51	-2.99528707614627e-07\\
  26.588	-2.96105110209055e-07\\
  26.664	-2.83613715623687e-07\\
  26.738	-2.62269363560108e-07\\
  26.81	-2.32406208056091e-07\\
  26.881	-1.93854777563729e-07\\
  26.951	-1.4675397963515e-07\\
  27.021	-9.04484700470221e-08\\
  27.09	-2.58299586164412e-08\\
  27.159	4.7884494591699e-08\\
  27.228	1.30674401077613e-07\\
  27.298	2.23849756508798e-07\\
  27.368	3.26119391047541e-07\\
  27.439	4.38907502342545e-07\\
  27.512	5.64061476637789e-07\\
  27.587	7.01920040313553e-07\\
  27.664	8.52689247210492e-07\\
  27.744	1.01854915257604e-06\\
  27.828	1.20192495600691e-06\\
  27.918	1.40771456003108e-06\\
  28.016	1.64116131884384e-06\\
  28.127	1.91501202806421e-06\\
  28.264	2.26263565750173e-06\\
  28.714	3.4118538643213e-06\\
  28.825	3.68165461850367e-06\\
  28.924	3.91323859361137e-06\\
  29.015	4.11707777914216e-06\\
  29.101	4.30064696033128e-06\\
  29.183	4.46659735686694e-06\\
  29.262	4.61739065471534e-06\\
  29.339	4.75520485920811e-06\\
  29.414	4.88026063294456e-06\\
  29.487	4.99288070088255e-06\\
  29.559	5.09482463684208e-06\\
  29.63	5.18618365319412e-06\\
  29.7	5.26710169523881e-06\\
  29.769	5.33776904276806e-06\\
  29.838	5.39924100095845e-06\\
  29.906	5.4506926616682e-06\\
  29.974	5.49297719842912e-06\\
  30	5.50670507237783e-06\\
};
\addlegendentry{$\varepsilon_2$}

\addplot [color=mycolor3]
  table[row sep=crcr]{%
  0.199000000000002	4.31914767062835e-07\\
  0.216999999999999	4.66931918907676e-07\\
  0.256	5.54643261096999e-07\\
  0.468	1.03778688398393e-06\\
  0.602	1.33312506633843e-06\\
  0.724	1.5929627537048e-06\\
  0.837	1.824609011436e-06\\
  0.945	2.03689781841376e-06\\
  1.048	2.23030209767217e-06\\
  1.148	2.40901221815193e-06\\
  1.246	2.5750035739236e-06\\
  1.342	2.72844160775776e-06\\
  1.436	2.86958372441859e-06\\
  1.529	3.0001156154924e-06\\
  1.621	3.12013007430778e-06\\
  1.712	3.22976228517291e-06\\
  1.803	3.33023394816223e-06\\
  1.893	3.42048116408478e-06\\
  1.983	3.50156186712525e-06\\
  2.072	3.57263981953793e-06\\
  2.161	3.63458143581852e-06\\
  2.249	3.68675780748617e-06\\
  2.337	3.72982396257271e-06\\
  2.425	3.76367867715999e-06\\
  2.512	3.78798367606237e-06\\
  2.598	3.8029353248703e-06\\
  2.684	3.80873845173824e-06\\
  2.769	3.80535154675954e-06\\
  2.853	3.79295901353771e-06\\
  2.937	3.7714367415731e-06\\
  3.02	3.74106848255451e-06\\
  3.103	3.70152220696696e-06\\
  3.185	3.65331973384286e-06\\
  3.267	3.59593254728452e-06\\
  3.348	3.53013834342164e-06\\
  3.429	3.45522367339868e-06\\
  3.51	3.37114235904323e-06\\
  3.591	3.27787505582933e-06\\
  3.672	3.17543281980193e-06\\
  3.753	3.06386028725569e-06\\
  3.835	2.94169313619363e-06\\
  3.917	2.81037599592082e-06\\
  4	2.66830940276463e-06\\
  4.084	2.51538212836522e-06\\
  4.169	2.35154318772857e-06\\
  4.256	2.17472426911058e-06\\
  4.345	1.98471148138424e-06\\
  4.437	1.77910843035534e-06\\
  4.532	1.55763414255716e-06\\
  4.632	1.31523045965309e-06\\
  4.737	1.05145449680322e-06\\
  4.85	7.58268761558156e-07\\
  4.974	4.27167929473171e-07\\
  5.116	3.8554713199801e-08\\
  5.299	-4.72059529244007e-07\\
  5.727	-1.66965727288471e-06\\
  5.865	-2.04440016204899e-06\\
  5.983	-2.35584069940842e-06\\
  6.089	-2.62656762473057e-06\\
  6.187	-2.86775149760388e-06\\
  6.278	-3.08268368343079e-06\\
  6.365	-3.27904679764401e-06\\
  6.448	-3.45722228800582e-06\\
  6.527	-3.61778356250397e-06\\
  6.604	-3.76517693823075e-06\\
  6.679	-3.89954621127231e-06\\
  6.752	-4.02112764774643e-06\\
  6.823	-4.13024212164714e-06\\
  6.893	-4.22862607152297e-06\\
  6.961	-4.31513850784881e-06\\
  7.028	-4.39137173557924e-06\\
  7.095	-4.45842657725848e-06\\
  7.161	-4.51530624800967e-06\\
  7.226	-4.56226220890699e-06\\
  7.291	-4.60009635006031e-06\\
  7.356	-4.62871114592645e-06\\
  7.42	-4.64781352960131e-06\\
  7.484	-4.65788039250015e-06\\
  7.548	-4.65890791190304e-06\\
  7.613	-4.65072539412859e-06\\
  7.678	-4.6333018346445e-06\\
  7.743	-4.60672348268076e-06\\
  7.809	-4.57048705726493e-06\\
  7.875	-4.52507860160267e-06\\
  7.942	-4.46978099688522e-06\\
  8.01	-4.40439401216963e-06\\
  8.079	-4.32875684452938e-06\\
  8.149	-4.24275010146857e-06\\
  8.22	-4.14629743872297e-06\\
  8.292	-4.03936672555005e-06\\
  8.366	-3.92030181473046e-06\\
  8.442	-3.78880628559841e-06\\
  8.52	-3.64465426727634e-06\\
  8.601	-3.48567747820994e-06\\
  8.684	-3.31360864436192e-06\\
  8.771	-3.12401871127577e-06\\
  8.862	-2.91644128580515e-06\\
  8.958	-2.68815534099076e-06\\
  9.06	-2.43627937379642e-06\\
  9.169	-2.15786792878703e-06\\
  9.289	-1.84203698694319e-06\\
  9.424	-1.47737933531289e-06\\
  9.587	-1.02758408715431e-06\\
  9.844	-3.0784562099484e-07\\
  10.111	4.37047567913851e-07\\
  10.27	8.71436313332197e-07\\
  10.404	1.22851996664508e-06\\
  10.524	1.53922632506465e-06\\
  10.634	1.81501050988686e-06\\
  10.737	2.06424319415532e-06\\
  10.835	2.29233312865063e-06\\
  10.929	2.50201103924041e-06\\
  11.019	2.69371913930172e-06\\
  11.106	2.87002171361905e-06\\
  11.191	3.03317830940841e-06\\
  11.274	3.18334613425009e-06\\
  11.355	3.32075251918695e-06\\
  11.434	3.44568750421104e-06\\
  11.512	3.55990211531321e-06\\
  11.589	3.66343265056912e-06\\
  11.664	3.75518903439342e-06\\
  11.738	3.83668515979707e-06\\
  11.811	3.90806067684935e-06\\
  11.884	3.97027240239822e-06\\
  11.956	4.02247082220697e-06\\
  12.027	4.0648722112735e-06\\
  12.098	4.09812055934822e-06\\
  12.168	4.12181605113915e-06\\
  12.238	4.13638561980179e-06\\
  12.308	4.14174165541681e-06\\
  12.377	4.13793691933506e-06\\
  12.446	4.12506571478843e-06\\
  12.515	4.10310039811179e-06\\
  12.584	4.07203388164135e-06\\
  12.653	4.0318805041295e-06\\
  12.722	3.98267682300002e-06\\
  12.792	3.9235731783549e-06\\
  12.862	3.85530297464243e-06\\
  12.933	3.77680860452756e-06\\
  13.004	3.68914288984001e-06\\
  13.076	3.59105179725816e-06\\
  13.149	3.48236079972253e-06\\
  13.223	3.36293790681452e-06\\
  13.298	3.2326979173547e-06\\
  13.374	3.09160654055063e-06\\
  13.452	2.93765378245325e-06\\
  13.532	2.77058397912811e-06\\
  13.615	2.58797675556366e-06\\
  13.7	2.39181116867826e-06\\
  13.789	2.17723227891042e-06\\
  13.883	1.94131445141466e-06\\
  13.983	1.68101044906166e-06\\
  14.091	1.39055246961561e-06\\
  14.211	1.05847351861144e-06\\
  14.353	6.55981349240164e-07\\
  14.554	7.61062679544011e-08\\
  14.853	-7.85920036605603e-07\\
  14.998	-1.19422051270135e-06\\
  15.121	-1.53155654203374e-06\\
  15.232	-1.82692106420745e-06\\
  15.335	-2.09194297795534e-06\\
  15.433	-2.33497040724728e-06\\
  15.526	-2.55656421188633e-06\\
  15.616	-2.76199315152326e-06\\
  15.704	-2.95375859238334e-06\\
  15.79	-3.13204563084923e-06\\
  15.874	-3.29713674673826e-06\\
  15.957	-3.45119699218799e-06\\
  16.039	-3.59432607766053e-06\\
  16.12	-3.72667620140987e-06\\
  16.201	-3.84990975987876e-06\\
  16.282	-3.96392245249899e-06\\
  16.362	-4.06740447544962e-06\\
  16.442	-4.16177457296385e-06\\
  16.522	-4.24700999701599e-06\\
  16.602	-4.32310706344197e-06\\
  16.682	-4.39007777330858e-06\\
  16.762	-4.44794653731151e-06\\
  16.842	-4.49674706004544e-06\\
  16.923	-4.53695971458501e-06\\
  17.004	-4.5679629998574e-06\\
  17.085	-4.58980462170189e-06\\
  17.166	-4.6025324031973e-06\\
  17.247	-4.60619301279053e-06\\
  17.329	-4.60070881302954e-06\\
  17.411	-4.58602223929461e-06\\
  17.493	-4.56217552979865e-06\\
  17.575	-4.5292097787808e-06\\
  17.657	-4.48716620127243e-06\\
  17.74	-4.43540933758868e-06\\
  17.823	-4.37444456125036e-06\\
  17.906	-4.30432686826521e-06\\
  17.989	-4.22512025011201e-06\\
  18.073	-4.13578289126804e-06\\
  18.157	-4.03730729559015e-06\\
  18.242	-3.92847120878059e-06\\
  18.327	-3.81052900877421e-06\\
  18.413	-3.68209643752948e-06\\
  18.5	-3.54305207039829e-06\\
  18.588	-3.39331618448568e-06\\
  18.678	-3.23100336885318e-06\\
  18.77	-3.05583950677146e-06\\
  18.864	-2.86762746171121e-06\\
  18.96	-2.6662597818472e-06\\
  19.06	-2.44726365750125e-06\\
  19.164	-2.2102366834531e-06\\
  19.273	-1.95257117852066e-06\\
  19.389	-1.66911505061762e-06\\
  19.515	-1.35195119099762e-06\\
  19.657	-9.85148268739522e-07\\
  19.83	-5.2866946020913e-07\\
  20.175	3.93697796141623e-07\\
  20.379	9.33232726652022e-07\\
  20.531	1.32629336135892e-06\\
  20.663	1.65867120927032e-06\\
  20.784	1.95428556182264e-06\\
  20.897	2.22126530502464e-06\\
  21.004	2.46501736356208e-06\\
  21.107	2.69057801816075e-06\\
  21.207	2.90042572359539e-06\\
  21.304	3.09485965743761e-06\\
  21.399	3.27615370920853e-06\\
  21.492	3.44451979117366e-06\\
  21.584	3.60189779868847e-06\\
  21.674	3.74675906655852e-06\\
  21.763	3.88093527803335e-06\\
  21.851	4.00452695714648e-06\\
  21.938	4.11765711305634e-06\\
  22.025	4.22160745827682e-06\\
  22.111	4.31517426946471e-06\\
  22.196	4.39851480393827e-06\\
  22.28	4.47179105833584e-06\\
  22.364	4.53587575677261e-06\\
  22.447	4.59000566976897e-06\\
  22.529	4.6343454052078e-06\\
  22.61	4.66906078244733e-06\\
  22.691	4.69457810226004e-06\\
  22.771	4.7105817948534e-06\\
  22.85	4.71724758099867e-06\\
  22.928	4.7147594059993e-06\\
  23.006	4.70310365585647e-06\\
  23.083	4.68245958273883e-06\\
  23.16	4.65260154669522e-06\\
  23.236	4.61397677398168e-06\\
  23.312	4.56615484267786e-06\\
  23.387	4.509864492519e-06\\
  23.462	4.4444777635988e-06\\
  23.537	4.36996028341241e-06\\
  23.612	4.28630488258364e-06\\
  23.687	4.19353420255675e-06\\
  23.762	4.09170299775496e-06\\
  23.838	3.97936257812148e-06\\
  23.914	3.85793912727195e-06\\
  23.991	3.72582702823365e-06\\
  24.069	3.58288962232223e-06\\
  24.148	3.42904643702013e-06\\
  24.229	3.26216352064534e-06\\
  24.312	3.08196437259767e-06\\
  24.397	2.88826803540587e-06\\
  24.485	2.6785691424891e-06\\
  24.577	2.45009561439247e-06\\
  24.673	2.20250366567143e-06\\
  24.776	1.92756006001105e-06\\
  24.887	1.62196139186221e-06\\
  25.011	1.27122518378542e-06\\
  25.158	8.45955135275744e-07\\
  25.377	2.02056039455556e-07\\
  25.646	-5.87053623490874e-07\\
  25.792	-1.00596691865462e-06\\
  25.915	-1.34990258260359e-06\\
  26.025	-1.64848637140835e-06\\
  26.127	-1.91628515722186e-06\\
  26.223	-2.15920481139165e-06\\
  26.314	-2.38036709987455e-06\\
  26.401	-2.58274607745079e-06\\
  26.485	-2.76908189178471e-06\\
  26.567	-2.94181563731399e-06\\
  26.646	-3.09917139418303e-06\\
  26.724	-3.24537571572137e-06\\
  26.8	-3.37869458277851e-06\\
  26.875	-3.50108035007679e-06\\
  26.949	-3.61259985837137e-06\\
  27.022	-3.71336444615622e-06\\
  27.094	-3.80352690498853e-06\\
  27.165	-3.88327829980994e-06\\
  27.235	-3.9528447040027e-06\\
  27.305	-4.01328241039778e-06\\
  27.375	-4.06448536693915e-06\\
  27.444	-4.10583977483725e-06\\
  27.513	-4.13808808730209e-06\\
  27.582	-4.16119940638282e-06\\
  27.651	-4.1751659338729e-06\\
  27.72	-4.18000254498452e-06\\
  27.789	-4.17574620570349e-06\\
  27.859	-4.16219658205819e-06\\
  27.929	-4.13943112675952e-06\\
  27.999	-4.10755304969257e-06\\
  28.07	-4.06603690450424e-06\\
  28.141	-4.01541990768806e-06\\
  28.213	-3.95496616789615e-06\\
  28.286	-3.8844877181532e-06\\
  28.36	-3.80382525477785e-06\\
  28.435	-3.7128496153116e-06\\
  28.511	-3.61146322092054e-06\\
  28.588	-3.49960142287387e-06\\
  28.667	-3.37560735630404e-06\\
  28.747	-3.24088845715664e-06\\
  28.829	-3.09364778772192e-06\\
  28.913	-2.93365873105245e-06\\
  29	-2.75869646060301e-06\\
  29.089	-2.57053060437329e-06\\
  29.181	-2.36689914245858e-06\\
  29.277	-2.14525703867707e-06\\
  29.378	-1.9028339011129e-06\\
  29.485	-1.63670642550073e-06\\
  29.599	-1.34389474837349e-06\\
  29.723	-1.01614022440799e-06\\
  29.863	-6.36743905602088e-07\\
  30	-2.58364313054926e-07\\
};
\addlegendentry{$\varepsilon_3$}

\end{axis}
\end{tikzpicture}%

%% file: TikZ/AsyCoSPM_1_dot_THETA_trans.tex
%
%
\definecolor{mycolor1}{rgb}{0.00000,0.44700,0.74100}%
\definecolor{mycolor2}{rgb}{0.85000,0.32500,0.09800}%
\definecolor{mycolor3}{rgb}{0.92900,0.69400,0.12500}%
\begin{tikzpicture}

\begin{axis}[%
width=4.521in,
height=2.5in,
at={(0.758in,0.481in)},
scale only axis,
xmin=0,
xmax=0.25,
xlabel style={font=\color{white!15!black}},
xlabel={$t$ $\mathrm{[s]}$},
xtick={0,0.05,0.1,0.15,0.2,0.25},
ymin=-1.5,
ymax=0,
ylabel style={font=\color{white!15!black}},
ylabel={$\dot{\bm{\theta}}$ $\mathrm{[rad/s]}$},
axis background/.style={fill=white},
xmajorgrids,
ymajorgrids,
legend style={legend cell align=left, align=left, draw=white!15!black}
]
\addplot [color=mycolor1]
  table[row sep=crcr]{%
  0	0\\
  0.002	0\\
  0.00299999999999989	-0.239027054395866\\
  0.00499999999999989	-0.920095602834158\\
  0.00600000000000001	-1.15873948305654\\
  0.0069999999999999	-1.26020894083045\\
  0.00800000000000001	-1.23310219264423\\
  0.0089999999999999	-1.12074936547396\\
  0.01	-0.97930352898702\\
  0.0109999999999999	-0.858413180019744\\
  0.012	-0.788369940395343\\
  0.0129999999999999	-0.776323023418267\\
  0.014	-0.810205657164469\\
  0.016	-0.923778037362076\\
  0.0169999999999999	-0.961984732341582\\
  0.018	-0.974178563731325\\
  0.0189999999999999	-0.962264972320367\\
  0.02	-0.93478892005169\\
  0.0209999999999999	-0.902726121195447\\
  0.022	-0.875631300615522\\
  0.0229999999999999	-0.85918926591696\\
  0.024	-0.854532585759938\\
  0.0249999999999999	-0.859075275879995\\
  0.0269999999999999	-0.877268182361012\\
  0.028	-0.882760358504562\\
  0.0289999999999999	-0.883244257158307\\
  0.03	-0.879155126577404\\
  0.0309999999999999	-0.872208377542858\\
  0.032	-0.864563533419718\\
  0.0329999999999999	-0.858071295791928\\
  0.034	-0.853804602879838\\
  0.0349999999999999	-0.851939304983088\\
  0.036	-0.851930122202914\\
  0.038	-0.853778940545495\\
  0.0389999999999999	-0.854036703281305\\
  0.04	-0.853353379892201\\
  0.0409999999999999	-0.851826343447286\\
  0.044	-0.845873891616899\\
  0.0449999999999999	-0.844534297990605\\
  0.046	-0.843701602166303\\
  0.0469999999999999	-0.843260528284473\\
  0.05	-0.842484104324383\\
  0.0509999999999999	-0.841996411963842\\
  0.0529999999999999	-0.840663976825246\\
  0.0549999999999999	-0.839352180485115\\
  0.056	-0.838848611019392\\
  0.0580000000000001	-0.838164779761737\\
  0.0680000000000001	-0.835750696233923\\
  0.0720000000000001	-0.835207751064804\\
  0.0780000000000001	-0.834553013272934\\
  0.083	-0.834225211355242\\
  0.0900000000000001	-0.833933084623146\\
  0.099	-0.833773163448879\\
  0.111	-0.833760251398452\\
  0.129	-0.833939469260418\\
  0.161	-0.834460504522603\\
  0.212	-0.835469825711839\\
  0.269	-0.83676879369452\\
  0.299	-0.837521058598413\\
};
\addlegendentry{$\dot{\theta}_1$}

\addplot [color=mycolor2]
  table[row sep=crcr]{%
  0	0\\
  0.002	0\\
  0.00299999999999989	-0.332111350921252\\
  0.00499999999999989	-1.11242276541038\\
  0.00600000000000001	-1.36128224446612\\
  0.0069999999999999	-1.44967437870524\\
  0.00800000000000001	-1.39488139859764\\
  0.0089999999999999	-1.2527128454133\\
  0.01	-1.0884523030856\\
  0.0109999999999999	-0.956689979428593\\
  0.012	-0.887726953863238\\
  0.0129999999999999	-0.885206526180048\\
  0.014	-0.931931403064476\\
  0.016	-1.06360280571047\\
  0.0169999999999999	-1.10295697535678\\
  0.018	-1.11163533244264\\
  0.0189999999999999	-1.09361383237469\\
  0.02	-1.05974769614659\\
  0.0209999999999999	-1.02282068674629\\
  0.022	-0.993289146379382\\
  0.0229999999999999	-0.976799548827684\\
  0.024	-0.973758329611921\\
  0.0249999999999999	-0.980555891282572\\
  0.026	-0.991660533463232\\
  0.0269999999999999	-1.00174215567915\\
  0.028	-1.00719184471203\\
  0.0289999999999999	-1.00674396531273\\
  0.03	-1.0012472687831\\
  0.032	-0.984122839093214\\
  0.0329999999999999	-0.97702281789601\\
  0.034	-0.972627361862953\\
  0.0349999999999999	-0.970963421105874\\
  0.036	-0.971281737927564\\
  0.0369999999999999	-0.972474569646363\\
  0.038	-0.973492262285435\\
  0.0389999999999999	-0.973636077729652\\
  0.04	-0.972672584345272\\
  0.0409999999999999	-0.970781570634008\\
  0.0429999999999999	-0.966005591957849\\
  0.044	-0.964000447626368\\
  0.0449999999999999	-0.96257623204\\
  0.046	-0.961726821210175\\
  0.0469999999999999	-0.96129619142165\\
  0.0489999999999999	-0.960815727808419\\
  0.05	-0.960422115770944\\
  0.0509999999999999	-0.959837037365475\\
  0.0529999999999999	-0.958288479487883\\
  0.0549999999999999	-0.95682241930558\\
  0.056	-0.956274679343833\\
  0.0569999999999999	-0.955858133402985\\
  0.0589999999999999	-0.955274154531844\\
  0.0620000000000001	-0.954454511665484\\
  0.0660000000000001	-0.953264676167445\\
  0.0680000000000001	-0.952838885582901\\
  0.0720000000000001	-0.952241093591355\\
  0.077	-0.951615103131719\\
  0.081	-0.951287628938503\\
  0.0880000000000001	-0.950920255132584\\
  0.095	-0.950742382335622\\
  0.104	-0.95067662496243\\
  0.117	-0.950763017307807\\
  0.138	-0.951093399074089\\
  0.182	-0.951992682861733\\
  0.258	-0.953721242789513\\
  0.299	-0.954717850320249\\
};
\addlegendentry{$\dot{\theta}_2$}

\addplot [color=mycolor3]
  table[row sep=crcr]{%
  0	0\\
  0.002	0\\
  0.00299999999999989	-0.285576084814451\\
  0.00499999999999989	-1.01639012144618\\
  0.00600000000000001	-1.2602827040801\\
  0.0069999999999999	-1.35536543828094\\
  0.00800000000000001	-1.31452423805601\\
  0.0089999999999999	-1.18729163551752\\
  0.01	-1.03438129888871\\
  0.0109999999999999	-0.907936299480675\\
  0.012	-0.838290699607768\\
  0.0129999999999999	-0.830878300476073\\
  0.014	-0.871094433971119\\
  0.016	-0.993693464536556\\
  0.0169999999999999	-1.03251403245248\\
  0.018	-1.04299330483593\\
  0.0189999999999999	-1.0280499522923\\
  0.02	-0.997371646677147\\
  0.0209999999999999	-0.962837622414189\\
  0.022	-0.934462787215265\\
  0.0229999999999999	-0.917927280321029\\
  0.024	-0.914014878667207\\
  0.0249999999999999	-0.919637641589192\\
  0.0269999999999999	-0.939287938826755\\
  0.028	-0.944756971829556\\
  0.0289999999999999	-0.944774224300351\\
  0.03	-0.939974873545633\\
  0.0329999999999999	-0.917249167191841\\
  0.034	-0.912885184153511\\
  0.0349999999999999	-0.911089901053027\\
  0.036	-0.911218756605322\\
  0.038	-0.913213252429718\\
  0.0389999999999999	-0.913401546720942\\
  0.04	-0.912566036517859\\
  0.0409999999999999	-0.910843475876541\\
  0.0429999999999999	-0.906356260300136\\
  0.044	-0.904423096724054\\
  0.0449999999999999	-0.903021313609939\\
  0.046	-0.902161138275761\\
  0.0469999999999999	-0.901707587215121\\
  0.0489999999999999	-0.901210923046776\\
  0.05	-0.900837357538742\\
  0.0509999999999999	-0.900287083593188\\
  0.0529999999999999	-0.898817725217994\\
  0.0549999999999999	-0.897398204148915\\
  0.056	-0.896857128567325\\
  0.0569999999999999	-0.896439935879423\\
  0.0589999999999999	-0.895849150233508\\
  0.0620000000000001	-0.895036851115785\\
  0.0660000000000001	-0.893857340142904\\
  0.0680000000000001	-0.893422358473627\\
  0.0720000000000001	-0.892798522942682\\
  0.077	-0.892136879383679\\
  0.081	-0.89176982737234\\
  0.0880000000000001	-0.891325292189186\\
  0.095	-0.891060620709266\\
  0.105	-0.890863490546665\\
  0.119	-0.89078108132632\\
  0.142	-0.890843426494216\\
  0.185	-0.891155184827574\\
  0.245	-0.891761712775974\\
  0.299	-0.892450513438322\\
};
\addlegendentry{$\dot{\theta}_3$}

\end{axis}
\end{tikzpicture}%

%% file: TikZ/AsyCoSPM_1_dot_THETA_perm.tex
%
%
\definecolor{mycolor1}{rgb}{0.00000,0.44700,0.74100}%
\definecolor{mycolor2}{rgb}{0.85000,0.32500,0.09800}%
\definecolor{mycolor3}{rgb}{0.92900,0.69400,0.12500}%
\begin{tikzpicture}

\begin{axis}[%
width=4.521in,
height=2.5in,
at={(0.758in,0.481in)},
scale only axis,
xmin=0.25,
xmax=30,
xlabel style={font=\color{white!15!black}},
xlabel={$t$ $\mathrm{[s]}$},
ymin=-1.2,
ymax=-0.8,
ylabel style={font=\color{white!15!black}},
ylabel={$\dot{\bm{\theta}}$ $\mathrm{[rad/s]}$},
axis background/.style={fill=white},
xmajorgrids,
ymajorgrids,
legend style={legend cell align=left, align=left, draw=white!15!black}
]
\addplot [color=mycolor1]
  table[row sep=crcr]{%
  0.199000000000002	-0.835198023697917\\
  0.274000000000001	-0.836890919579464\\
  0.350999999999999	-0.838934593714047\\
  0.428999999999998	-0.84130893831556\\
  0.509	-0.844049550263978\\
  0.591000000000001	-0.847165423567244\\
  0.675000000000001	-0.85066334108031\\
  0.762	-0.854594611187153\\
  0.852	-0.858971181994782\\
  0.945	-0.86380147678916\\
  1.042	-0.869146751594688\\
  1.144	-0.875075940883853\\
  1.252	-0.881663684554436\\
  1.367	-0.888988044642534\\
  1.491	-0.89719521402521\\
  1.627	-0.906507297675166\\
  1.78	-0.917297236476177\\
  1.957	-0.930094728167525\\
  2.177	-0.946321887730591\\
  2.51	-0.97122701812129\\
  3.033	-1.01032207541845\\
  3.286	-1.02891366490574\\
  3.491	-1.04367984705397\\
  3.666	-1.05598607605558\\
  3.82	-1.06651688506991\\
  3.959	-1.07572218406164\\
  4.086	-1.08383304465639\\
  4.204	-1.09106812082345\\
  4.315	-1.09756997430357\\
  4.419	-1.10335991097924\\
  4.518	-1.10856961774631\\
  4.613	-1.11326462058783\\
  4.704	-1.11745808434596\\
  4.792	-1.12120850217087\\
  4.877	-1.12452681275269\\
  4.96	-1.12746068213233\\
  5.04	-1.12998622445067\\
  5.119	-1.13217468752554\\
  5.196	-1.13400280981926\\
  5.272	-1.13550049680715\\
  5.346	-1.13665595622705\\
  5.419	-1.13749424601948\\
  5.491	-1.13802002471103\\
  5.563	-1.1382401323688\\
  5.634	-1.13815216475795\\
  5.705	-1.13775656077984\\
  5.775	-1.13706163277278\\
  5.845	-1.13606118585947\\
  5.915	-1.13475336964397\\
  5.985	-1.13313727644863\\
  6.055	-1.13121295977695\\
  6.125	-1.12898144711659\\
  6.195	-1.12644474716275\\
  6.265	-1.12360585159114\\
  6.336	-1.12042177783924\\
  6.408	-1.11688484427587\\
  6.48	-1.11304416620067\\
  6.553	-1.10884773970461\\
  6.628	-1.10422853872465\\
  6.704	-1.09924049555166\\
  6.782	-1.09381256421096\\
  6.862	-1.0879359501793\\
  6.944	-1.08160494705683\\
  7.029	-1.07473478276813\\
  7.118	-1.06723114757061\\
  7.211	-1.05908185429088\\
  7.31	-1.05009851645267\\
  7.418	-1.03998597184149\\
  7.538	-1.02843680685427\\
  7.681	-1.0143559559282\\
  7.896	-0.992838519112269\\
  8.148	-0.967688723657641\\
  8.289	-0.953929582787772\\
  8.408	-0.942619295512571\\
  8.515	-0.932753822334384\\
  8.614	-0.923931184911304\\
  8.707	-0.915947689362469\\
  8.796	-0.908613709207106\\
  8.881	-0.901913335687432\\
  8.963	-0.895751237792485\\
  9.043	-0.890042338167362\\
  9.121	-0.884778993435727\\
  9.198	-0.879889312386819\\
  9.273	-0.875430128216568\\
  9.347	-0.871333444290379\\
  9.42	-0.867595227673817\\
  9.492	-0.864210080762987\\
  9.564	-0.861130736593811\\
  9.635	-0.858398668121197\\
  9.706	-0.855973190618752\\
  9.776	-0.853885371105822\\
  9.846	-0.852101500109637\\
  9.916	-0.850623480430098\\
  9.986	-0.849452540935964\\
  10.056	-0.84858924972098\\
  10.126	-0.848033527046294\\
  10.196	-0.847784657821197\\
  10.266	-0.847841303428769\\
  10.336	-0.848201512760475\\
  10.407	-0.848874344559693\\
  10.478	-0.849853559266581\\
  10.549	-0.851135250588115\\
  10.621	-0.852739254417699\\
  10.694	-0.854672877200365\\
  10.767	-0.856909756623573\\
  10.841	-0.859479730001429\\
  10.916	-0.862387121249995\\
  10.993	-0.865679764288462\\
  11.071	-0.869322484205377\\
  11.151	-0.873367791489986\\
  11.232	-0.877769075310571\\
  11.315	-0.882582418345887\\
  11.401	-0.88787583060677\\
  11.49	-0.893661098845993\\
  11.583	-0.900015538053299\\
  11.68	-0.906950625889806\\
  11.783	-0.914622385940763\\
  11.894	-0.923199334199285\\
  12.017	-0.933016703018207\\
  12.159	-0.944667905406661\\
  12.343	-0.960092926938788\\
  12.785	-0.997262864919758\\
  12.932	-1.00924026730351\\
  13.059	-1.01928871826522\\
  13.175	-1.02816698409221\\
  13.284	-1.03620722806198\\
  13.388	-1.04357368665946\\
  13.488	-1.05035052152257\\
  13.584	-1.05655343142322\\
  13.678	-1.06232289261961\\
  13.77	-1.06766419401691\\
  13.86	-1.07258548783703\\
  13.949	-1.07714685339414\\
  14.037	-1.08135040822184\\
  14.124	-1.08519979657346\\
  14.21	-1.08869998578265\\
  14.295	-1.09185707740889\\
  14.38	-1.09470991658257\\
  14.465	-1.09725561627272\\
  14.549	-1.09946744525296\\
  14.633	-1.1013755739016\\
  14.717	-1.10297895829684\\
  14.801	-1.10427703387175\\
  14.885	-1.10526967388252\\
  14.969	-1.10595715304087\\
  15.053	-1.10634011672424\\
  15.137	-1.10641955605434\\
  15.221	-1.10619678904738\\
  15.306	-1.10566541477021\\
  15.391	-1.1048282797775\\
  15.476	-1.10368771816141\\
  15.562	-1.10222765696673\\
  15.648	-1.10046282526593\\
  15.735	-1.09837092818562\\
  15.822	-1.09597472693835\\
  15.91	-1.0932459841851\\
  15.999	-1.09017947083959\\
  16.089	-1.08677077765475\\
  16.18	-1.0830164412443\\
  16.272	-1.07891408419303\\
  16.365	-1.0744625689388\\
  16.46	-1.06960950928756\\
  16.557	-1.06434746624733\\
  16.656	-1.0586713698919\\
  16.758	-1.05251713918649\\
  16.864	-1.04581258733906\\
  16.974	-1.03854668256817\\
  17.09	-1.03057505636366\\
  17.214	-1.02174241768606\\
  17.349	-1.01181493422686\\
  17.503	-1.00017612977259\\
  17.7	-0.984961023906063\\
  18.177	-0.947994874777994\\
  18.326	-0.936835083054749\\
  18.454	-0.927546218270535\\
  18.57	-0.919428510370238\\
  18.678	-0.912174256957556\\
  18.779	-0.905692005952972\\
  18.875	-0.899831187111467\\
  18.968	-0.894458550650747\\
  19.057	-0.889619944104329\\
  19.144	-0.885195575749407\\
  19.228	-0.881226882192774\\
  19.31	-0.877654361666067\\
  19.391	-0.874430912412944\\
  19.47	-0.871590859754164\\
  19.548	-0.869091139093253\\
  19.625	-0.866929008107686\\
  19.701	-0.865100464445568\\
  19.776	-0.86360033675285\\
  19.85	-0.862422376050976\\
  19.924	-0.861549606736101\\
  19.997	-0.860991881867292\\
  20.07	-0.860738819967356\\
  20.143	-0.860793177455637\\
  20.215	-0.861149914935805\\
  20.287	-0.861808960736326\\
  20.359	-0.862770881826286\\
  20.431	-0.86403553659952\\
  20.503	-0.865602066468206\\
  20.576	-0.867496921063132\\
  20.649	-0.869698050085514\\
  20.722	-0.872202262129754\\
  20.796	-0.875046044457971\\
  20.871	-0.878236797521382\\
  20.946	-0.881732168828279\\
  21.022	-0.885577829811147\\
  21.099	-0.889777625448694\\
  21.178	-0.894394301161345\\
  21.258	-0.899375858579912\\
  21.34	-0.904788903946585\\
  21.424	-0.910640994177559\\
  21.511	-0.917011728730579\\
  21.601	-0.923912526460871\\
  21.694	-0.931350057757999\\
  21.792	-0.939494625850109\\
  21.896	-0.948445173668361\\
  22.009	-0.958480139638095\\
  22.135	-0.969981925324017\\
  22.285	-0.983993630638366\\
  22.507	-1.00507662709473\\
  22.779	-1.03084640960103\\
  22.927	-1.04455222588723\\
  23.051	-1.05573461610446\\
  23.162	-1.06544388142354\\
  23.265	-1.07415022854909\\
  23.362	-1.08204445237094\\
  23.454	-1.08922767654319\\
  23.542	-1.09579620494625\\
  23.627	-1.10183865044051\\
  23.71	-1.10743378092972\\
  23.791	-1.1125870981055\\
  23.87	-1.11730683823621\\
  23.947	-1.12160363685869\\
  24.023	-1.12553999454053\\
  24.098	-1.12911821856638\\
  24.172	-1.13234197371617\\
  24.245	-1.13521617234169\\
  24.317	-1.13774686687539\\
  24.388	-1.139941145232\\
  24.459	-1.14183152515584\\
  24.53	-1.143413963716\\
  24.6	-1.14466936876245\\
  24.67	-1.1456196068317\\
  24.74	-1.14626276841641\\
  24.81	-1.14659761016864\\
  24.88	-1.14662356203506\\
  24.95	-1.14634073389315\\
  25.02	-1.14574992160438\\
  25.09	-1.14485261239511\\
  25.16	-1.14365098946608\\
  25.231	-1.14212429627245\\
  25.302	-1.14029124655067\\
  25.374	-1.13812413099879\\
  25.446	-1.13565205567546\\
  25.519	-1.13284072009706\\
  25.593	-1.1296850037656\\
  25.668	-1.12618110935558\\
  25.744	-1.12232669403314\\
  25.822	-1.1180644477099\\
  25.902	-1.11338401749778\\
  25.984	-1.10827765744505\\
  26.068	-1.10274061249553\\
  26.155	-1.09670041052564\\
  26.246	-1.09007537507002\\
  26.342	-1.08277706987854\\
  26.444	-1.07471358654825\\
  26.555	-1.06562786703556\\
  26.68	-1.05508118370014\\
  26.83	-1.04210536406494\\
  27.081	-1.02002740136009\\
  27.296	-1.00124513581774\\
  27.436	-0.989317618266607\\
  27.555	-0.97947915229172\\
  27.662	-0.970932910542761\\
  27.762	-0.963249452095042\\
  27.856	-0.956330276901994\\
  27.946	-0.950009439725111\\
  28.033	-0.944205201286991\\
  28.117	-0.93890553597511\\
  28.199	-0.93403631741548\\
  28.279	-0.929588496640189\\
  28.358	-0.925500290791017\\
  28.436	-0.921768944530616\\
  28.513	-0.918389704353945\\
  28.589	-0.915356019933697\\
  28.665	-0.912625806429421\\
  28.741	-0.910201476689089\\
  28.817	-0.908084133040933\\
  28.893	-0.906273607756575\\
  28.969	-0.904768514692165\\
  29.045	-0.90356631141438\\
  29.122	-0.902653465246285\\
  29.2	-0.902037020224331\\
  29.278	-0.901724821858849\\
  29.357	-0.90171164803559\\
  29.437	-0.902000892246587\\
  29.519	-0.902603898838244\\
  29.602	-0.903519674096355\\
  29.687	-0.904764133196124\\
  29.774	-0.906346082886841\\
  29.863	-0.908272718990411\\
  29.954	-0.910549664897385\\
  30	-0.91181377931235\\  
};
\addlegendentry{$\dot{\theta}_1$}

\addplot [color=mycolor2]
  table[row sep=crcr]{%
  0.199000000000002	-0.952364910367255\\
  0.305	-0.954867253731482\\
  0.419	-0.957868095725932\\
  0.542000000000002	-0.961417099459283\\
  0.678000000000001	-0.965651877045516\\
  0.838000000000001	-0.970950574795978\\
  1.055	-0.978469218902287\\
  1.473	-0.992997062950334\\
  1.646	-0.99866249587458\\
  1.797	-1.00330698905402\\
  1.936	-1.00728130930717\\
  2.069	-1.01077985183137\\
  2.198	-1.01386724703433\\
  2.324	-1.01657950719175\\
  2.45	-1.01898673603737\\
  2.576	-1.02109003704101\\
  2.704	-1.02292254321089\\
  2.835	-1.02449341182192\\
  2.971	-1.02581727637214\\
  3.113	-1.02689138431631\\
  3.263	-1.02771667336249\\
  3.422	-1.0282839277682\\
  3.593	-1.02858659303217\\
  3.778	-1.02860501975513\\
  3.975	-1.02831742069553\\
  4.179	-1.02771522136343\\
  4.381	-1.02681651956124\\
  4.572	-1.02566681527412\\
  4.75	-1.02429400970258\\
  4.914	-1.02272814832443\\
  5.066	-1.02097699999906\\
  5.209	-1.01902795264778\\
  5.344	-1.0168861570859\\
  5.473	-1.01453679706171\\
  5.597	-1.01197506567212\\
  5.717	-1.00919231024363\\
  5.834	-1.00617488113887\\
  5.949	-1.0029031956724\\
  6.062	-0.999383568595587\\
  6.174	-0.995590995635382\\
  6.286	-0.991493428084365\\
  6.399	-0.987051914619773\\
  6.513	-0.982264484216536\\
  6.629	-0.977088402665011\\
  6.749	-0.971428518509981\\
  6.875	-0.965178627521333\\
  7.01	-0.95817411066497\\
  7.161	-0.950027133259145\\
  7.346	-0.939723520590039\\
  7.893	-0.90904862699162\\
  8.03	-0.901800674254613\\
  8.15	-0.895751399910132\\
  8.259	-0.890557059058274\\
  8.361	-0.886000065212734\\
  8.457	-0.882015245432047\\
  8.548	-0.878539543575812\\
  8.636	-0.875482477469397\\
  8.721	-0.872834841199918\\
  8.803	-0.870583163624744\\
  8.883	-0.868688529786372\\
  8.961	-0.86714225538374\\
  9.038	-0.8659193453245\\
  9.114	-0.865018630922872\\
  9.189	-0.864437298564035\\
  9.263	-0.86417098062277\\
  9.336	-0.864213851877189\\
  9.408	-0.864558730697372\\
  9.48	-0.865208308187459\\
  9.552	-0.866165705015145\\
  9.623	-0.867413224208367\\
  9.694	-0.868962922143528\\
  9.765	-0.870814572877713\\
  9.837	-0.872999273942128\\
  9.909	-0.875490485398615\\
  9.981	-0.878284363333389\\
  10.054	-0.881420830742346\\
  10.128	-0.884905719954304\\
  10.203	-0.888743095196141\\
  10.279	-0.892935069416129\\
  10.357	-0.897542597515748\\
  10.437	-0.902574563355678\\
  10.52	-0.908104917034528\\
  10.606	-0.914145866072118\\
  10.695	-0.920704140043977\\
  10.789	-0.927937032534487\\
  10.89	-0.936017582437572\\
  11.001	-0.945211406529104\\
  11.127	-0.955963606342987\\
  11.283	-0.969598110942059\\
  11.799	-1.01494699434089\\
  11.926	-1.02565181676295\\
  12.039	-1.03487667099233\\
  12.144	-1.04314489069859\\
  12.243	-1.05063521744838\\
  12.337	-1.05744382308016\\
  12.428	-1.06373019815135\\
  12.516	-1.06950503446186\\
  12.602	-1.07484389136778\\
  12.686	-1.07975490438877\\
  12.769	-1.08430196868465\\
  12.851	-1.08848722373734\\
  12.932	-1.09231468115248\\
  13.012	-1.09579002081778\\
  13.092	-1.0989580858357\\
  13.171	-1.10178198307063\\
  13.25	-1.10430134019014\\
  13.329	-1.10651534526049\\
  13.408	-1.10842421610375\\
  13.487	-1.11002916153623\\
  13.567	-1.11134691368271\\
  13.647	-1.11235839858542\\
  13.728	-1.11307467777206\\
  13.809	-1.11348630557936\\
  13.891	-1.11359879893601\\
  13.974	-1.11340808453738\\
  14.058	-1.11291126506253\\
  14.144	-1.11209542089619\\
  14.231	-1.11096443702164\\
  14.32	-1.1095012949948\\
  14.411	-1.10769850240094\\
  14.504	-1.10555049890646\\
  14.6	-1.10302591471581\\
  14.699	-1.10011424404279\\
  14.801	-1.09680792651009\\
  14.907	-1.09306585015804\\
  15.018	-1.08883945251676\\
  15.134	-1.08411628440977\\
  15.257	-1.07880074164889\\
  15.388	-1.07283192259821\\
  15.53	-1.06605183367926\\
  15.685	-1.05834124676858\\
  15.859	-1.04937414219487\\
  16.061	-1.03865013455486\\
  16.312	-1.02500525958661\\
  16.693	-1.00395094518345\\
  17.33	-0.96876018599124\\
  17.638	-0.952067884107542\\
  17.887	-0.938870174722439\\
  18.098	-0.927983934269857\\
  18.283	-0.918737457468126\\
  18.449	-0.91074039804278\\
  18.6	-0.903765346849092\\
  18.74	-0.897598999742378\\
  18.871	-0.892130763612073\\
  18.995	-0.887258410546011\\
  19.112	-0.88296231748339\\
  19.224	-0.879150538223481\\
  19.332	-0.875777270382443\\
  19.436	-0.872830640905701\\
  19.537	-0.870271193330804\\
  19.635	-0.868089163584965\\
  19.731	-0.86625455876155\\
  19.825	-0.864762155320669\\
  19.917	-0.863604786650633\\
  20.008	-0.862765956677162\\
  20.097	-0.862249393686948\\
  20.185	-0.86204241726151\\
  20.272	-0.86214204455316\\
  20.358	-0.862544155210674\\
  20.443	-0.863243553687223\\
  20.528	-0.864247635731932\\
  20.612	-0.865542825924326\\
  20.696	-0.867141908958047\\
  20.78	-0.869046827308139\\
  20.864	-0.871258693848318\\
  20.948	-0.87377777268712\\
  21.032	-0.87660346156105\\
  21.116	-0.87973427559772\\
  21.2	-0.883167832289779\\
  21.285	-0.886947065666433\\
  21.371	-0.891078337012317\\
  21.457	-0.895513446264623\\
  21.545	-0.900359329206331\\
  21.634	-0.905568445542027\\
  21.724	-0.911141372706929\\
  21.816	-0.917143704499605\\
  21.91	-0.923581674152238\\
  22.007	-0.930531755579583\\
  22.107	-0.938002793470464\\
  22.212	-0.946156928140503\\
  22.322	-0.955008602283609\\
  22.44	-0.964815740303447\\
  22.568	-0.975764832468336\\
  22.714	-0.988567593354379\\
  22.899	-1.00511561352488\\
  23.378	-1.04811234035992\\
  23.52	-1.06045615681617\\
  23.643	-1.07084636419423\\
  23.754	-1.07992206524176\\
  23.858	-1.08812138911027\\
  23.956	-1.09554308140136\\
  24.049	-1.10228432540159\\
  24.139	-1.10850458139321\\
  24.226	-1.11421380924358\\
  24.311	-1.11948650885413\\
  24.394	-1.12432957343664\\
  24.475	-1.12875271742258\\
  24.555	-1.1328170085076\\
  24.634	-1.13652518977628\\
  24.712	-1.13988171364453\\
  24.789	-1.14289259569983\\
  24.866	-1.14559843601341\\
  24.942	-1.14796687510041\\
  25.018	-1.15003303801158\\
  25.094	-1.15179603950147\\
  25.17	-1.15325607760841\\
  25.246	-1.15441441718171\\
  25.323	-1.15528268780135\\
  25.4	-1.1558471070493\\
  25.478	-1.15611367347319\\
  25.557	-1.15607641129802\\
  25.637	-1.15573070609761\\
  25.718	-1.15507339295157\\
  25.8	-1.1541028347239\\
  25.884	-1.15280172071295\\
  25.97	-1.15116113163174\\
  26.058	-1.14917451938082\\
  26.148	-1.14683791465571\\
  26.241	-1.14411930321507\\
  26.338	-1.14097734428337\\
  26.439	-1.13739989580245\\
  26.546	-1.13330099030305\\
  26.659	-1.12866446738122\\
  26.781	-1.12334928633669\\
  26.914	-1.11724566665225\\
  27.064	-1.11004994464859\\
  27.241	-1.10124219621833\\
  27.473	-1.08937077724558\\
  28.309	-1.04633057144506\\
  28.582	-1.0327352707585\\
  28.853	-1.01954510839014\\
  29.127	-1.00651437888407\\
  29.392	-0.99421351708488\\
  29.633	-0.983325360706342\\
  29.845	-0.974046129963028\\
  30	-0.967488635429692\\
};
\addlegendentry{$\dot{\theta}_2$}

\addplot [color=mycolor3]
  table[row sep=crcr]{%
  0.199000000000002	-0.891281311941228\\
  0.280000000000001	-0.892192913276123\\
  0.361999999999998	-0.893422133152914\\
  0.445	-0.894974274894825\\
  0.529	-0.896853540015954\\
  0.614000000000001	-0.899062422623338\\
  0.699999999999999	-0.901601614092623\\
  0.788	-0.904504598140115\\
  0.878	-0.907778161378612\\
  0.971	-0.911467620699341\\
  1.067	-0.915583409416886\\
  1.167	-0.920179329171425\\
  1.271	-0.925265789129888\\
  1.381	-0.930953156971601\\
  1.499	-0.93736502096348\\
  1.627	-0.944632778045314\\
  1.769	-0.953008148163114\\
  1.934	-0.963054788718466\\
  2.15	-0.976533697459541\\
  2.717	-1.01206032338752\\
  2.898	-1.02300717810828\\
  3.057	-1.03232300767769\\
  3.202	-1.04051767817282\\
  3.338	-1.04790059859634\\
  3.466	-1.05454714942375\\
  3.588	-1.06058061339642\\
  3.705	-1.06606504107715\\
  3.818	-1.07105866876991\\
  3.927	-1.07557239648431\\
  4.033	-1.07965724609371\\
  4.136	-1.0833210151891\\
  4.236	-1.08657371339099\\
  4.334	-1.0894552894693\\
  4.429	-1.09194566149177\\
  4.522	-1.09408209119528\\
  4.614	-1.09589040087587\\
  4.704	-1.0973551844161\\
  4.793	-1.0984982697967\\
  4.881	-1.09932135294516\\
  4.967	-1.09982328485817\\
  5.053	-1.1000198806904\\
  5.138	-1.09990887228211\\
  5.222	-1.09949681209517\\
  5.306	-1.09878112159355\\
  5.39	-1.09775987392695\\
  5.474	-1.09643234569475\\
  5.558	-1.09479909273488\\
  5.642	-1.09286201616373\\
  5.727	-1.09059598655294\\
  5.813	-1.08799516352463\\
  5.9	-1.08505556678333\\
  5.988	-1.08177529127161\\
  6.078	-1.07811234879254\\
  6.17	-1.07405978079786\\
  6.264	-1.06961407631369\\
  6.362	-1.06467177398117\\
  6.464	-1.05922027553667\\
  6.572	-1.05313901373324\\
  6.688	-1.04629586055972\\
  6.815	-1.03849185272344\\
  6.961	-1.02920432357057\\
  7.147	-1.01704644427782\\
  7.652	-0.983874515730701\\
  7.804	-0.974298615104651\\
  7.938	-0.96615947915231\\
  8.061	-0.958991284757282\\
  8.177	-0.952534354570993\\
  8.288	-0.946659886331329\\
  8.395	-0.941300796522007\\
  8.499	-0.936395205272866\\
  8.601	-0.931888532599036\\
  8.701	-0.927774616606975\\
  8.8	-0.924008312744093\\
  8.897	-0.920621719349377\\
  8.993	-0.917572477504102\\
  9.088	-0.91485667915715\\
  9.183	-0.912445736929971\\
  9.277	-0.910364349379904\\
  9.371	-0.908589192782539\\
  9.464	-0.90713754590961\\
  9.557	-0.905991886939219\\
  9.649	-0.905162353441472\\
  9.741	-0.904637541660325\\
  9.833	-0.904419856096251\\
  9.924	-0.904508880765356\\
  10.015	-0.90490259770819\\
  10.106	-0.905602790605634\\
  10.196	-0.906598204849935\\
  10.286	-0.907895975770277\\
  10.376	-0.909496785503389\\
  10.466	-0.91140079401308\\
  10.556	-0.913607536321855\\
  10.647	-0.916145371241267\\
  10.738	-0.918989362858387\\
  10.829	-0.922136397831601\\
  10.921	-0.925621759882571\\
  11.014	-0.929449680016972\\
  11.108	-0.933622675546893\\
  11.204	-0.938190423670417\\
  11.302	-0.943160423418732\\
  11.402	-0.948537016708826\\
  11.506	-0.954437140678763\\
  11.614	-0.96087368188407\\
  11.727	-0.967915998990804\\
  11.848	-0.975766494414632\\
  11.98	-0.984641376469138\\
  12.131	-0.995107741234353\\
  12.325	-1.00888027398656\\
  12.785	-1.04165296506488\\
  12.935	-1.05195684118527\\
  13.065	-1.06058484573208\\
  13.183	-1.06811259678311\\
  13.293	-1.07482513971809\\
  13.397	-1.08086667335498\\
  13.496	-1.08631520126357\\
  13.592	-1.09129435908596\\
  13.685	-1.09581357481181\\
  13.776	-1.09992992142484\\
  13.865	-1.10365003699125\\
  13.952	-1.10698316721361\\
  14.038	-1.10997378485401\\
  14.123	-1.11262451762991\\
  14.207	-1.11493948551258\\
  14.29	-1.11692414038833\\
  14.373	-1.11860351396323\\
  14.455	-1.11995952783816\\
  14.537	-1.12101179226818\\
  14.619	-1.12175866764809\\
  14.701	-1.1221992923071\\
  14.783	-1.12233355169037\\
  14.865	-1.12216204749285\\
  14.947	-1.1216860672252\\
  15.029	-1.12090755462966\\
  15.112	-1.11981409040667\\
  15.195	-1.11841659980306\\
  15.279	-1.11669671570337\\
  15.364	-1.11464819240695\\
  15.449	-1.11229508490554\\
  15.535	-1.10961046051867\\
  15.622	-1.10659060748948\\
  15.71	-1.10323279254569\\
  15.8	-1.09949209931274\\
  15.891	-1.09540460255729\\
  15.984	-1.09092139665678\\
  16.079	-1.08603524665077\\
  16.176	-1.08074095138054\\
  16.276	-1.07497647618104\\
  16.379	-1.06873280638234\\
  16.486	-1.06193978441183\\
  16.598	-1.05452093248352\\
  16.716	-1.04639582851627\\
  16.842	-1.03741088791302\\
  16.98	-1.02725871305877\\
  17.137	-1.01539319997078\\
  17.333	-1.00025617429311\\
  17.896	-0.95658418805192\\
  18.047	-0.945293716641288\\
  18.179	-0.935723353226503\\
  18.299	-0.927323286290175\\
  18.411	-0.919784996990977\\
  18.517	-0.912953925649607\\
  18.618	-0.906748347613615\\
  18.715	-0.901091270071284\\
  18.809	-0.895912795261992\\
  18.9	-0.891202010871044\\
  18.989	-0.886897949848134\\
  19.076	-0.882993825394664\\
  19.162	-0.879441085600345\\
  19.246	-0.876275467184715\\
  19.329	-0.873452088926822\\
  19.411	-0.870967801026129\\
  19.492	-0.868818198304609\\
  19.572	-0.866997718553062\\
  19.652	-0.865482705680641\\
  19.731	-0.864290244533841\\
  19.81	-0.863402518696866\\
  19.889	-0.862821764645258\\
  19.967	-0.862550950765073\\
  20.045	-0.862581447960679\\
  20.123	-0.862913151007476\\
  20.202	-0.863555252952786\\
  20.281	-0.864503788176204\\
  20.36	-0.865756308724247\\
  20.44	-0.867331166249155\\
  20.52	-0.869210303991395\\
  20.601	-0.871417887004522\\
  20.683	-0.873959227474575\\
  20.766	-0.876838478603755\\
  20.85	-0.880058538954451\\
  20.935	-0.88362095812899\\
  21.022	-0.887572983259446\\
  21.111	-0.891923338522364\\
  21.202	-0.896678502923706\\
  21.296	-0.901899578112385\\
  21.393	-0.907597075838098\\
  21.493	-0.913777563189885\\
  21.598	-0.920575376999089\\
  21.709	-0.928072359108722\\
  21.827	-0.936351936221659\\
  21.955	-0.945642300671143\\
  22.099	-0.956406178658966\\
  22.272	-0.969657282033335\\
  22.54	-0.990537381879335\\
  22.83	-1.01304440157594\\
  23	-1.0259301424494\\
  23.144	-1.03654543288211\\
  23.274	-1.04582593923878\\
  23.394	-1.05409024242656\\
  23.507	-1.06157079371823\\
  23.615	-1.06841753450982\\
  23.719	-1.07470651656432\\
  23.819	-1.08045164922093\\
  23.917	-1.08577683446104\\
  24.012	-1.09063538842146\\
  24.105	-1.09508869691673\\
  24.196	-1.09914434618107\\
  24.286	-1.10285124770032\\
  24.375	-1.10621010384362\\
  24.462	-1.10919033196151\\
  24.548	-1.11183452457911\\
  24.633	-1.11414680805924\\
  24.718	-1.11615403909877\\
  24.802	-1.11783305037072\\
  24.886	-1.11920498876454\\
  24.969	-1.12025541543208\\
  25.052	-1.1209995588356\\
  25.134	-1.12143171171625\\
  25.216	-1.12156105693272\\
  25.298	-1.12138661358462\\
  25.38	-1.12090805596632\\
  25.462	-1.1201257406452\\
  25.544	-1.11904073498536\\
  25.627	-1.11763609843883\\
  25.71	-1.11592587606409\\
  25.794	-1.11388751778097\\
  25.878	-1.11154432822621\\
  25.963	-1.10886857247126\\
  26.049	-1.10585627376138\\
  26.136	-1.10250492037207\\
  26.225	-1.09876998546627\\
  26.316	-1.0946424217756\\
  26.409	-1.09011577164279\\
  26.504	-1.08518657553096\\
  26.603	-1.07974183380195\\
  26.706	-1.07376808867368\\
  26.814	-1.06719634948597\\
  26.929	-1.05989070201044\\
  27.054	-1.05164094260222\\
  27.195	-1.04202348189267\\
  27.368	-1.02990181032408\\
  27.999	-0.985384750925938\\
  28.139	-0.976001333913121\\
  28.264	-0.967923812934586\\
  28.38	-0.960727775701535\\
  28.49	-0.95420496628094\\
  28.596	-0.948223431970618\\
  28.698	-0.942770569644537\\
  28.798	-0.937730260471955\\
  28.895	-0.933144024207774\\
  28.991	-0.928909943793222\\
  29.085	-0.925066939060656\\
  29.178	-0.921566929447426\\
  29.271	-0.918373216835782\\
  29.363	-0.915519751164659\\
  29.454	-0.913000025671732\\
  29.545	-0.910783876888097\\
  29.636	-0.908873047896169\\
  29.727	-0.907268627028014\\
  29.818	-0.905971139210536\\
  29.909	-0.904980629268774\\
  30	-0.904296735476748\\
};
\addlegendentry{$\dot{\theta}_3$}

\end{axis}
\end{tikzpicture}%

%% file: TikZ/AsyCoSPM_1_theta_s1.tex
%
%
\definecolor{mycolor1}{rgb}{0.00000,0.44700,0.74100}%
\definecolor{mycolor2}{rgb}{0.85000,0.32500,0.09800}%
\definecolor{mycolor3}{rgb}{0.92900,0.69400,0.12500}%
\begin{tikzpicture}

\begin{axis}[%
width=4.521in,
height=2.5in,
at={(0.758in,0.481in)},
scale only axis,
xmin=0,
xmax=30,
xlabel style={font=\color{white!15!black}},
xlabel={$t$ $\mathrm{[s]}$},
ymin=1.3,
ymax=2.2,
ylabel style={font=\color{white!15!black}},
ylabel={$\bm{\theta}$ $\mathrm{[rad]}$},
axis background/.style={fill=white},
xmajorgrids,
ymajorgrids,
legend style={legend cell align=left, align=left, draw=white!15!black}
]
\addplot [color=mycolor1]
  table[row sep=crcr]{%
  0	1.5707963267949\\
  0.00199999999999889	1.57165777252033\\
  0.00600000000000023	1.57446371597491\\
  0.00799999999999912	1.57472302047232\\
  0.0130000000000017	1.57446886365903\\
  0.0210000000000008	1.5753557039057\\
  0.0279999999999987	1.57614280294976\\
  0.0489999999999995	1.57911729847644\\
  0.097999999999999	1.58713206323393\\
  0.312000000000001	1.62235379982451\\
  0.466999999999999	1.64714527089435\\
  0.605	1.66853526299408\\
  0.733000000000001	1.68769149728887\\
  0.853999999999999	1.70511376152115\\
  0.969999999999999	1.72112605446528\\
  1.082	1.73589398468528\\
  1.19	1.74944910790284\\
  1.296	1.76206469663212\\
  1.4	1.77375092191621\\
  1.502	1.78452370057847\\
  1.603	1.79449912374527\\
  1.702	1.80359285748069\\
  1.801	1.8119958099841\\
  1.899	1.8196225044619\\
  1.996	1.82648512288243\\
  2.093	1.83265790087281\\
  2.189	1.83808234012539\\
  2.285	1.84282146995413\\
  2.381	1.84687229507972\\
  2.477	1.85023283970693\\
  2.573	1.85290204030738\\
  2.669	1.85487965662995\\
  2.765	1.85616620306526\\
  2.861	1.85676290198785\\
  2.957	1.85667166016954\\
  3.053	1.85589506881179\\
  3.149	1.85443642719564\\
  3.246	1.85227398118542\\
  3.343	1.849424378483\\
  3.441	1.84585368457769\\
  3.539	1.84159498929285\\
  3.638	1.83660296350761\\
  3.738	1.83086700364492\\
  3.838	1.82444625825999\\
  3.94	1.81720629420072\\
  4.043	1.80920484361151\\
  4.148	1.8003547531542\\
  4.255	1.79064117077399\\
  4.364	1.78005485567579\\
  4.476	1.76848693263643\\
  4.592	1.75581306985774\\
  4.713	1.74189768910727\\
  4.841	1.72647668849201\\
  4.978	1.70926913374332\\
  5.129	1.68959764611368\\
  5.305	1.66595458762551\\
  5.553	1.63187479752277\\
  5.906	1.58342713541105\\
  6.077	1.56069294855673\\
  6.22	1.54236250804496\\
  6.348	1.52663978809965\\
  6.466	1.5128327384728\\
  6.576	1.50064293359983\\
  6.681	1.4896909052058\\
  6.782	1.47984497954701\\
  6.879	1.47107405032726\\
  6.973	1.46325629450879\\
  7.065	1.45629042629035\\
  7.155	1.45016281788021\\
  7.243	1.44485423012818\\
  7.33	1.44029198383103\\
  7.416	1.43647150855856\\
  7.501	1.43338471525208\\
  7.585	1.4310202505911\\
  7.668	1.42936375979446\\
  7.751	1.42839051765743\\
  7.834	1.42810561425846\\
  7.917	1.42851175485447\\
  8	1.42960925324121\\
  8.083	1.43139603644957\\
  8.166	1.43386766073918\\
  8.25	1.43705938569287\\
  8.334	1.44093615715613\\
  8.419	1.4455443360985\\
  8.505	1.45089334397074\\
  8.592	1.45698935939112\\
  8.681	1.46391672849497\\
  8.771	1.47160781196281\\
  8.864	1.48025012223554\\
  8.959	1.48977282691311\\
  9.057	1.50029113527371\\
  9.158	1.51182248357346\\
  9.264	1.52462229995647\\
  9.375	1.53872332326946\\
  9.493	1.55441163234946\\
  9.621	1.57213406828984\\
  9.763	1.59250432576284\\
  9.929	1.61703545458424\\
  10.156	1.65133747727343\\
  10.557	1.71198604200571\\
  10.727	1.73692839074548\\
  10.872	1.75752162599966\\
  11.003	1.77543909434903\\
  11.124	1.79130360792476\\
  11.238	1.80556782472025\\
  11.347	1.81852285108467\\
  11.452	1.83031781321062\\
  11.554	1.84108836706509\\
  11.653	1.85085749897057\\
  11.75	1.85974418853548\\
  11.845	1.8677644846044\\
  11.939	1.87501317965596\\
  12.032	1.88149402961533\\
  12.124	1.88721458919856\\
  12.215	1.89218591279962\\
  12.306	1.8964654947496\\
  12.396	1.90001210396618\\
  12.486	1.90287367367344\\
  12.576	1.90504991322318\\
  12.666	1.90654306637602\\
  12.757	1.90736312181692\\
  12.848	1.90749704020203\\
  12.94	1.90694495601777\\
  13.033	1.905697099329\\
  13.127	1.90374672583183\\
  13.222	1.90109025975146\\
  13.319	1.89768888121588\\
  13.418	1.89352512912125\\
  13.519	1.88858619766238\\
  13.623	1.88280552216161\\
  13.73	1.87616177378013\\
  13.84	1.86864108547511\\
  13.955	1.86008328527418\\
  14.075	1.85045853443122\\
  14.202	1.8395751450316\\
  14.338	1.82721936645437\\
  14.486	1.81307029388586\\
  14.652	1.79649515335354\\
  14.852	1.7758048140603\\
  15.167	1.74241740625615\\
  15.489	1.70850625725781\\
  15.683	1.688768237267\\
  15.846	1.67285873692531\\
  15.993	1.65919253658151\\
  16.128	1.64732130198052\\
  16.255	1.63683282238874\\
  16.376	1.62752212137113\\
  16.492	1.61927947069042\\
  16.604	1.61200531327077\\
  16.713	1.60561437070697\\
  16.819	1.60008770120082\\
  16.922	1.59539957647302\\
  17.023	1.59148295061622\\
  17.123	1.5882931025573\\
  17.221	1.58585196962738\\
  17.318	1.58412111365189\\
  17.414	1.58309429334831\\
  17.509	1.58276228211278\\
  17.604	1.58312041537571\\
  17.698	1.58416109131122\\
  17.792	1.58588899409574\\
  17.886	1.58830617624949\\
  17.98	1.5914121584878\\
  18.074	1.59520388027545\\
  18.169	1.5997268631325\\
  18.264	1.6049358439054\\
  18.36	1.6108849228646\\
  18.457	1.61758077846739\\
  18.556	1.6251057979795\\
  18.656	1.63339296524401\\
  18.759	1.64262197357301\\
  18.864	1.65272112302393\\
  18.972	1.66379684771321\\
  19.084	1.67597171822696\\
  19.201	1.68938037358714\\
  19.325	1.70428696409817\\
  19.458	1.72097490893951\\
  19.604	1.73999582445823\\
  19.773	1.76272777196836\\
  19.994	1.79319735304123\\
  20.461	1.85775723389872\\
  20.629	1.88015704638823\\
  20.772	1.89854567771718\\
  20.901	1.91445400641688\\
  21.021	1.92856652242678\\
  21.134	1.94116761927818\\
  21.241	1.95241697646095\\
  21.344	1.96256257988759\\
  21.444	1.97172383855768\\
  21.541	1.97992111086388\\
  21.635	1.98718281383196\\
  21.727	1.99361040397671\\
  21.818	1.99928104840085\\
  21.907	2.00414420110479\\
  21.995	2.00827013984106\\
  22.082	2.01166636773248\\
  22.168	2.01434345945131\\
  22.254	2.01633399520933\\
  22.339	2.0176188747405\\
  22.424	2.01821967488056\\
  22.509	2.01813311288169\\
  22.594	2.01735817039039\\
  22.679	2.01589609912054\\
  22.764	2.0137504170785\\
  22.85	2.01088967316046\\
  22.936	2.00734335280957\\
  23.023	2.00306905679151\\
  23.111	1.99805655005501\\
  23.2	1.99229871316896\\
  23.29	1.98579178003282\\
  23.382	1.97845214159523\\
  23.476	1.97026182329941\\
  23.572	1.96120823721743\\
  23.671	1.95118015823059\\
  23.773	1.94015798444682\\
  23.879	1.92801382764488\\
  23.991	1.91448278109894\\
  24.109	1.89952854749503\\
  24.236	1.88273438376334\\
  24.376	1.86351783670098\\
  24.538	1.84056825598944\\
  24.751	1.80964846266165\\
  25.212	1.74253845107216\\
  25.376	1.71949596547136\\
  25.517	1.7003676067609\\
  25.644	1.68382024359566\\
  25.762	1.66912703491382\\
  25.874	1.65586819677221\\
  25.981	1.64389109171678\\
  26.084	1.63305169989252\\
  26.184	1.62322006135239\\
  26.281	1.61437193829734\\
  26.376	1.60639471560848\\
  26.469	1.59927140393399\\
  26.561	1.59291425324213\\
  26.652	1.58731886233805\\
  26.742	1.58247687964166\\
  26.831	1.57837628721887\\
  26.92	1.57496725352785\\
  27.008	1.57228117601538\\
  27.096	1.57027764491849\\
  27.184	1.56895552461101\\
  27.273	1.568307344907\\
  27.362	1.56834492082026\\
  27.452	1.56906999622032\\
  27.543	1.57049196065921\\
  27.635	1.572616460604\\
  27.729	1.57547920508088\\
  27.824	1.57905773165196\\
  27.922	1.58344059049691\\
  28.022	1.58860054600902\\
  28.126	1.59466043294121\\
  28.234	1.60165000037956\\
  28.347	1.60966040049122\\
  28.466	1.618790901641\\
  28.594	1.62931167772317\\
  28.734	1.64152328610886\\
  28.892	1.65601222400985\\
  29.085	1.67443450452621\\
  29.416	1.70686158866441\\
  29.698	1.73418497841882\\
  29.887	1.75181179431466\\
  30	1.7619551484841\\
};
\addlegendentry{$\theta_1$}

\addplot [color=mycolor2]
  table[row sep=crcr]{%
  0	1.5707963267949\\
  0.00199999999999889	1.57165777252033\\
  0.00499999999999901	1.57388423018782\\
  0.00700000000000145	1.57432923724545\\
  0.0100000000000016	1.57361025457583\\
  0.0130000000000017	1.57321394847064\\
  0.0199999999999996	1.57317645506857\\
  0.0259999999999998	1.57300711714826\\
  0.0549999999999997	1.57372328031714\\
  0.114000000000001	1.57654625010292\\
  0.300000000000001	1.58537199308173\\
  0.462	1.59236982168329\\
  0.614999999999998	1.59829596587436\\
  0.762	1.60330354918513\\
  0.905000000000001	1.60748849902441\\
  1.046	1.6109264046585\\
  1.186	1.61365065848646\\
  1.326	1.61568671787856\\
  1.467	1.61705114445653\\
  1.611	1.6177541604823\\
  1.758	1.61778372714318\\
  1.91	1.61712724940542\\
  2.069	1.61575158642474\\
  2.238	1.61359419323725\\
  2.419	1.61058921317276\\
  2.618	1.60658901108203\\
  2.844	1.60134432363331\\
  3.116	1.59431894892494\\
  3.492	1.5838617714473\\
  4.428	1.55758254866133\\
  4.752	1.54931539894653\\
  5.019	1.54317223099336\\
  5.249	1.53855293875366\\
  5.453	1.5351315838786\\
  5.637	1.53272079798892\\
  5.806	1.5311817009981\\
  5.963	1.53042632086618\\
  6.111	1.53039178996709\\
  6.251	1.53103608325539\\
  6.385	1.53233190782696\\
  6.514	1.53426176935142\\
  6.639	1.53681837255323\\
  6.76	1.53997818459971\\
  6.878	1.54374401000487\\
  6.993	1.54809384365903\\
  7.107	1.55309281289516\\
  7.219	1.558689841613\\
  7.33	1.56492242360512\\
  7.44	1.57178221419167\\
  7.55	1.57932762035687\\
  7.66	1.58755906073648\\
  7.771	1.59655592080004\\
  7.883	1.60632638799932\\
  7.996	1.61687228635404\\
  8.112	1.62839285883015\\
  8.231	1.64090942249457\\
  8.354	1.65454529546287\\
  8.482	1.6694309332231\\
  8.618	1.68594504400555\\
  8.766	1.70462164012075\\
  8.933	1.72640689503515\\
  9.142	1.7544021641067\\
  9.699	1.82937818566809\\
  9.858	1.84985690671031\\
  9.996	1.86695122061597\\
  10.121	1.8817554619445\\
  10.238	1.89492611208246\\
  10.348	1.90662523070261\\
  10.453	1.91711145533913\\
  10.554	1.92651876596688\\
  10.652	1.93496678176139\\
  10.748	1.94255682069702\\
  10.842	1.94930079128101\\
  10.934	1.95521692950534\\
  11.025	1.9603820112344\\
  11.115	1.96480181069338\\
  11.204	1.96848615649945\\
  11.292	1.97144858990258\\
  11.38	1.97372806102246\\
  11.468	1.97532131334282\\
  11.556	1.976228061937\\
  11.644	1.97645094126776\\
  11.733	1.97598638850978\\
  11.822	1.97483649048286\\
  11.912	1.97298829350107\\
  12.003	1.97043354242042\\
  12.095	1.96716748380832\\
  12.189	1.96314263416158\\
  12.285	1.95834051165856\\
  12.383	1.95274779630434\\
  12.484	1.9462895504434\\
  12.588	1.93894470688421\\
  12.696	1.93062039607682\\
  12.808	1.92129529803044\\
  12.926	1.91077687629977\\
  13.052	1.89884452602667\\
  13.188	1.88526082838447\\
  13.338	1.86957393535403\\
  13.512	1.85066207146749\\
  13.736	1.82557974583062\\
  14.32	1.75986059968249\\
  14.504	1.74004905694933\\
  14.666	1.72328819906319\\
  14.814	1.70865663069312\\
  14.953	1.69559403586308\\
  15.086	1.6837775780555\\
  15.214	1.6730874934565\\
  15.339	1.66333586007291\\
  15.461	1.65450807885965\\
  15.58	1.64658171736928\\
  15.697	1.63947011560415\\
  15.813	1.63310507595573\\
  15.928	1.62748424073309\\
  16.042	1.62260210161265\\
  16.155	1.6184504403205\\
  16.267	1.61501873704392\\
  16.379	1.61227307785233\\
  16.491	1.61021779353838\\
  16.602	1.60886496532292\\
  16.713	1.60819495977355\\
  16.824	1.60820855882185\\
  16.935	1.60890566591544\\
  17.046	1.61028539917427\\
  17.157	1.6123461583001\\
  17.269	1.6151134199992\\
  17.381	1.61856867721669\\
  17.493	1.62270817928793\\
  17.606	1.62757348124602\\
  17.719	1.63312517840382\\
  17.833	1.63941491328868\\
  17.947	1.64638874711223\\
  18.062	1.65410752109299\\
  18.178	1.66257796907956\\
  18.296	1.67188630260841\\
  18.415	1.68196394537603\\
  18.536	1.6929023343942\\
  18.659	1.70471254399153\\
  18.785	1.71750462664302\\
  18.914	1.7312940369094\\
  19.047	1.7462023269077\\
  19.186	1.7624797652234\\
  19.332	1.78027521644457\\
  19.488	1.79998965526459\\
  19.659	1.82230364207242\\
  19.856	1.8487226592111\\
  20.12	1.8848773310869\\
  20.592	1.94958698488574\\
  20.783	1.97499201627326\\
  20.943	1.99559708087774\\
  21.086	2.01333565542756\\
  21.218	2.02902845985654\\
  21.342	2.04308422720285\\
  21.459	2.05566377899034\\
  21.571	2.06702312051157\\
  21.679	2.07729162753884\\
  21.783	2.08649870667999\\
  21.884	2.0947612089239\\
  21.983	2.10217602816303\\
  22.08	2.10875343397955\\
  22.175	2.11450844835522\\
  22.268	2.11946042634562\\
  22.36	2.12367470786645\\
  22.451	2.12715545909414\\
  22.541	2.12990970673174\\
  22.63	2.1319471964015\\
  22.718	2.13328024006266\\
  22.806	2.13392702252662\\
  22.893	2.13388549304765\\
  22.98	2.13316262747878\\
  23.067	2.1317562465942\\
  23.154	2.1296662801246\\
  23.241	2.12689480216881\\
  23.329	2.12340250118701\\
  23.417	2.11922394094441\\
  23.506	2.11430904023642\\
  23.596	2.10864628366469\\
  23.687	2.10222717900571\\
  23.779	2.09504651012241\\
  23.872	2.08710257790766\\
  23.967	2.07830125988361\\
  24.064	2.06862776987456\\
  24.164	2.05796289982157\\
  24.267	2.0462839049614\\
  24.374	2.0334535130777\\
  24.485	2.01944897977666\\
  24.602	2.00399087551288\\
  24.727	1.98677208229075\\
  24.862	1.96746982451246\\
  25.012	1.94531339557144\\
  25.187	1.91874821941643\\
  25.424	1.8820202431547\\
  25.886	1.81031350234036\\
  26.073	1.78208636589713\\
  26.234	1.75846685647597\\
  26.381	1.73758727075732\\
  26.519	1.71867459103436\\
  26.65	1.70140581495666\\
  26.777	1.68535248756563\\
  26.9	1.67049019571044\\
  27.021	1.65655851122398\\
  27.14	1.64354730104515\\
  27.257	1.63143985888915\\
  27.373	1.62011947925881\\
  27.489	1.6094888227686\\
  27.604	1.59963630527378\\
  27.719	1.59047013503215\\
  27.834	1.58199075062614\\
  27.949	1.57419662405993\\
  28.065	1.56702578741701\\
  28.181	1.56054458463073\\
  28.297	1.55474767457146\\
  28.414	1.54958791440785\\
  28.531	1.54511177395289\\
  28.649	1.54128304280643\\
  28.768	1.53811230415918\\
  28.887	1.53562798130907\\
  29.007	1.53381088215241\\
  29.127	1.53267774260846\\
  29.248	1.53222044819771\\
  29.37	1.53244836790207\\
  29.492	1.53335981362538\\
  29.615	1.53496137725038\\
  29.739	1.53725902677818\\
  29.865	1.54028335178008\\
  29.992	1.54401955290598\\
  30	1.54427761554752\\
};
\addlegendentry{$\theta_2$}

\addplot [color=mycolor3]
  table[row sep=crcr]{%
  0	1.5707963267949\\
  0.00199999999999889	1.57165777252033\\
  0.00499999999999901	1.5739315895574\\
  0.00700000000000145	1.57451349937535\\
  0.0100000000000016	1.57404798483656\\
  0.0130000000000017	1.57383886086165\\
  0.0210000000000008	1.57422268719631\\
  0.0279999999999987	1.57458270169665\\
  0.0479999999999983	1.57620808390604\\
  0.0919999999999987	1.58084090810817\\
  0.436	1.61791707082858\\
  0.611000000000001	1.63599002800416\\
  0.763000000000002	1.65101172503093\\
  0.902000000000001	1.66407221749589\\
  1.033	1.67569693441973\\
  1.157	1.68601810863947\\
  1.276	1.69524330667156\\
  1.392	1.70354984464905\\
  1.505	1.71095340923373\\
  1.615	1.71747839271884\\
  1.724	1.7232561134793\\
  1.831	1.72824245744896\\
  1.937	1.73249755643985\\
  2.042	1.73602931346107\\
  2.147	1.73887294341778\\
  2.251	1.74100605343628\\
  2.355	1.74245665090473\\
  2.459	1.74322497578576\\
  2.564	1.74331127572412\\
  2.669	1.74270999839474\\
  2.775	1.74141311617627\\
  2.881	1.7394329191028\\
  2.989	1.73672500363243\\
  3.098	1.73330053457645\\
  3.208	1.72915810945467\\
  3.32	1.72425256042042\\
  3.434	1.71857096315781\\
  3.551	1.7120462035604\\
  3.67	1.70472219604496\\
  3.793	1.69646216790924\\
  3.92	1.68724409994325\\
  4.053	1.67689632933153\\
  4.193	1.66530722090306\\
  4.342	1.65227847154375\\
  4.505	1.63732814156219\\
  4.692	1.61946687689062\\
  4.932	1.59580643304264\\
  5.489	1.54064706071937\\
  5.671	1.52348300421711\\
  5.827	1.50944629597895\\
  5.968	1.49743518325596\\
  6.099	1.48695260238384\\
  6.223	1.47770942073242\\
  6.342	1.46952356251017\\
  6.456	1.46236161826197\\
  6.567	1.45606939198537\\
  6.675	1.45062651936708\\
  6.781	1.4459637127645\\
  6.886	1.44203083008725\\
  6.99	1.43882608718788\\
  7.093	1.43634286410272\\
  7.195	1.43457007174181\\
  7.297	1.43348526150245\\
  7.399	1.43308967960638\\
  7.501	1.43338091940059\\
  7.604	1.43436591291517\\
  7.707	1.43603557989141\\
  7.811	1.43840294341717\\
  7.917	1.44150336988719\\
  8.025	1.44535477252842\\
  8.135	1.44997012552646\\
  8.247	1.45535707790551\\
  8.362	1.46157455780298\\
  8.481	1.4686980649326\\
  8.604	1.47674876022546\\
  8.733	1.48588397965582\\
  8.869	1.4962100781166\\
  9.014	1.50791648706084\\
  9.171	1.52128872278659\\
  9.347	1.53698294547725\\
  9.557	1.55642834363889\\
  9.872	1.58637931074936\\
  10.237	1.62092322149449\\
  10.44	1.639432853445\\
  10.61	1.65425583694576\\
  10.761	1.6667463016491\\
  10.9	1.67756827497185\\
  11.031	1.68708479136924\\
  11.155	1.69540974749696\\
  11.274	1.70271350158684\\
  11.388	1.70902978070517\\
  11.499	1.71449637135576\\
  11.607	1.71913070658867\\
  11.713	1.72299062899835\\
  11.817	1.72608735520055\\
  11.919	1.72843735477556\\
  12.02	1.73007466222185\\
  12.119	1.73099830082116\\
  12.218	1.7312355552168\\
  12.316	1.7307857518672\\
  12.414	1.72964925324085\\
  12.512	1.72782368157657\\
  12.61	1.72530980989536\\
  12.708	1.72211158136109\\
  12.807	1.718193098858\\
  12.907	1.71354416709187\\
  13.008	1.70815839453366\\
  13.11	1.70203345344667\\
  13.214	1.69510143834884\\
  13.32	1.68734981534174\\
  13.429	1.67868956559564\\
  13.541	1.66910285683869\\
  13.658	1.65839201388391\\
  13.78	1.64652519230872\\
  13.908	1.63338116886145\\
  14.045	1.61861810080785\\
  14.195	1.60175297681651\\
  14.365	1.58193087591131\\
  14.575	1.55671780974753\\
  15.282	1.47124365460296\\
  15.449	1.45210304062175\\
  15.597	1.43581801811641\\
  15.733	1.4215329756317\\
  15.861	1.40877311348421\\
  15.982	1.39739357407\\
  16.098	1.38716619536434\\
  16.21	1.37797465735347\\
  16.319	1.36971650058033\\
  16.425	1.36237250460951\\
  16.529	1.35585729121888\\
  16.63	1.35021165795833\\
  16.73	1.34530741889102\\
  16.828	1.34118337405261\\
  16.925	1.33778378322829\\
  17.021	1.33510241236497\\
  17.116	1.33313024719805\\
  17.211	1.33184575639341\\
  17.305	1.33125927639229\\
  17.399	1.3313591027271\\
  17.493	1.33214848975844\\
  17.587	1.33362849233986\\
  17.681	1.33579793261963\\
  17.775	1.33865337498208\\
  17.87	1.34223035430294\\
  17.965	1.34649391238451\\
  18.061	1.35148895751485\\
  18.158	1.3572234077768\\
  18.256	1.36370198673827\\
  18.356	1.37100235557842\\
  18.458	1.37914213739287\\
  18.562	1.38813392717274\\
  18.669	1.39808078533321\\
  18.779	1.40900186466402\\
  18.893	1.4210159948045\\
  19.012	1.43425480484581\\
  19.137	1.44885783617798\\
  19.271	1.46521502382057\\
  19.416	1.48361607116619\\
  19.58	1.50513697167501\\
  19.78	1.53210914265202\\
  20.193	1.58873826287824\\
  20.421	1.61952826001544\\
  20.593	1.6420790669592\\
  20.742	1.66093797748774\\
  20.878	1.67747010125954\\
  21.005	1.69222224982049\\
  21.125	1.70547529739587\\
  21.24	1.71748786509865\\
  21.351	1.72839152366151\\
  21.458	1.73821621419691\\
  21.562	1.7470835696665\\
  21.664	1.75509642306431\\
  21.764	1.76226818811135\\
  21.863	1.76867849175912\\
  21.96	1.77427474005502\\
  22.056	1.77913045968504\\
  22.151	1.78325412161137\\
  22.246	1.78668965967907\\
  22.34	1.78940408376701\\
  22.434	1.79143189695471\\
  22.528	1.79277011575896\\
  22.622	1.79341801894409\\
  22.716	1.79337712489525\\
  22.81	1.79265116094925\\
  22.905	1.79122745298829\\
  23	1.78911834395831\\
  23.096	1.78630143905376\\
  23.193	1.78276838952824\\
  23.291	1.7785136835128\\
  23.391	1.773481105608\\
  23.492	1.76771167498949\\
  23.595	1.76114295457996\\
  23.701	1.75369056669937\\
  23.809	1.74540977600438\\
  23.921	1.73612956494449\\
  24.037	1.72582212245771\\
  24.158	1.7143729010132\\
  24.285	1.7016585966984\\
  24.42	1.68744537492369\\
  24.566	1.67137418881956\\
  24.729	1.65272502452675\\
  24.922	1.62992319417693\\
  25.202	1.59606897240556\\
  25.57	1.55167977521572\\
  25.761	1.52936024791175\\
  25.921	1.51133998100431\\
  26.065	1.4958040111\\
  26.198	1.48213954453987\\
  26.323	1.46998129341406\\
  26.442	1.45909051915358\\
  26.556	1.44933813168934\\
  26.667	1.44052874409902\\
  26.775	1.43264630593719\\
  26.88	1.42566651838889\\
  26.983	1.41950144669758\\
  27.085	1.41408400279707\\
  27.185	1.40945519399889\\
  27.284	1.40555232983198\\
  27.383	1.40233637612281\\
  27.481	1.39983670058165\\
  27.579	1.39802137214349\\
  27.677	1.39689124600016\\
  27.775	1.39644405371363\\
  27.874	1.39668027151792\\
  27.973	1.3975993672712\\
  28.073	1.3992091627864\\
  28.175	1.40154119486242\\
  28.278	1.40458409772755\\
  28.383	1.40837452648889\\
  28.49	1.4129246896685\\
  28.6	1.41829329194005\\
  28.713	1.42449934634389\\
  28.83	1.43161791941357\\
  28.951	1.43966852482188\\
  29.078	1.44880858245333\\
  29.213	1.45922247704248\\
  29.357	1.47102791353682\\
  29.515	1.48468410138682\\
  29.693	1.50077496240332\\
  29.91	1.52111132311851\\
  30	1.52969507916231\\
};
\addlegendentry{$\theta_3$}

\end{axis}
\end{tikzpicture}%

%% file: TikZ/AsyCoSPM_1_TAU_trans.tex
%
%
\definecolor{mycolor1}{rgb}{0.00000,0.44700,0.74100}%
\definecolor{mycolor2}{rgb}{0.85000,0.32500,0.09800}%
\definecolor{mycolor3}{rgb}{0.92900,0.69400,0.12500}%
\begin{tikzpicture}

\begin{axis}[%
width=4.521in,
height=2.5in,
at={(0.758in,0.481in)},
scale only axis,
xmin=0,
xmax=0.25,
xtick={0,0.05,0.1,0.15,0.2,0.25},
xlabel style={font=\color{white!15!black}},
xlabel={$t$ $\mathrm{[s]}$},
ymin=-5,
ymax=2,
ylabel style={font=\color{white!15!black}},
ylabel={$\bm{\tau}$ $\mathrm{[Nm]}$},
axis background/.style={fill=white},
xmajorgrids,
ymajorgrids,
legend style={legend cell align=left, align=left, draw=white!15!black}
]
\addplot [color=mycolor1]
  table[row sep=crcr]{%
0	1.26232381999017\\
0.000200950914520659	1.27494977384201\\
0.00099999999999989	1.30722602953107\\
0.00179333169064222	1.32003048640589\\
0.002	1.32130701733532\\
0.00299999999999989	1.31998026704314\\
0.004	1.30471096189648\\
0.00499999999999989	1.27829226456568\\
0.00600000000000001	1.24928621200457\\
0.0069999999999999	1.22516012197254\\
0.00800000000000001	1.21095833438609\\
0.0089999999999999	1.207934054544\\
0.01	1.21409693199677\\
0.0129999999999999	1.24749797674181\\
0.014	1.25299603481489\\
0.0149999999999999	1.25421541579732\\
0.016	1.25239336134369\\
0.0169999999999999	1.24932698653099\\
0.018	1.24669144350941\\
0.0189999999999999	1.2455724813432\\
0.02	1.24629525053109\\
0.0209999999999999	1.2485230474903\\
0.0229999999999999	1.25450356419703\\
0.024	1.25683190560576\\
0.0249999999999999	1.25821330582612\\
0.026	1.25867895306975\\
0.0269999999999999	1.2584990097252\\
0.0289999999999999	1.25765456536799\\
0.03	1.25754745455964\\
0.0309999999999999	1.25778915849227\\
0.032	1.25831516188026\\
0.034	1.25964335848322\\
0.0349999999999999	1.26017357101048\\
0.036	1.26052102999511\\
0.0369999999999999	1.26069563960115\\
0.0409999999999999	1.26091203370086\\
0.0429999999999999	1.26134583773166\\
0.0469999999999999	1.26245371216917\\
0.0529999999999999	1.26383398285969\\
0.0580000000000001	1.26533047065752\\
0.0649999999999999	1.26753396187454\\
0.0800000000000001	1.27264937336779\\
0.093	1.27699939121132\\
0.104	1.28048760271907\\
0.116	1.28409423655222\\
0.13	1.28810059211479\\
0.147	1.29277726168163\\
0.172	1.29945729973151\\
0.21	1.30941338749288\\
0.257	1.32155162096412\\
0.296	1.33149485271728\\
};
\addlegendentry{$\tau_1$}

\addplot [color=mycolor2]
  table[row sep=crcr]{%
0	0.699412589421265\\
0.00020095091452077	0.70147109923367\\
0.001	0.701320607428003\\
0.002	0.689623018579017\\
0.005	0.63721141756243\\
0.00600000000000001	0.627489030334756\\
0.00700000000000001	0.62550811238638\\
0.00800000000000001	0.630460942147942\\
0.00900000000000001	0.639955515986384\\
0.01	0.651027868077351\\
0.011	0.661071911313804\\
0.012	0.668460667456717\\
0.013	0.672734775277949\\
0.014	0.674413949373924\\
0.015	0.674577751745974\\
0.016	0.674396234577819\\
0.017	0.67475913401333\\
0.018	0.676086152044683\\
0.019	0.678326667410983\\
0.021	0.683896035324437\\
0.022	0.686267525428049\\
0.023	0.687949279283096\\
0.024	0.688897077726849\\
0.025	0.689249713006849\\
0.026	0.689246728372749\\
0.028	0.689108150748423\\
0.029	0.689250227427459\\
0.03	0.689558795972027\\
0.033	0.690690494706754\\
0.034	0.690882033785074\\
0.035	0.690938609633169\\
0.037	0.690788003668482\\
0.039	0.69060422822393\\
0.041	0.69059860995364\\
0.052	0.691226792640104\\
0.0570000000000001	0.691870434981194\\
0.0620000000000001	0.692643146251542\\
0.0700000000000001	0.694125465570406\\
0.086	0.697132867676006\\
0.094	0.698436704110317\\
0.102	0.699557217459937\\
0.11	0.700501538716263\\
0.119	0.701381725643507\\
0.129	0.702179803115777\\
0.141	0.702956880278929\\
0.156	0.703750997365824\\
0.178	0.704730600617611\\
0.222	0.706478880994668\\
0.296	0.709360438920293\\
};
\addlegendentry{$\tau_2$}

\addplot [color=mycolor3]
  table[row sep=crcr]{%
0	-1.96173640941144\\
0.000200950914520881	-1.98447875434383\\
0.00100000000000033	-2.22439254494554\\
0.00199999999999978	-2.71967879393447\\
0.00300000000000011	-3.29484342593854\\
0.00399999999999956	-3.78185422536271\\
0.00499999999999989	-4.03078506185707\\
0.00600000000000023	-3.98845533831962\\
0.00699999999999967	-3.68718571860136\\
0.00800000000000001	-3.22265739241461\\
0.00900000000000034	-2.71238193957325\\
0.00999999999999979	-2.25868336312534\\
0.0110000000000001	-1.924106336418\\
0.0119999999999996	-1.72438840689812\\
0.0129999999999999	-1.6364427337716\\
0.0140000000000002	-1.61541983739749\\
0.0149999999999997	-1.61355090006528\\
0.016	-1.59485631283069\\
0.0170000000000003	-1.54249203143905\\
0.0179999999999998	-1.45849775831271\\
0.0199999999999996	-1.26100683163792\\
0.0209999999999999	-1.18489222494422\\
0.0220000000000002	-1.13955072261588\\
0.0229999999999997	-1.12604248933706\\
0.024	-1.13820009880321\\
0.0250000000000004	-1.16598045429189\\
0.0270000000000001	-1.22997575614866\\
0.0279999999999996	-1.25481665257251\\
0.0289999999999999	-1.2737460949712\\
0.0309999999999997	-1.3056583046463\\
0.032	-1.32546071017809\\
0.0330000000000004	-1.35056968506244\\
0.0339999999999998	-1.38097325452824\\
0.0359999999999996	-1.45142479601669\\
0.0380000000000003	-1.52140719370766\\
0.04	-1.58213348864092\\
0.0419999999999998	-1.6353807258853\\
0.0460000000000003	-1.7358641180432\\
0.048	-1.78230036970729\\
0.0499999999999998	-1.82310925366156\\
0.0519999999999996	-1.85808278269195\\
0.0540000000000003	-1.88848163262358\\
0.056	-1.91536263650481\\
0.0579999999999998	-1.93889894095804\\
0.0590000000000002	-1.94933807390968\\
0.0599999999999996	-1.95887020321651\\
0.0609999999999999	-1.96751583628579\\
0.0620000000000003	-1.97532349867036\\
0.0629999999999997	-1.98235944775958\\
0.0640000000000001	-1.98869469598643\\
0.0650000000000004	-1.99439380293635\\
0.0659999999999998	-1.99950824110916\\
0.0670000000000002	-2.00407508088623\\
0.0679999999999996	-2.00811998752925\\
0.069	-2.01166251541414\\
0.0700000000000003	-2.01472152863496\\
0.0709999999999997	-2.01731910064716\\
0.0720000000000001	-2.0194821204048\\
0.0730000000000004	-2.0212417105325\\
0.0739999999999998	-2.02263118470126\\
0.0750000000000002	-2.02368351889769\\
0.0759999999999996	-2.02442920753504\\
0.077	-2.02489504092838\\
0.0780000000000003	-2.02510393330635\\
0.0789999999999997	-2.02507558910816\\
0.0800000000000001	-2.02482760368456\\
0.0810000000000004	-2.02437657085348\\
0.0819999999999999	-2.02373887808432\\
0.0830000000000002	-2.02293104464124\\
0.0839999999999996	-2.02196963039733\\
0.0860000000000003	-2.01965017631919\\
0.0880000000000001	-2.01689871803254\\
0.0899999999999999	-2.01381312975993\\
0.093	-2.00872890820717\\
0.0970000000000004	-2.00143725265302\\
0.107	-1.98292265963999\\
0.111	-1.97598263917147\\
0.114	-1.9710642567998\\
0.117	-1.96641573661138\\
0.12	-1.96204673586918\\
0.123	-1.95795838003265\\
0.126	-1.95414515251476\\
0.129	-1.95059688844433\\
0.132	-1.94730030673526\\
0.135	-1.94424004734709\\
0.138	-1.94139954008182\\
0.141	-1.93876173260074\\
0.144	-1.93630960447599\\
0.148	-1.93330028545336\\
0.152	-1.93055408867997\\
0.156	-1.92803633436258\\
0.16	-1.92571551515779\\
0.165	-1.92304876659786\\
0.17	-1.92059789589438\\
0.175	-1.91832232486047\\
0.181	-1.915775753493\\
0.188	-1.9129997166578\\
0.196	-1.91001437463305\\
0.206	-1.90647347984824\\
0.219	-1.90206438064603\\
0.238	-1.89582104093461\\
0.273	-1.88453444744533\\
0.296	-1.87715504463241\\
};
\addlegendentry{$\tau_3$}

\end{axis}
\end{tikzpicture}%

%% file: TikZ/AsyCoSPM_1_TAU_perm.tex
%
%
\definecolor{mycolor1}{rgb}{0.00000,0.44700,0.74100}%
\definecolor{mycolor2}{rgb}{0.85000,0.32500,0.09800}%
\definecolor{mycolor3}{rgb}{0.92900,0.69400,0.12500}%
\begin{tikzpicture}

\begin{axis}[%
width=4.521in,
height=2.5in,
at={(0.758in,0.481in)},
scale only axis,
xmin=0.25,
xmax=30,
xlabel style={font=\color{white!15!black}},
xlabel={$t$ $\mathrm{[s]}$},
ymin=-2.5,
ymax=2,
ylabel style={font=\color{white!15!black}},
ylabel={$\bm{\tau}$ $\mathrm{[Nm]}$},
axis background/.style={fill=white},
xmajorgrids,
ymajorgrids,
legend style={legend cell align=left, align=left, draw=white!15!black}
]
\addplot [color=mycolor1]
  table[row sep=crcr]{%
0.196000000000002	1.30576325383851\\
0.405999999999999	1.35886591413619\\
0.597000000000001	1.4037005667367\\
0.774000000000001	1.44173859343161\\
0.942	1.47434992656021\\
1.105	1.5025102105782\\
1.266	1.52683592474419\\
1.427	1.54765911339586\\
1.589	1.56512232122669\\
1.754	1.57943338213844\\
1.924	1.5907071093716\\
2.101	1.59897397182687\\
2.287	1.6041899614223\\
2.484	1.60624185066807\\
2.694	1.60493898580137\\
2.915	1.60009809969239\\
3.145	1.59159862782715\\
3.377	1.57959196913486\\
3.607	1.56424869241189\\
3.83	1.54594347237082\\
4.047	1.52468816445665\\
4.259	1.50046358226474\\
4.468	1.47312081172188\\
4.678	1.44216607780988\\
4.893	1.40696322389117\\
5.118	1.36659142222554\\
5.363	1.31905855031718\\
5.649	1.25992710673563\\
6.092	1.16427036262922\\
6.528	1.07129414085931\\
6.792	1.01852175688251\\
7.015	0.977389116273876\\
7.216	0.943772282090027\\
7.402	0.916110805976711\\
7.578	0.893382047591974\\
7.747	0.875019192121155\\
7.91	0.860765396144949\\
8.069	0.850325598452745\\
8.224	0.843591053255231\\
8.377	0.840388550015412\\
8.528	0.840660301107334\\
8.679	0.844384651622029\\
8.83	0.851569508720516\\
8.982	0.862267458326372\\
9.136	0.876580835605779\\
9.293	0.894653514187265\\
9.455	0.916805378015859\\
9.623	0.943280168460895\\
9.801	0.974857693739761\\
9.994	1.01266508027664\\
10.211	1.0587791796765\\
10.48	1.11966047062696\\
11.295	1.30658836335513\\
11.51	1.35084337749525\\
11.702	1.38688329854723\\
11.88	1.41681839875205\\
12.05	1.44192309694579\\
12.214	1.46267870114578\\
12.375	1.47960473102096\\
12.535	1.49297599640873\\
12.696	1.50296705977279\\
12.859	1.50961886755629\\
13.026	1.51296697311323\\
13.199	1.512959850625\\
13.38	1.50948016470038\\
13.573	1.5022837604736\\
13.783	1.49094395604065\\
14.018	1.47472320397191\\
14.295	1.45203165262708\\
14.666	1.41792117124175\\
15.693	1.32177552445204\\
16.063	1.29164687582069\\
16.408	1.26697887957187\\
16.744	1.24639038305401\\
17.077	1.2294270295268\\
17.407	1.21605024394729\\
17.731	1.2063459359122\\
18.043	1.20041672573116\\
18.34	1.19818099759735\\
18.622	1.19946416892881\\
18.893	1.20412290804269\\
19.157	1.21210430194857\\
19.42	1.22351679080216\\
19.69	1.23871944213487\\
19.98	1.25856716991671\\
20.325	1.28579779680942\\
21.27	1.36248240540806\\
21.519	1.37785364943014\\
21.738	1.38796611050106\\
21.94	1.39387903912428\\
22.132	1.39607205352228\\
22.318	1.39475928822049\\
22.501	1.39002503453177\\
22.684	1.38183385737529\\
22.869	1.37009626609866\\
23.06	1.35449949615783\\
23.261	1.33457936409317\\
23.478	1.30955160597722\\
23.726	1.27736691816577\\
24.05	1.23157551678913\\
24.716	1.13665076595891\\
24.968	1.10484151883268\\
25.187	1.08062543678653\\
25.389	1.06173497224307\\
25.58	1.04730371579604\\
25.766	1.03670106506262\\
25.949	1.02973120981154\\
26.131	1.02625409345655\\
26.315	1.02620905835259\\
26.502	1.02962768525094\\
26.695	1.03662530365527\\
26.897	1.04743814495194\\
27.111	1.06239190208983\\
27.342	1.08204321608837\\
27.597	1.10725606377291\\
27.889	1.13967889009474\\
28.243	1.18259910951014\\
28.733	1.24577080320624\\
29.935	1.40193543380799\\
30	1.40996823626956\\
};
\addlegendentry{$\tau_1$}

\addplot [color=mycolor2]
  table[row sep=crcr]{%
0.196000000000002	0.705460751162924\\
1.701	0.763411433641281\\
2.081	0.773118580604976\\
2.432	0.77867022639364\\
2.773	0.780625924213805\\
3.114	0.779133237272436\\
3.464	0.774142607544324\\
3.836	0.765351583925973\\
4.258	0.751832075550201\\
4.875	0.728170168520705\\
5.559	0.702827816220193\\
5.98	0.690784887788631\\
6.378	0.682872993166708\\
6.819	0.677678694666241\\
8.497	0.661731381373517\\
9.132	0.648814449583899\\
10	0.631469830512859\\
10.443	0.626267453278068\\
10.826	0.625154293376848\\
11.179	0.627526882463936\\
11.517	0.633217928501839\\
11.854	0.642345074409491\\
12.21	0.655486517185011\\
12.643	0.675121488029959\\
13.512	0.715387403470736\\
13.824	0.725685644013339\\
14.099	0.731346686387855\\
14.358	0.733248650271438\\
14.613	0.731674642157749\\
14.874	0.726589850678319\\
15.154	0.717619466285598\\
15.48	0.703588671964472\\
16.058	0.674429764286259\\
16.463	0.65586566323454\\
16.756	0.645850594779137\\
17.018	0.640307975047826\\
17.27	0.638422914518866\\
17.524	0.639991751700119\\
17.794	0.64515477422546\\
18.108	0.654731969985075\\
18.633	0.674880314770135\\
19.074	0.690263836211145\\
19.382	0.697560631942988\\
19.66	0.700724842318547\\
19.93	0.700347738539485\\
20.204	0.696500877687583\\
20.497	0.688883820277361\\
20.833	0.676587661939941\\
21.316	0.655062951550381\\
21.953	0.627207928562704\\
22.305	0.615426975788441\\
22.618	0.608370334440981\\
22.919	0.605028114624023\\
23.222	0.605131382631942\\
23.539	0.608721129931972\\
23.889	0.616203851382988\\
24.316	0.628954288244675\\
25.641	0.671068581808278\\
25.997	0.677292248226742\\
26.335	0.679758192646606\\
26.686	0.678824530724409\\
27.115	0.674024900310645\\
27.965	0.663735925953254\\
28.296	0.663860043349107\\
28.6	0.66741424629501\\
28.904	0.67443784703568\\
29.233	0.685558764877488\\
29.643	0.703085151170054\\
30	0.719764050633785\\
};
\addlegendentry{$\tau_2$}

\addplot [color=mycolor3]
  table[row sep=crcr]{%
0.196000000000002	-1.91001437463305\\
0.260000000000002	-1.88871266594782\\
0.728000000000002	-1.73994455389564\\
0.972000000000001	-1.6658999476222\\
1.172	-1.60873218489678\\
1.349	-1.56167495420577\\
1.511	-1.52213965781104\\
1.662	-1.48880341179976\\
1.805	-1.46073351840116\\
1.942	-1.43733741424085\\
2.074	-1.41827704010243\\
2.202	-1.40325639519113\\
2.327	-1.39203660935368\\
2.45	-1.38444713644422\\
2.572	-1.38039391100153\\
2.693	-1.37985498753587\\
2.813	-1.38277789845707\\
2.933	-1.38914835195349\\
3.054	-1.39903780063178\\
3.176	-1.41247708411831\\
3.3	-1.42961662098621\\
3.427	-1.45068141651035\\
3.557	-1.47575555951832\\
3.692	-1.50533554984969\\
3.833	-1.53980024597348\\
3.982	-1.57981840588813\\
4.142	-1.62642386267373\\
4.318	-1.6813669858501\\
4.52	-1.74816607888405\\
4.777	-1.83702433302606\\
5.48	-2.08208200293462\\
5.686	-2.14900775245165\\
5.866	-2.20391607940077\\
6.03	-2.25039344544062\\
6.184	-2.29048323449411\\
6.33	-2.3249537125258\\
6.47	-2.35449534003484\\
6.606	-2.37967912181414\\
6.738	-2.40063739123905\\
6.868	-2.41778794557632\\
6.996	-2.43118962195607\\
7.123	-2.44099831839739\\
7.249	-2.44725421119618\\
7.375	-2.45003379378826\\
7.501	-2.44934967552221\\
7.628	-2.44519513967344\\
7.756	-2.43755628669249\\
7.887	-2.42625044448576\\
8.02	-2.41129580011248\\
8.157	-2.3924019077988\\
8.299	-2.36930067451663\\
8.447	-2.34168922591514\\
8.603	-2.30903077418535\\
8.77	-2.27047906241163\\
8.951	-2.22509946143932\\
9.155	-2.17031354508292\\
9.4	-2.10078863982064\\
9.769	-1.99201434229593\\
10.23	-1.85683163229853\\
10.487	-1.78515476324565\\
10.706	-1.72759169640857\\
10.904	-1.67905407594992\\
11.089	-1.63721425471409\\
11.264	-1.60113917216487\\
11.431	-1.57019287172755\\
11.592	-1.54382698852493\\
11.748	-1.52174096557276\\
11.9	-1.50368103718634\\
12.049	-1.48945097298993\\
12.195	-1.47897944009683\\
12.338	-1.47216829746603\\
12.479	-1.46888223094295\\
12.619	-1.46905841454544\\
12.758	-1.47266905704189\\
12.897	-1.47972292455441\\
13.037	-1.4902988491552\\
13.178	-1.50442908172025\\
13.321	-1.52224441743281\\
13.467	-1.54392671745617\\
13.618	-1.56987883457773\\
13.775	-1.60040561970514\\
13.941	-1.63624979225867\\
14.12	-1.67849839856567\\
14.319	-1.72909113958349\\
14.556	-1.79304130521829\\
14.914	-1.89369290386993\\
15.352	-2.01607619845194\\
15.602	-2.08226955588376\\
15.818	-2.13596323231279\\
16.017	-2.1819246873613\\
16.206	-2.2220541788116\\
16.387	-2.2569842484557\\
16.563	-2.28745046084617\\
16.734	-2.31357173925177\\
16.901	-2.33561529259411\\
17.064	-2.35367647325526\\
17.223	-2.3678554500844\\
17.379	-2.37831174104337\\
17.531	-2.38505362899326\\
17.68	-2.38820963630954\\
17.826	-2.38784371781099\\
17.969	-2.38403939257567\\
18.11	-2.37684168699018\\
18.25	-2.36622562286594\\
18.389	-2.35221162850243\\
18.528	-2.33471769601471\\
18.668	-2.3136028662206\\
18.81	-2.2886749625421\\
18.955	-2.25970250431687\\
19.105	-2.22619755425375\\
19.263	-2.18733165726493\\
19.433	-2.14188752323172\\
19.621	-2.08798086002954\\
19.846	-2.01972396996594\\
20.216	-1.90320866604345\\
20.546	-1.80073382751252\\
20.761	-1.73756522552388\\
20.945	-1.68703411557706\\
21.113	-1.64442427278182\\
21.271	-1.6078732975614\\
21.422	-1.5764442607236\\
21.569	-1.54935693253637\\
21.713	-1.52633366987458\\
21.855	-1.50713721003824\\
21.995	-1.49168444700179\\
22.135	-1.47971396437305\\
22.274	-1.47128357254802\\
22.414	-1.46626310274747\\
22.554	-1.46470290179366\\
22.695	-1.4665958716268\\
22.837	-1.47197515857084\\
22.98	-1.48086751687545\\
23.124	-1.49329371687552\\
23.269	-1.50926761455228\\
23.416	-1.52893925586768\\
23.565	-1.55236947960134\\
23.717	-1.57978960070903\\
23.872	-1.61127674846627\\
24.032	-1.64733592482551\\
24.198	-1.68832294026513\\
24.373	-1.73513312993458\\
24.562	-1.78933358515165\\
24.774	-1.85382038510088\\
25.038	-1.93793366274721\\
25.72	-2.15702075755888\\
25.918	-2.21587097041973\\
26.09	-2.26346938124079\\
26.247	-2.30337621571118\\
26.393	-2.3369720933082\\
26.532	-2.36544917721252\\
26.666	-2.3893874906083\\
26.796	-2.40910504957895\\
26.923	-2.42487769994802\\
27.048	-2.43692325453483\\
27.172	-2.44539061236999\\
27.295	-2.45032671080493\\
27.418	-2.45180902267999\\
27.542	-2.44983313965649\\
27.667	-2.4443704638432\\
27.794	-2.43533755997566\\
27.923	-2.42268522873317\\
28.055	-2.40626020174336\\
28.191	-2.38584251636231\\
28.332	-2.36115560218306\\
28.479	-2.33187953587667\\
28.633	-2.29766846354788\\
28.797	-2.25766529292271\\
28.974	-2.2108861086904\\
29.17	-2.15543185921027\\
29.396	-2.08778799556105\\
29.691	-1.99563456848684\\
30	-1.89746860096169\\
};
\addlegendentry{$\tau_3$}

\end{axis}
\end{tikzpicture}%

%% file: TikZ/spm_chaine_cinematique.tex
\def\ScaleFactor{1}

    \begin{tikzpicture}[>=latex',every text node part/.style={align=center},rounded corners]
    	\node at (0,0) (F0) {$\bm{z}_\frakb$};
    	
    	\node[right of=F0,right=3.5*\ScaleFactor cm] (u2) {$\rep{\bm{u}_2}{\frakb}$};
    	\node[right of=u2,right=2.25*\ScaleFactor cm] (w2) {$\rep{\bm{w}_2}{\frakb}$};
    	\node[right of=w2,right=2.25*\ScaleFactor cm] (v2) {$\rep{\bm{v}_2}{\frakb}$};
    	
    	\node[above of=u2,above=0.25*\ScaleFactor cm] (u1) {$\rep{\bm{u}_1}{\frakb}$};
    	\node[above of=w2,above=0.25*\ScaleFactor cm] (w1) {$\rep{\bm{w}_1}{\frakb}$};
    	\node[above of=v2,above=0.25*\ScaleFactor cm] (v1) {$\rep{\bm{v}_1}{\frakb}$};
    	
    	\node[below of=u2,below=0.25*\ScaleFactor cm] (u3) {$\rep{\bm{u}_3}{\frakb}$};
    	\node[below of=w2,below=0.25*\ScaleFactor cm] (w3) {$\rep{\bm{w}_3}{\frakb}$};
    	\node[below of=v2,below=0.25*\ScaleFactor cm] (v3) {$\rep{\bm{v}_3}{\frakb}$};
    	
    	\node[right of=v2,right=3.5*\ScaleFactor cm] (F0t) {$\bm{z}_\star$};
    	
    	\draw[->] (u1)->(w1)node[midway,above]{$\bm{R}_z{\left(\theta_1\right)}\,\bm{R}_x{\left( \alpha_{1,1} \right)}$};
    	\draw[->] (u2)->(w2)node[midway,above]{$\bm{R}_z{\left(\theta_2\right)}\,\bm{R}_x{\left( \alpha_{1,2} \right)}$};
    	\draw[->] (u3)->(w3)node[midway,above]{$\bm{R}_z{\left(\theta_3\right)}\,\bm{R}_x{\left( \alpha_{1,3} \right)}$};
    	
    	\draw[->] (F0)--($(u1)+(-4*\ScaleFactor,0)$)--(u1)node[midway,above]{$\bm{R}_z{\left( \eta_1 \right)}\,\bm{R}_x{\left( \beta_1-\pi \right)}$};
    	\draw[->] (F0)--(u2)node[midway,above]{$\bm{R}_z{\left( \eta_2 \right)}\,\bm{R}_x{\left( \beta_1-\pi \right)}$};
    	\draw[->] (F0)--($(u3)+(-4*\ScaleFactor,0)$)--(u3)node[midway,above]{$\bm{R}_z{\left( \eta_3 \right)}\,\bm{R}_x{\left( \beta_1-\pi \right)}$};
    	
    	\draw[->] (F0t)--($(v1)+(4*\ScaleFactor,0)$)--(v1)node[midway,above]{$\bm{R}_z{\left( \eta_{1} \right)}\,\bm{R}_x{\left( -\beta_{2} \right)}$};
    	\draw[->] (F0t)--(v2)node[midway,above]{$\bm{R}_z{\left( \eta_{2} \right)}\,\bm{R}_x{\left( -\beta_{2} \right)}$};
    	\draw[->] (F0t)--($(v3)+(4*\ScaleFactor,0)$)--(v3)node[midway,above]{$\bm{R}_z{\left( \eta_{3} \right)}\,\bm{R}_x{\left( -\beta_{2} \right)}$};


		\draw[draw=none,pattern=north west lines, pattern color=black!80] (-1.125*\ScaleFactor,-3*\ScaleFactor) rectangle (-0.125*\ScaleFactor,-2.75*\ScaleFactor)node[right]{$\calB_\frakb\bydef\left\{\bm{x}_\frakb,\bm{y}_\frakb,\bm{z}_\frakb\right\}$};
		\draw[] (-1.125*\ScaleFactor,-2.75*\ScaleFactor)--(-0.125*\ScaleFactor,-2.75*\ScaleFactor);
    	\draw[->] (-0.625*\ScaleFactor,-2.75*\ScaleFactor)--(-0.625*\ScaleFactor,-1*\ScaleFactor)->(F0);
    	
    	\draw[->] (F0)--($(F0)+(-1.25*\ScaleFactor,0)$)|-node[near end,above]{$\bm{R}_z{\left(\chi_3\right)}\bm{R}_y{\left(\chi_2\right)}\bm{R}_x{\left(\chi_1\right)}$}($(F0t)+(1.25*\ScaleFactor,4*\ScaleFactor)$)|-(F0t);
    	
    	\draw[blue,<->] (w1)--(v1)node[midway,above]{$\bm{w}_1^\sfT\bm{v}_1=\cos{\left(\alpha_{2,1}\right)}$};
    	\draw[blue,<->] (w2)--(v2)node[midway,above]{$\bm{w}_2^\sfT\bm{v}_2=\cos{\left(\alpha_{2,2}\right)}$};
    	\draw[blue,<->] (w3)--(v3)node[midway,above]{$\bm{w}_3^\sfT\bm{v}_3=\cos{\left(\alpha_{2,3}\right)}$};
    \end{tikzpicture}

%% file: contents/appendix.tex
\section{Derivation of the geometric model of the 3-\texorpdfstring{\underline{R}}{R}RR 3-DOF SPM of interest}\label{a:gm}

\subsection{Expressions of the unit vectors}

Based on the convention used in \cite{Le23} and \cite{BHA08}, one can obtain the unit vectors $\bm{u}_i$, $\bm{w}_i$, and $\bm{v}_i$, with $i\in\itvd{1,3}$ by exploiting the kinematic chain of the robot depicted in Figure \ref{fig:spm_chaine_cinematique}. 

Given that $\bm{z}_\frakb\bydef\mat{0&0&1}^\sfT$, one can show that they can be written in the reference frame $\calF_\frakb\byaff\left(O,\bm{x}_\frakb,\bm{y}_\frakb,\bm{z}_\frakb\right)$ as:
\begin{equation}
	\begin{aligned}
		\rep{\bm{u}_i}{\frakb} &= \bm{R}_z{\left( \eta_i \right)}\,\bm{R}_x{\left( \beta_1-\pi \right)}\,\bm{z}_\frakb\\
		\rep{\bm{w}_i}{\frakb} &= \bm{R}_z{\left( \eta_i \right)}\,\bm{R}_x{\left( \beta_1-\pi \right)}\,\bm{R}_z{\left(\theta_i\right)}\,\bm{R}_x{\left( \alpha_{1,i} \right)}\,\bm{z}_\frakb\\
		\rep{\bm{v}_i}{\frakb} &= \bm{R}_z{\left(\chi_3\right)}\bm{R}_y{\left(\chi_2\right)}\bm{R}_x{\left(\chi_1\right)}\bm{R}_z{\left( \eta_{i} \right)}\,\bm{R}_x{\left( -\beta_{2} \right)}\,\bm{z}_\frakb
	\end{aligned}
\end{equation}

where $\bm{R}_x$, $\bm{R}_y$ and $\bm{R}_z$ denote the rotation matrices around their respective local axis. The latter matrices are defined as follows:
\begin{subequations}
	\begin{align}
		\bm{R}_x(\chi_1) &\bydef \mat{1 & 0 & 0 
		\\
		0 & \cos\left(\chi_{1}\right) & -\sin\left(\chi_{1}\right) 
		\\
		0 & \sin\left(\chi_{1}\right) & \cos\left(\chi_{1}\right) 
		}\\
		\bm{R}_y(\chi_2) &\bydef \mat{\cos\left(\chi_{2}\right) & 0 & \sin\left(\chi_{2}\right) 
		\\
		 0 & 1 & 0 
		\\
		 -\sin\left(\chi_{2}\right) & 0 & \cos\left(\chi_{2}\right) 
		}\\
		\bm{R}_z(\chi_3) &\bydef \mat{\cos\left(\chi_{3}\right) & -\sin\left(\chi_{3}\right) & 0 
		\\
		 \sin\left(\chi_{3}\right) & \cos\left(\chi_{3}\right) & 0 
		\\
		 0 & 0 & 1
		}
	\end{align}
\end{subequations}

\begin{remark}
	The disposition of the pivot linkage $\eta_i$, $i\in\itvd{1,3}$ is supposed to be identical for the base and the upper platform, as in \cite{Le23}. However, as our robot has coaxial input shafts ($\beta_1=0$), such parameters does not appear ``physically'' but still in the equations.
\end{remark}

\begin{remark}
	In the special case of SPMs with coaxial input shafts ($\beta_1=0$), one always has
	$$
	\rep{\bm{u}_i}{\frakb}=\mat{0 \\ 0 \\ -1}, \qquad\forall\,i\in\itvd{1,3}
	$$
	as mentioned in \cite{TS20}.
\end{remark}

\subsection{Expression of the geometric model}

Let $x_i=\cos{\left(\chi_i\right)}$, $y_i=\sin{\left(\chi_i\right)}$, $c_i=\cos{\left(\theta_i\right)}$, and $s_i=\sin{\left(\theta_i\right)}$, $i\in\itvd{1,3}$. The detailed geometric model of the mechanism of interest is given by:

\begin{equation}
	\begin{aligned}
		&\hspace*{1.25em}\bm{f}(\bm{\theta},\bm{\chi}) = \mat{f_1(\theta_1,\chi_1,\chi_2,\chi_3)\\f_2(\theta_2,\chi_1,\chi_2,\chi_3)\\f_3(\theta_3,\chi_1,\chi_2,\chi_3)},\quad\text{with}\\[3ex]
		f_1 &= -c_{1} x_{3} y_{2} y_{1}+s_{1} x_{3} y_{2} y_{1}+c_{1} y_{3} y_{2} y_{1}+s_{1} y_{3} y_{2} y_{1}\\
		&\hspace*{3em}-x_{2} y_{1} \sqrt{2}+c_{1} x_{3} x_{1}+s_{1} x_{3} x_{1}+c_{1} y_{3} x_{1}\\
		&\hspace*{3em}-s_{1} y_{3} x_{1}+c_{1} x_{3} x_{2}-s_{1} x_{3} x_{2}\\
		&\hspace*{3em}-c_{1} y_{3} x_{2}-s_{1} y_{3} x_{2}-y_{2} \sqrt{2}\\[2ex]
		f_2 &= c_{2} x_{3} y_{2} y_{1}+s_{2} x_{3} y_{2} y_{1}+c_{2} y_{3} y_{2} y_{1}-s_{2} y_{3} y_{2} y_{1}\\
		&\hspace*{3em}-x_{2} y_{1} \sqrt{2}+c_{2} x_{3} x_{1}-s_{2} x_{3} x_{1}-c_{2} y_{3} x_{1}\\
		&\hspace*{3em}-s_{2} y_{3} x_{1}+c_{2} x_{3} x_{2}+s_{2} x_{3} x_{2}\\
		&\hspace*{3em}+c_{2} y_{3} x_{2}-s_{2} y_{3} x_{2}+y_{2} \sqrt{2}\\[2ex]
		f_3 &= c_{3} y_{1} y_{2} y_{3}+s_{3} x_{3} y_{1} y_{2}+c_{3} x_{1} x_{3}-s_{3} x_{1} y_{3}
	\end{aligned}
\end{equation}

As highlighted in Figure \ref{fig:spm_chaine_cinematique}, such a system is obtained using a kinematic closure being the dot product between $\bm{w}_i$ and $\bm{v}_i$, $i\in\itvd{1,3}$.

%% file: main.bbl
\begin{thebibliography}{}

\bibitem[Bai et~al., 2009]{BHA08}
Bai, S., Hansen, M.~R., and Angeles, J. (2009).
\newblock A robust forward-displacement analysis of spherical parallel robots.
\newblock {\em Mechanism and Machine Theory}, 44:2204--2216.

\bibitem[Gosselin and Hamel, 1994]{GH94}
Gosselin, C. and Hamel, J.-F. (1994).
\newblock The agile eye: a high-performance three-degree-of-freedom
  camera-orienting device.
\newblock In {\em Proceedings of the 1994 IEEE International Conference on
  Robotics and Automation}, pages 781--786 vol.1.

\bibitem[Hilkert, 2008]{Hil08}
Hilkert, J. (2008).
\newblock Inertially stabilized platform technology concepts and principles.
\newblock {\em IEEE Control Systems Magazine}, 28(1):26--46.

\bibitem[Kantorovich, 1948]{Kan48}
Kantorovich, L.~V. (1948).
\newblock On {N}ewton’s method for functional equations.
\newblock {\em {F}unctional {A}nalysis and {A}pplied {M}athematics},
  59(7):1237--1240.

\bibitem[Lazard and Rouillier, 2007]{LazRou07}
Lazard, D. and Rouillier, F. (2007).
\newblock Solving parametric polynomial systems.
\newblock {\em Journal of Symbolic Computation}, 42(6):636--667.

\bibitem[Lê et~al., 2023]{Le23}
Lê, A., Rouillier, F., Rance, G., and Chablat, D. (2023).
\newblock {O}n the {C}ertification of the {K}inematics of 3-{DOF} {S}pherical
  {P}arallel {M}anipulators.
\newblock {\em Maple Transactions}.

\bibitem[Masten, 2008]{Mas08}
Masten, M.~K. (2008).
\newblock Inertially stabilized platforms for optical imaging systems.
\newblock {\em IEEE Control Systems Magazine}, 28(1):47--64.

\bibitem[Merlet, 2006]{Mer06}
Merlet, J.-P. (2006).
\newblock {\em Parallel Robots (Second Edition)}.
\newblock Solid Mechanics and Its Application. Springer.

\bibitem[Rouillier, 2007]{Rou07}
Rouillier, F. (2007).
\newblock {\em {Algorithmes pour l'{\'e}tude des solutions r{\'e}elles des
  syst{\`e}mes polynomiaux}}.
\newblock Habilitation {\`a} diriger des recherches, {Universit{\'e} Pierre \&
  Marie Curie - Paris 6}.

\bibitem[Tursynbek and Shintemirov, 2020]{TS20}
Tursynbek, I. and Shintemirov, A. (2020).
\newblock Infinite torsional motion generation of a spherical parallel
  manipulator with coaxial input axes.
\newblock {\em 2020 IEEE/ASME International Conference on Advanced Intelligent
  Mechatronics (AIM)}, pages 1780--1785.

\end{thebibliography}
